\documentclass[lettersize,journal]{IEEEtran}
\usepackage{amsmath,amsfonts}
\usepackage{algorithmic}
\usepackage{algorithm}
\usepackage{array}
\usepackage{textcomp}
\usepackage{stfloats}
\usepackage{url}
\usepackage{verbatim}
\usepackage{graphicx}
\usepackage{cite}
\usepackage[hidelinks]{hyperref}
\usepackage{subfigure}
\usepackage{multirow}
\usepackage{threeparttable}

\begin{document}

\title{Towards Efficient SDRTV-to-HDRTV by Learning from Image Formation}

\author{Xiangyu~Chen\textsuperscript{\rm *},
        Zheyuan~Li\textsuperscript{\rm *},
        Zhengwen~Zhang,
        Jimmy~S.~Ren,
        Yihao Liu,
        Jingwen He, \\
        Yu Qiao,~\IEEEmembership{Senior~Member,~IEEE,} 
        Jiantao Zhou,~\IEEEmembership{Senior~Member,~IEEE,}
        Chao~Dong 
\thanks{\IEEEcompsocthanksitem Xiangyu Chen, Zheyuan Li and Jiantao Zhou are with State Key Laboratory of Internet of Things for Smart City, University of Macau.
\IEEEcompsocthanksitem Xiangyu Chen, Zheyuan Li, Zhengwen Zhang, Yihao Liu, Jingwen He, Yu Qiao and Chao Dong are also with Shenzhen Institutes of Advanced Technology, Chinese Academy of Sciences, Shenzhen, China.
\IEEEcompsocthanksitem Xiangyu Chen, Yihao Liu, Yu Qiao and Chao Dong are also with Shanghai Artificial Intelligence Laboratory, Shanghai, China.
\IEEEcompsocthanksitem Jimmy S. Ren is with SenseTime Research, Hong Kong, China.}
\thanks{Co-first Authors: Xiangyu Chen and Zheyuan Li; Corresponding Authors: Jiantao Zhou (jtzhou@um.edu.mo) and Chao Dong (chao.dong@siat.ac.cn).}
}

\markboth{Journal of \LaTeX\ Class Files,~Vol.~14, No.~8, August~2021}%
{Shell \MakeLowercase{\textit{et al.}}: A Sample Article Using IEEEtran.cls for IEEE Journals}


\maketitle

\begin{abstract}
Modern displays can render video content with high dynamic range (HDR) and wide color gamut (WCG). However, most resources are still in standard dynamic range (SDR). Therefore, transforming existing SDR content into the HDRTV standard holds significant value. This paper defines and analyzes the SDRTV-to-HDRTV task by modeling the formation of SDRTV/HDRTV content. Our findings reveal that a naive end-to-end supervised training approach suffers from severe gamut transition errors. To address this, we propose a new three-step solution called HDRTVNet++, which includes adaptive global color mapping, local enhancement, and highlight refinement. The adaptive global color mapping step utilizes global statistics for image-adaptive color adjustments. A local enhancement network further enhances details, and the two sub-networks are combined as a generator to achieve highlight consistency through GAN-based joint training. Designed for ultra-high-definition TV content, our method is both effective and lightweight for processing 4K resolution images. We also constructed a dataset using HDR videos in the HDR10 standard, named HDRTV1K, containing 1235 training and 117 testing images, all in 4K resolution. Additionally, we employ five metrics to evaluate SDRTV-to-HDRTV performance. Our results demonstrate state-of-the-art performance both quantitatively and visually. The codes and models are available at \url{https://github.com/xiaom233/HDRTVNet-plus}.
\end{abstract}

\begin{IEEEkeywords}
Image processing, Image Enhancement, Gamut extension.
\end{IEEEkeywords}

\section{Introduction}
\IEEEPARstart{T}{HE} evolution of television and film content resolution has progressed from standard definition (SD) to full high definition (FHD), and most recently, to ultra-high definition (UHD). 
A key feature of UHDTV is high dynamic range (HDR), which offers a wider color gamut and higher dynamic range than standard dynamic range (SDR) content, allowing viewers to experience images and videos closer to real life.
%
%
Although HDR display devices are now common, most available resources remain in the SDR format, necessitating algorithms to convert SDR content to HDR.
This task, known as SDRTV-to-HDRTV, holds significant practical value but has been relatively underexplored. The primary reasons are twofold: first, HDRTV standards (e.g., HDR10 and HLG) have only recently become well-defined; second, there is a scarcity of large-scale datasets for training and testing.

\begin{figure}
	\tiny
	\centering
	\setlength{\tabcolsep}{0.01cm}
	\begin{tabular}{ccccc}
		\includegraphics[height=0.6in]{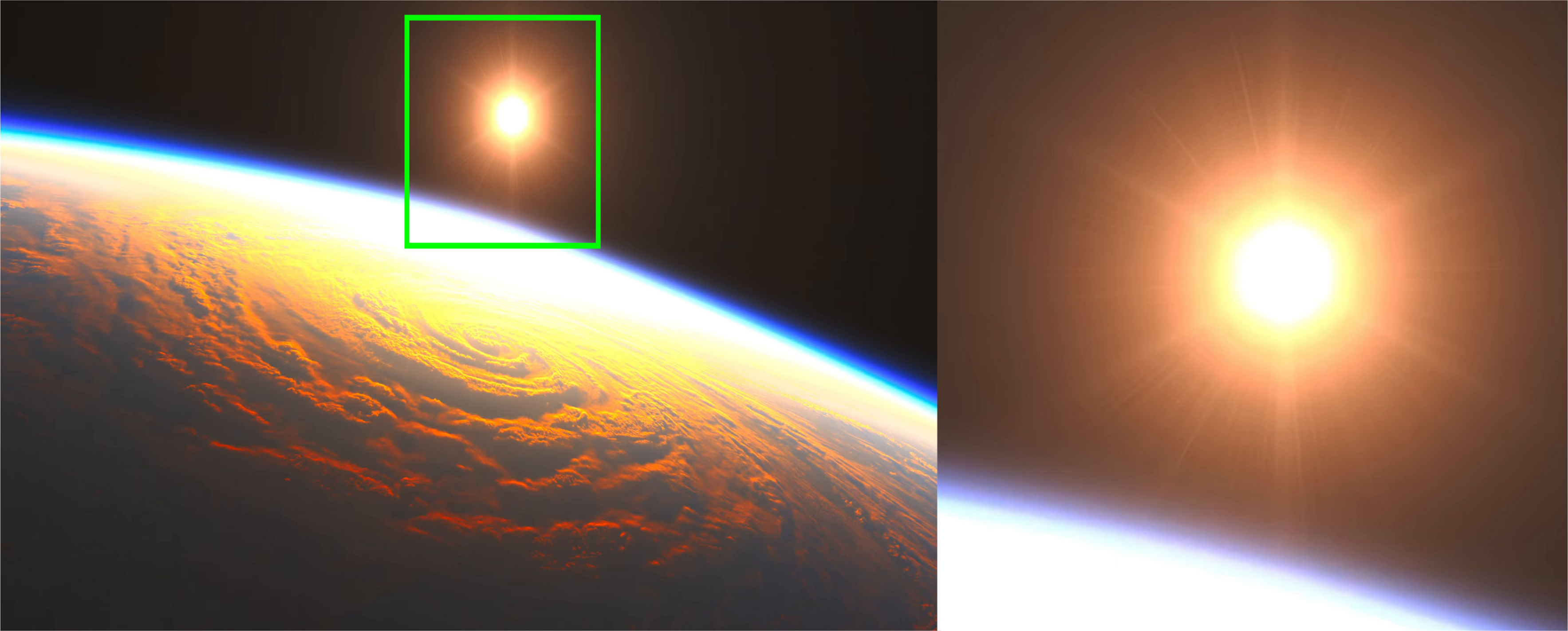}
		&\includegraphics[height=0.6in]{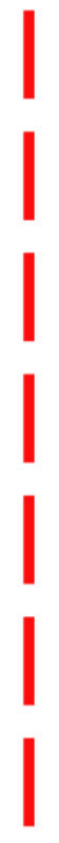}
		&\includegraphics[height=0.6in]{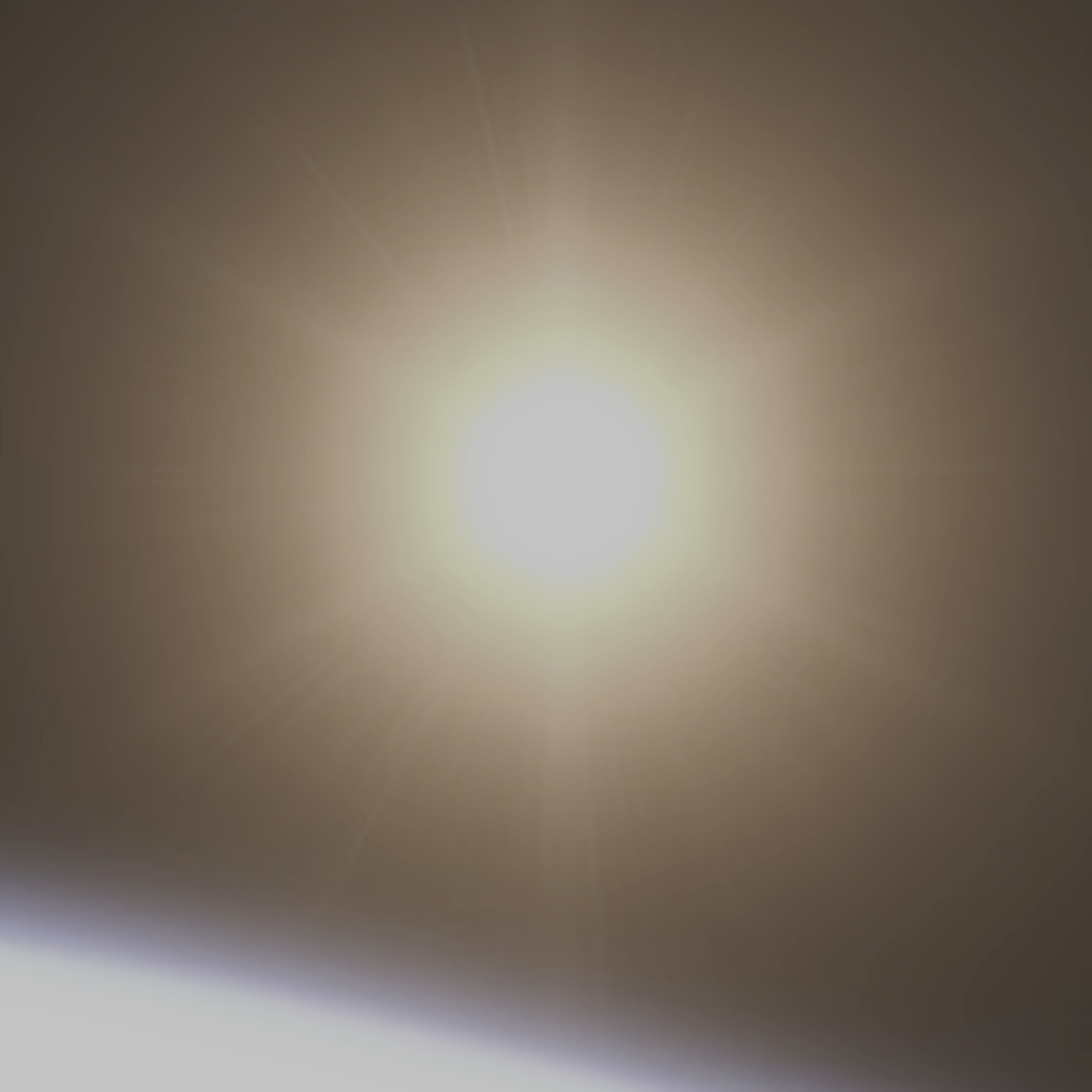}
		&\includegraphics[height=0.6in]{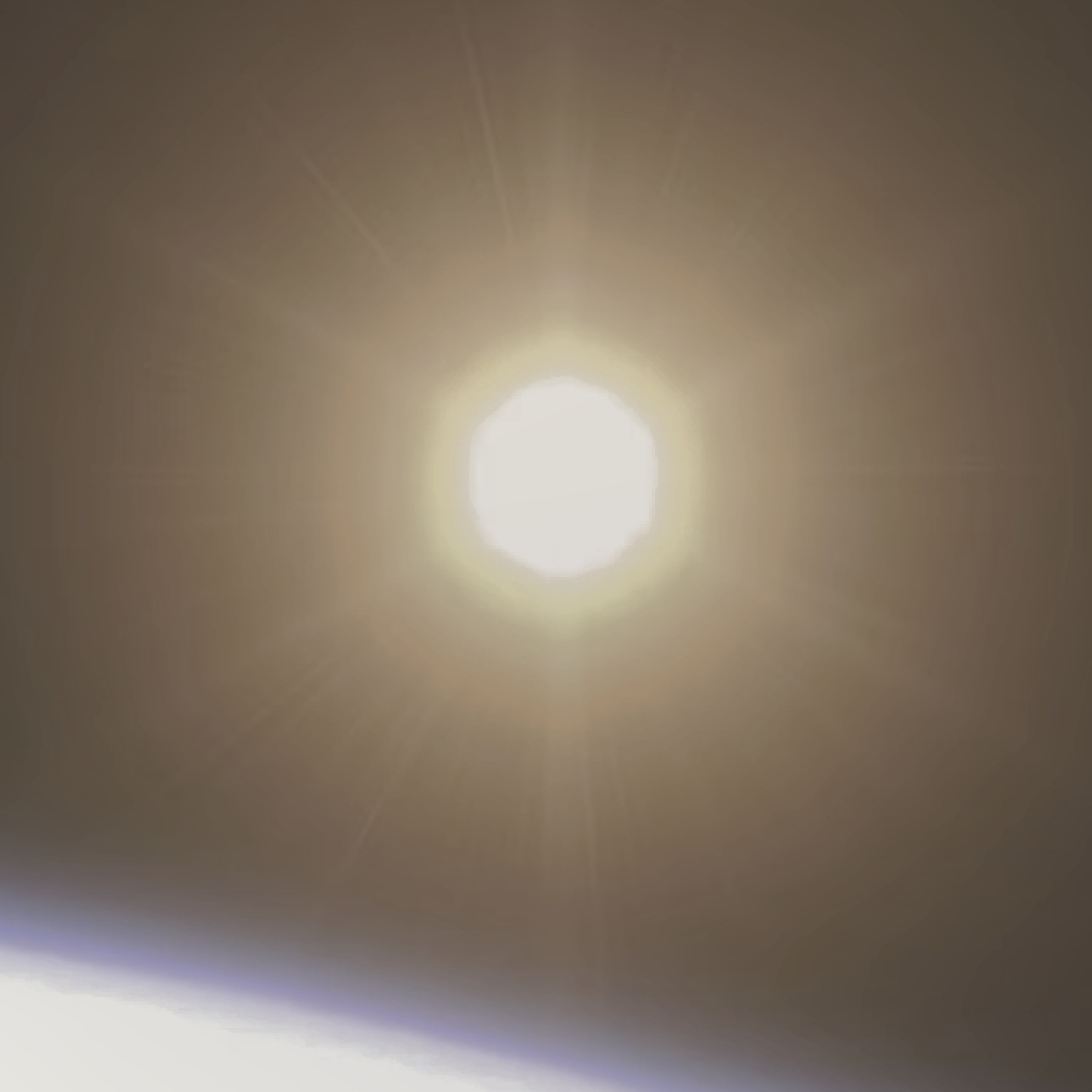}
		&\includegraphics[height=0.6in]{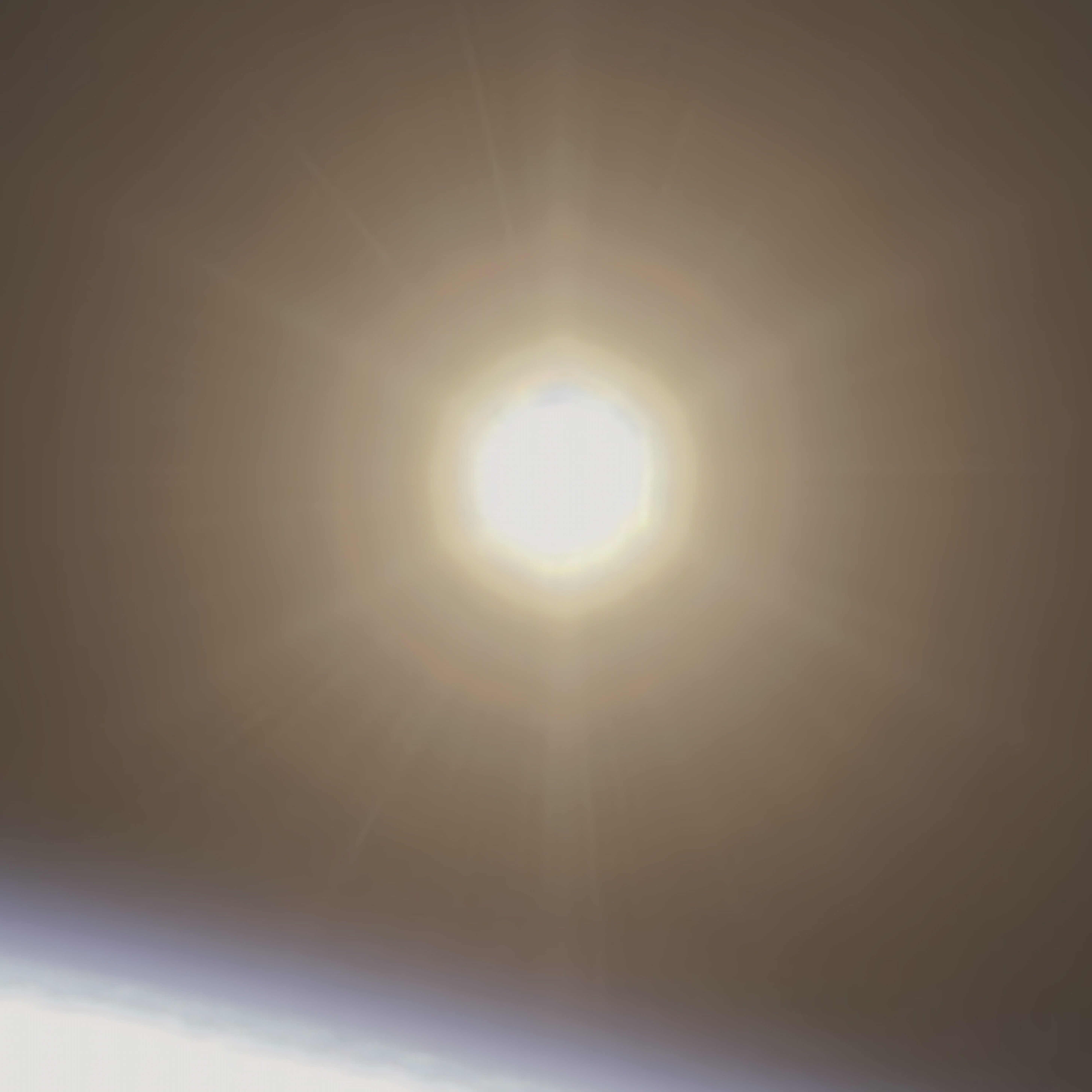}
		\\
		SDRTV& &HuoPhyEO \cite{huo2014physiological}&HDRNet\cite{gharbi2017deep}& HDRTVNet\cite{hdrtvnet_conf}
		\\ 
        \includegraphics[height=0.6in]{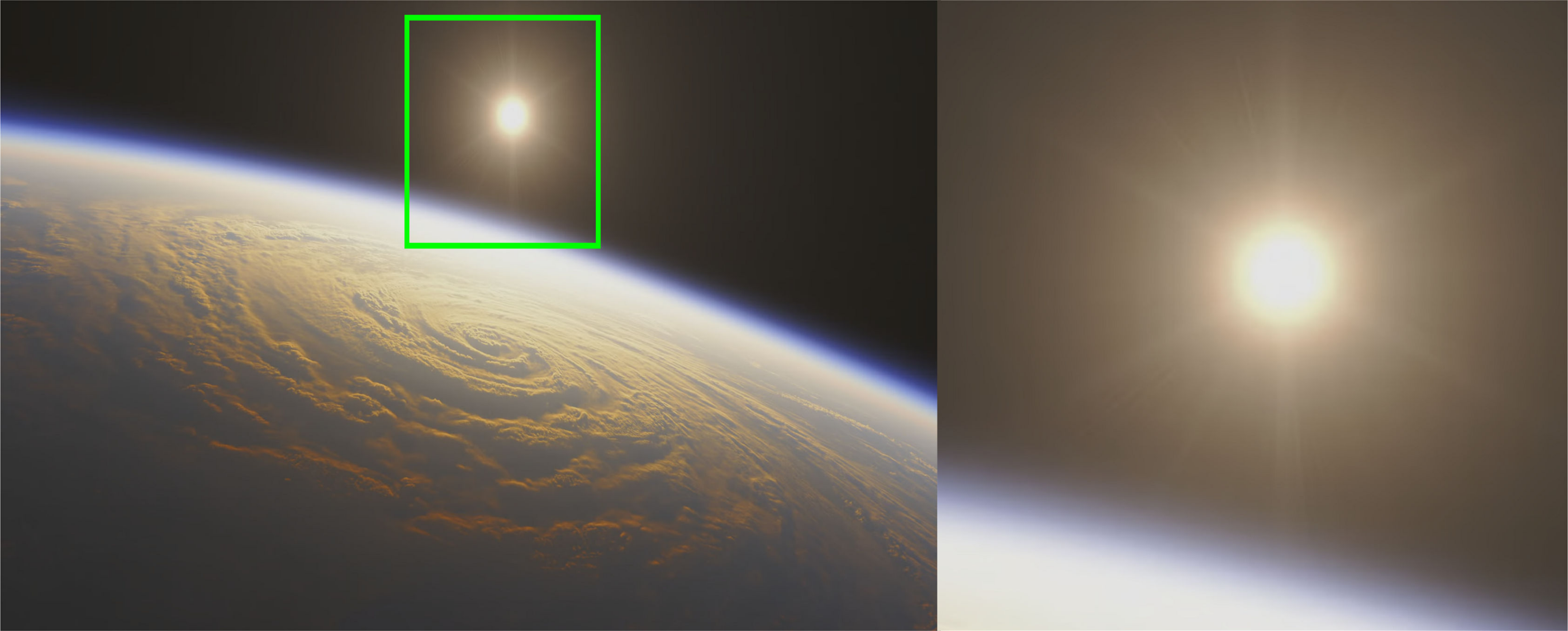}
        &\includegraphics[height=0.6in]{line.pdf}
		&\includegraphics[height=0.6in]{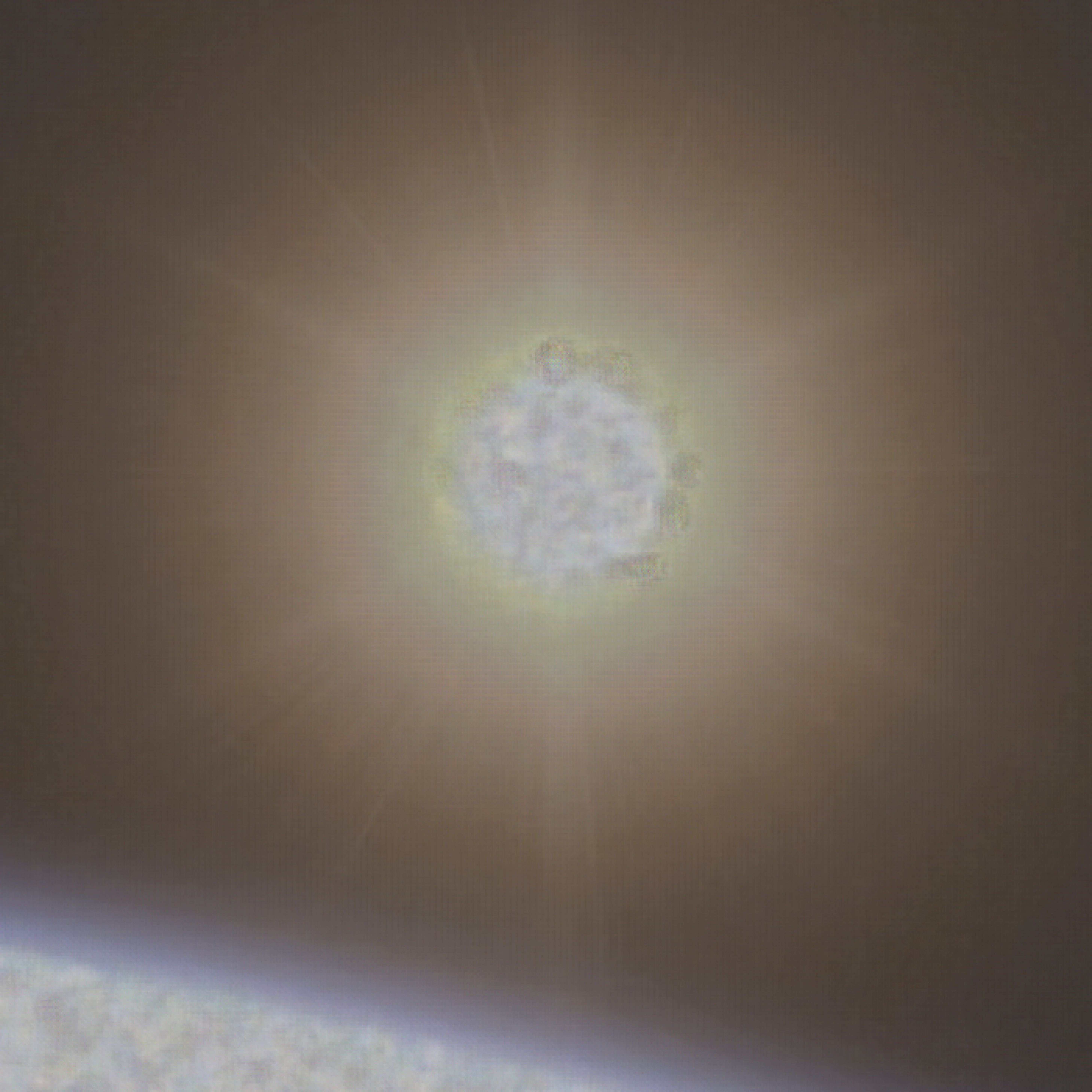}
		&\includegraphics[height=0.6in]{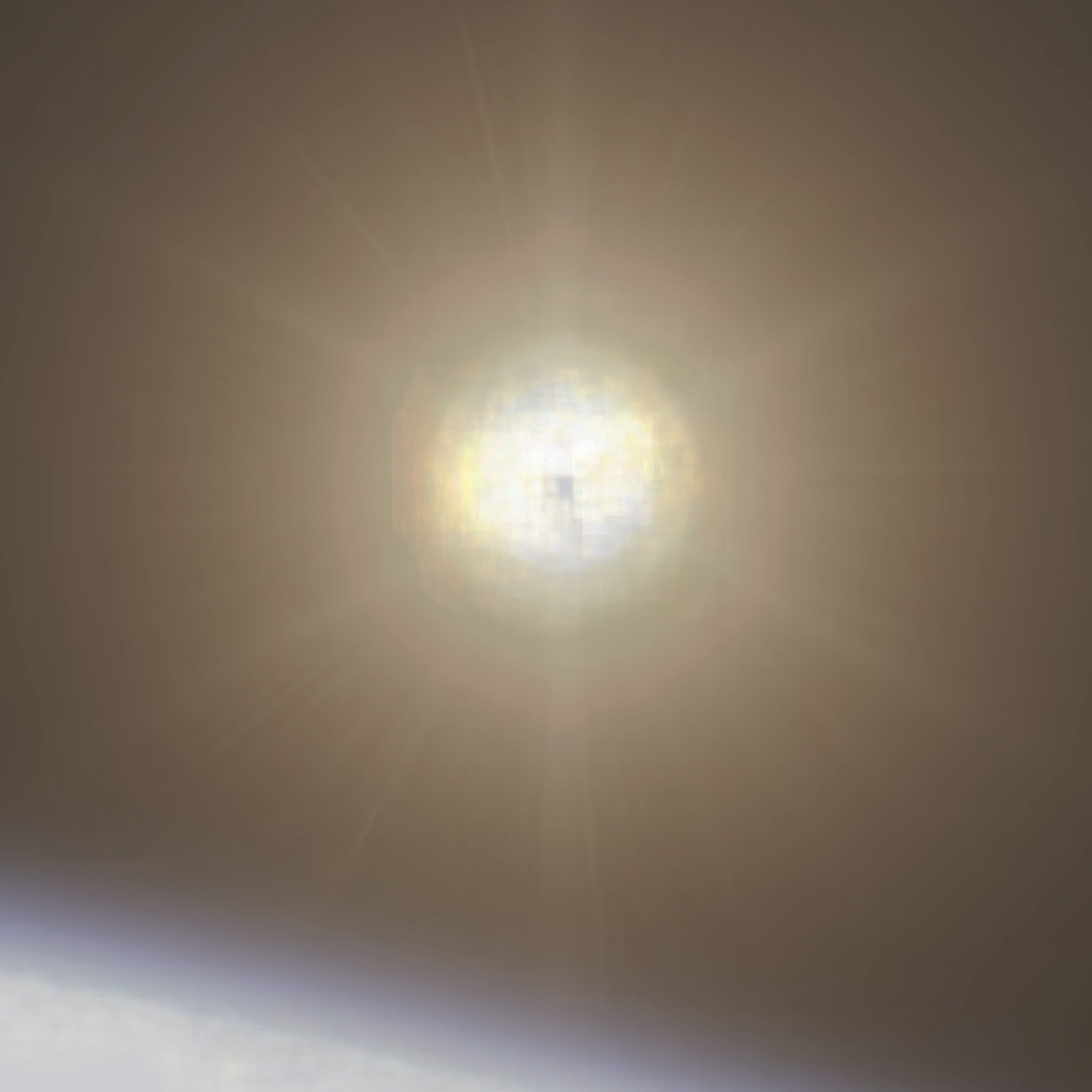}
		&\includegraphics[height=0.6in]{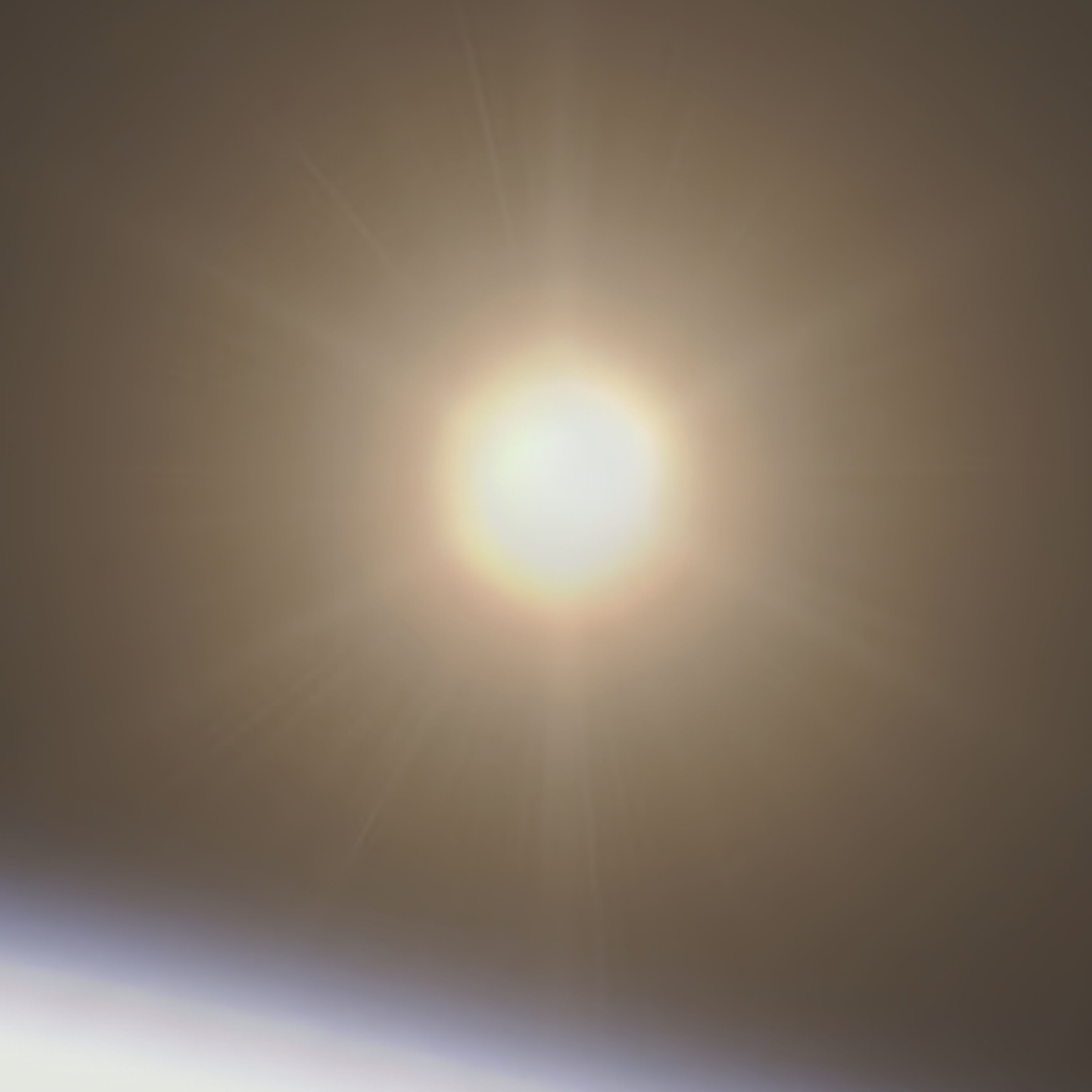}
		\\
		HDRTV& &Pixel2Pixel\cite{isola2017image}&Deep SR-ITM\cite{kim2019deep}&HDRTVNet++
		\\ 
	\end{tabular}
    \caption{Visual comparison of different methods to solve SDRTV-to-HDRTV. }
   \label{fig:teaser}
   \vspace{-10pt}
\end{figure}

To advance this emerging field, this paper conducts an in-depth study of the SDRTV-to-HDRTV problem. 
This task is challenging due to differences in dynamic range, color gamuts, and bit-depths between the two content types. While both SDRTV and HDRTV are derived from the same raw files, they adhere to different processing standards. 
It is important to note that SDRTV-to-HDRTV differs from the low dynamic range (LDR)-to-HDR task, which involves predicting HDR scene luminance in the linear domain, closer to the raw file.
In Section \ref{Preliminary}, we provide detailed explanations of SDRTV/HDRTV concepts and the SDRTV-to-HDRTV task. 
From an imaging formation perspective, SDRTV-to-HDRTV can be viewed as an image-to-image translation task. Due to substantial differences in color gamut, photo retouching methods can be applied to manage color transformations in this task. 
Although not exclusively focused on SDRTV-to-HDRTV, early works like Depp SR-ITM \cite{kim2019deep} and JSI-GAN \cite{kim2020jsi} address this task by combining super-resolution with SDRTV-to-HDRTV.
%
%
We illustrate SDRTV-to-HDRTV and compare various methods in Figure~\ref{fig:teaser}. As shown, previous methods struggle to effectively address this task.
Our paper aims to tackle SDRTV-to-HDRTV through a deep understanding of the underlying challenges. We introduce a simplified formation pipeline for SDRTV/HDRTV content, comprising tone mapping, gamut mapping, transfer function, and quantization. Based on this, we propose HDRTVNet++, which includes adaptive global color mapping (AGCM), local enhancement (LE), and highlight refinement (HR).
Specifically, AGCM employs a new color condition block to extract global image priors and adapt to different images. It uses only 1×1 filters to achieve superior performance with fewer parameters compared to other photo retouching methods such as CSRNet \cite{he2020conditional}, HDRNet \cite{gharbi2017deep} and Ada-3DLUT \cite{zeng2020learning}.
AGCM uses a new color condition block to extract global image priors and adapt to different images, utilizing only 1×1 filters for superior performance with fewer parameters. Following AGCM, we design a U-shape network with spatial conditions for LE to achieve local enhancements. This approach avoids color transition artifacts often produced by end-to-end networks. After training AGCM and LE, joint finetuning further improves results. Despite these advancements, highlight areas remain challenging due to severe information loss. To address this, we adopt generative adversarial training for highlight refinement.
To advance research in this area, we have constructed a new dataset called HDRTV1K and selected five evaluation metrics: PSNR, SSIM, SR-SIM \cite{zhang2012sr}, $\Delta E_{ITP}$ \cite{ITP} and HDR-VDP3 \cite{mantiuk2011hdr}.
In summary, our contributions are four-fold:
\begin{itemize}
\item We conduct a detailed analysis of the SDRTV-to-HDRTV task by modeling SDRTV/HDRTV content formation.
\item We propose an efficient SDRTV-to-HDRTV method achieving state-of-the-art performance.
\item We present a global color mapping network with outstanding accuracy and only 35K parameters.
\item We provide an HDRTV dataset and select five metrics for evaluating SDRTV-to-HDRTV algorithms.
\end{itemize}
A preliminary version of this work was presented at ICCV 2021 \cite{hdrtvnet_conf}.
Since then, several studies have explored the SDRTV-to-HDRTV problem \cite{xu2022fmnet,he2022sdrtv,cheng2022towards,yao2023bidirectional,guo2023learning,zhang2024effihdr}.
This paper introduces HDRTVNet++, an enhanced version that significantly improves upon the initial method through a refined pipeline and more effective network design. Key advancements include: 1) \textbf{Enhanced Network Design:} We propose HDRTVNet++ with an improved pipeline and network architecture. Notably, we jointly train AGCM and LE, achieving better restoration accuracy. The heavy sub-network for highlight generation is replaced with a joint adversarial training strategy, leading to a performance gain of 0.99 dB on PSNR with reduced parameters. 2) \textbf{Detailed Analysis:} We provide detailed explanations of the motivation and rationale behind our solution pipeline. By formulating pixel-independent and region-dependent operations, we highlight the importance of the correct sequence of global color mapping and local enhancement. Additional experiments further illustrate this crucial point for SDRTV-to-HDRTV conversion. 3) \textbf{Extensive Experiments:} We conduct comprehensive ablation studies and detailed investigations of the network design. These experiments demonstrate the effectiveness of our proposed modules and the overall method.
\section{Preliminary}
\label{Preliminary}
In this section, we clarify the concepts of SDRTV/HDRTV and the distinctions between SDRTV-to-HDRTV and LDR-to-HDR, as these terms are often confused and underexplored in existing literature.
\textbf{Concept.}
We use SDRTV/HDRTV to denote content (images and videos) adhering to the respective standards. SDRTV is defined by standards such as Rec.709~\cite{rec709} and BT.1886~\cite{bt1886}, while HDRTV is specified in Rec.2020~\cite{rec2020} and BT.2100~\cite{bt2100}. Key elements of HDRTV include a wide color gamut~\cite{rec2020}, PQ or HLG OETF~\cite{rec2020}, and 10-16 bit depth. Content not conforming to HDRTV is generally considered SDR, with clearer requirements outlined under the SDRTV standard, such as the Rec.709 color gamut and gamma-OETF.
Both SDRTV and HDRTV can encode the same content, but they differ in information capacity, resulting in distinct visual experiences. The terms LDR and SDR both refer to low dynamic range content, but their usage varies. SDR typically pertains to display standards used in content production, often derived from high dynamic range RAW files. In contrast, LDR content refers to images captured at specific exposure levels, with dynamic range determined during imaging. Thus, their formation processes differ.
For clarity, we use the terms \textbf{LDR-to-HDR} and \textbf{SDRTV-to-HDRTV} to represent the conventional image HDR reconstruction and the up-conversion of content from SDRTV to HDRTV standard.
\textbf{Explanation.}
SDRTV-to-HDRTV differs functionally from LDR-to-HDR. While both involve HDR, the meaning of HDR varies. LDR-to-HDR methods predict luminance in the linear domain, representing the physical brightness of a scene. Here, HDR refers to dynamic range information beyond the capture capabilities of LDR imaging.
Conversely, SDRTV-to-HDRTV involves predicting HDR images in the pixel domain using HDR display formats like HDR10, HLG, and Dolby Vision. Both SDRTV and HDRTV content originate from the same HDR scene radiance. While HDRTV content can derive from linear domain HDR content, this requires additional operations such as tone mapping and gamut mapping. As a result, methods for these tasks are not interchangeable.

\section{Related Work}
\label{related_work}

\subsection{SDRTV-to-HDRTV}
The SDRTV-to-HDRTV task, central to this paper, is initially presented in \cite{kim2019itm}, where SDRTV/HDRTV content is referred to as LDR/HDR. Notable advancements, such as Deep SR-ITM~\cite{kim2019deep} and JSI-GAN~\cite{kim2020jsi}, have successfully combined image super-resolution with SDRTV-to-HDRTV, significantly drawing research interest. 
Our conference version, HDRTVNet \cite{hdrtvnet_conf}, provides an in-depth analysis, a foundational solution, and a dataset for this task. Building on that, several works have emerged. 
For instance, He et al.\cite{he2022sdrtv} introduce HDCFM, employing hierarchical global and local feature modulation. Xu et al. \cite{xu2022fmnet} propose FMNet, a Frequency-aware Modulation Network to minimize structural distortions and artifacts. Cheng et al.\cite{cheng2022towards} develop a learning-based approach for synthesizing realistic SDRTV-HDRTV pairs, enhancing method generalization. Guo et al.~\cite{guo2023learning} contribute a new dataset and degradation models for practical conversion. Zhang et al.~\cite{zhang2024effihdr} design a efficient framework with reconstruction and enhancement models for this task.
In this work, we enhance the previous HDRTVNet \cite{hdrtvnet_conf} by introducing more effective networks, along with more comprehensive analysis and experiments.

\subsection{LDR-to-HDR}
In general, LDR-to-HDR, also known as inverse tone mapping, aims to predict HDR images from LDR photographs. Common methods include fusing multi-exposure LDR images \cite{kou2017intelligent,tan2021deep,perez2022ntire,yan2019attention} and reconstructing HDR images from a single image~\cite{banterle2009high,eilertsen2017hdr,liu2020single,hu2022hdr}. The latter is more relevant to this work. Traditional single-image LDR-to-HDR methods exploit internal image characteristics to predict scene luminance. For example, \cite{banterle2009high} estimate the density of light sources to expand the dynamic range, and \cite{huo2014physiological} apply a cross-bilateral filter to enhance input LDR images. Deep learning-based methods have also emerged. \cite{eilertsen2017hdr} propose HDRCNN to recover missing details in over-exposed regions, and \cite{liu2020single} learns the LDR-to-HDR mapping by reversing the camera pipeline. However, these approaches primarily aim at predicting linear HDR luminance and are not designed for content conversion under two display standards. Consequently, they are either hard to use for SDRTV-to-HDRTV or perform poorly when applied.

\subsection{Gamut Extension}
Gamut extension, a key concept in color science, involves converting content to a wider color gamut, essential in transitioning from Rec.709 to Rec.2020 during SDRTV-to-HDRTV conversion. Despite ITU-R~\cite{colorconversion} offering a color conversion matrix, it cannot consider multiple mappings involving color (i.e., tone mapping) used in production, limiting its effectiveness.
Several gamut extension algorithms have been proposed~\cite{zamir2014gamut,zamir2015gamut,schweiger2017luminance,zamir2017gamut,xu2018color,zamir2019vision}. For example, \cite{zamir2014gamut} introduces a perceptually-based variational framework for spatial gamut mapping, and \cite{zamir2019vision} presented a PDE-based optimization procedure considering hue, chroma, and saturation. However, these algorithms primarily address color mapping, lacking the capability for complex detail enhancement needed in SDRTV-to-HDRTV.

\begin{figure}[t]
\centering
\subfigure[SDRTV-to-HDRTV formation pipeline.]{
\label{SDRTV_HDRTV_formation_pipeline}
\begin{minipage}[t]{\linewidth}
\centering
\includegraphics[width=\linewidth]{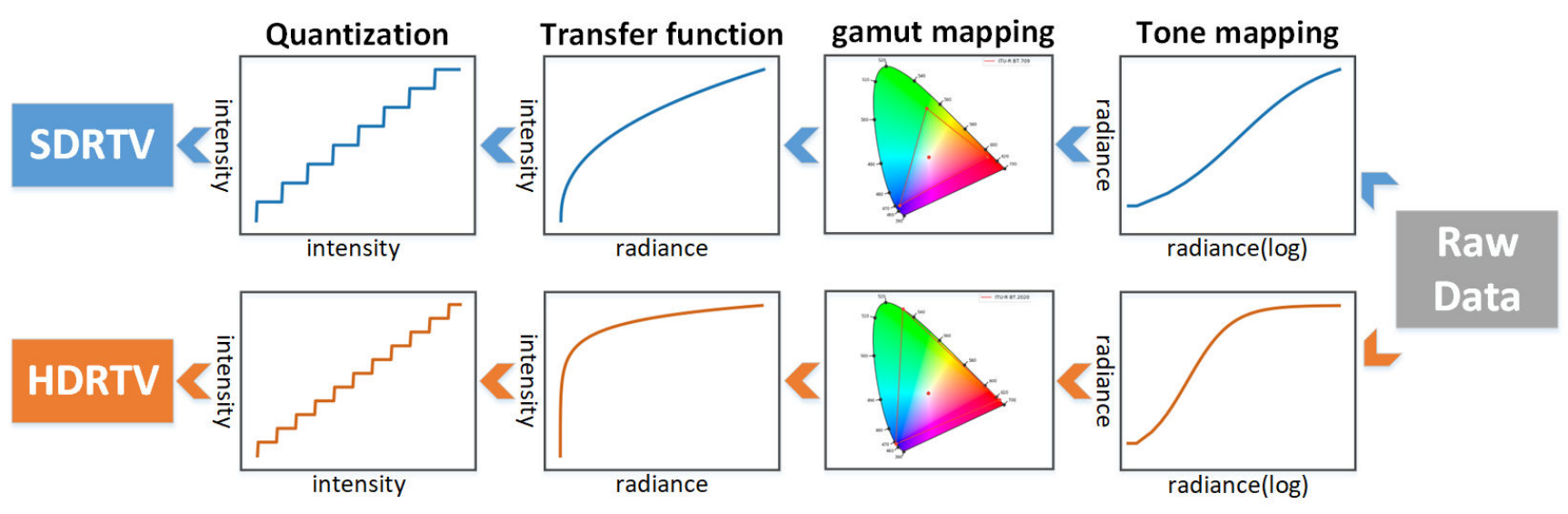}
\end{minipage}%
}

\subfigure[LDR image formation pipeline in SingleHDR \cite{liu2020single}.]{
\label{LDR_imgae_formation_pipeline_in_SingleHDR}
\begin{minipage}[t]{\linewidth}
\centering
\includegraphics[width=\linewidth]{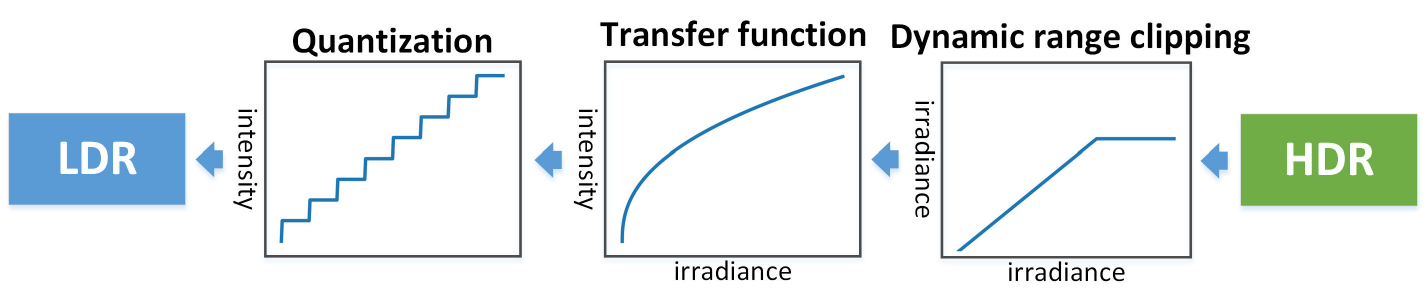}
\end{minipage}%
}
 \caption{Analysis of SDRTV-to-HDRTV and LDR-to-HDR formations.}
\label{fig2:image_formation_pipeline}
\vspace{-10pt}
\end{figure}

\section{Analysis and Method}
In this section, we introduce a streamlined SDRTV/HDRTV formation pipeline, highlighting critical steps in actual production. We then analyze this pipeline and propose a new solution using a divide-and-conquer approach.

\subsection{SDRTV/HDRTV Formation Pipeline}
We present a simplified pipeline for SDRTV and HDRTV formation, grounded in camera ISP and HDRTV content production \cite{bt2390}, as shown in Figure \ref{SDRTV_HDRTV_formation_pipeline}. While operations like denoising, white balance, and color grading are not covered, we focus on the key differences: tone mapping, gamut mapping, opto-electronic transfer function, and quantization. In the following equations, ``S'' represents SDRTV, and ``H'' represents HDRTV.

\textbf{Tone mapping}. This process converts high dynamic range signals to low dynamic range for display compatibility. It includes global tone mapping \cite{drago2003adaptive, reinhard2002parameter, tumblin1993tone} and local tone mapping \cite{larson1997visibility, lischinski2006interactive}. Global tone mapping applies a uniform function to all pixels, based on global image statistics like average luminance, whereas local tone mapping adapts to content but is computationally intensive. Thus, global tone mapping is preferred in SDRTV/HDRTV. The global tone mapping can be formulated as:
\begin{equation}
   I_{tS}=T_{S}(I|\theta_{S}),\ I_{tH}=T_{H}(I|\theta_{H}),
\end{equation}
where $T_{S}$ and $T_{H}$ are the tone mapping functions, and $ \theta_{S} $ and $ \theta_{H} $ are coefficients related to image statistics. S-shape curves are commonly used for global tone mapping, while clipping operations often occur during the actual production. We take Hable tone mapping \cite{hable2010uncharted} as an example and show its curves when processing SDRTV (0 - 100$cd/m^2$) and HDRTV (0 - 10000$cd/m^2$) in Figure \ref{SDRTV_HDRTV_formation_pipeline}.

\textbf{Gamut mapping}. This converts colors from the source to the target gamut while preserving scene appearance. According to ITU-R standards \cite{rec709, rec2020}, transformations from XYZ space to SDRTV (Rec.709) and HDRTV (Rec.2020) are:
\begin{equation}
  \begin{bmatrix}
      R_{709}\\ G_{709}\\ B_{709}
      \end{bmatrix}=M_{S}
      \begin{bmatrix}
         X \\ Y \\ Z
      \end{bmatrix}, 
    \begin{bmatrix}
      R_{2020}\\ G_{2020}\\ B_{2020}
      \end{bmatrix}=M_{H}
      \begin{bmatrix}
         X \\ Y \\ Z
      \end{bmatrix},
\end{equation}
where $M_{S}$ and $M_{H}$ are constant $3\times 3$ matrices. The CIE chromaticity diagram in Figure~\ref{SDRTV_HDRTV_formation_pipeline} differentiates the target gamuts of SDRTV and HDRTV.

\textbf{Opto-electronic transfer function}. This function converts linear optical signals into non-linear electronic signals. For SDRTV, it approximates a gamma function as $I_{fS}=f_{S}(I)=I^{1/2.2}$. For HDRTV, several OETFs exist for different standards, such as PQ-OETF \cite{PQ} for the HDR10 and HLG-OETF \cite{bt2100} for the HLG (Hybrid Log-Gamma). The PQ-OETF is:
\begin{equation}
  I_{fH}=f_{H}(I)=(\frac{a_{1}+a_{2} I^{b_{1}}}{1+a_{3} I^{b_{1}}})^{b_{2}},
\end{equation}
where $a_{1}, a_{2}, a_{3}, b_{1}, b_{2}$ are constants. The curves of gamma-OETF for SDRTV (0 - 100$cd/m^2$) and PQ-OETF for HDRTV (0 - 10000$cd/m^2$) are depicted in Figure \ref{SDRTV_HDRTV_formation_pipeline}. 

\textbf{Quantization}. This involves encoding pixel values using:
\begin{equation}
  I_{q}=Q(I,n)=\frac{\lfloor (2^{n}-1)\times I+0.5 \rfloor}{2^{n}-1},
\end{equation}
where $n$ is 8 for SDRTV and 10-16 for HDRTV. Figure \ref{SDRTV_HDRTV_formation_pipeline} also presents the two quantization curves. 

In summary, the SDRTV and HDRTV content formation pipelines are expressed as:
\begin{equation}
   I_{S}=Q_{S}\circ f_{S}\circ M_{S}\circ T_{S}(I_{raw}),
\label{S}
\end{equation}
\begin{equation}
   I_{H}=Q_{H}\circ f_{H}\circ M_{H}\circ T_{H}(I_{raw}),
\label{H}
\end{equation}
where $\circ$ denotes the connection between two operations.

\textbf{Comparison with LDR formation pipeline}. 
In SingleHDR \cite{liu2020single}, the LDR image formation pipeline comprises dynamic range clipping, non-linear mapping, and quantization.
Unlike LDR-to-HDR, SDRTV and HDRTV are generated from the same raw data using different operations, as shown in Eqs. (\ref{S}) and (\ref{H}). 
Gamut extension is critical in SDRTV-to-HDRTV, illustrating key production differences.

\subsection{SDRTV-to-HDRTV Solution Pipeline}
\begin{figure}[!t]
\centering

\subfigure[The proposed solution pipeline.]{
\label{proposed_solution_pipeline}
\begin{minipage}[t]{\linewidth}
\centering
\includegraphics[width=\linewidth]{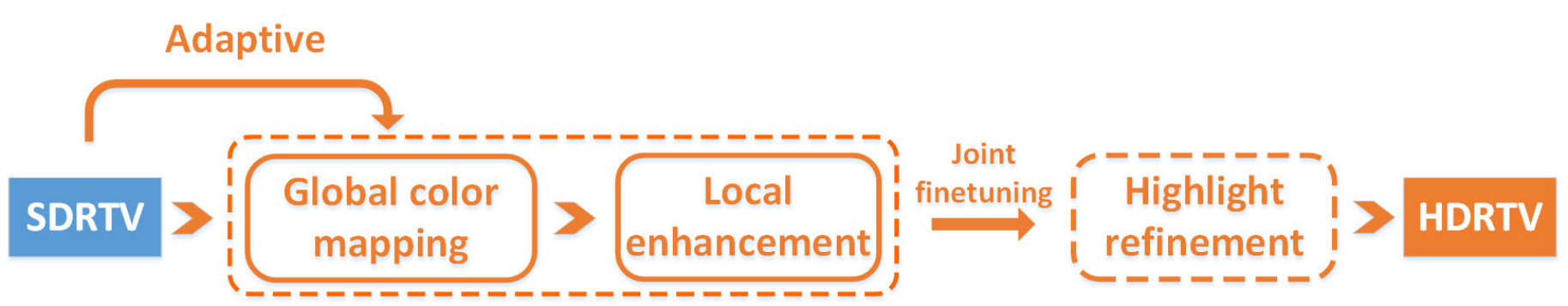}
\end{minipage}%
}

\subfigure[The end-to-end solution.]{
\label{end2end_solution}
\begin{minipage}[t]{\linewidth}
\centering
\includegraphics[width=\linewidth]{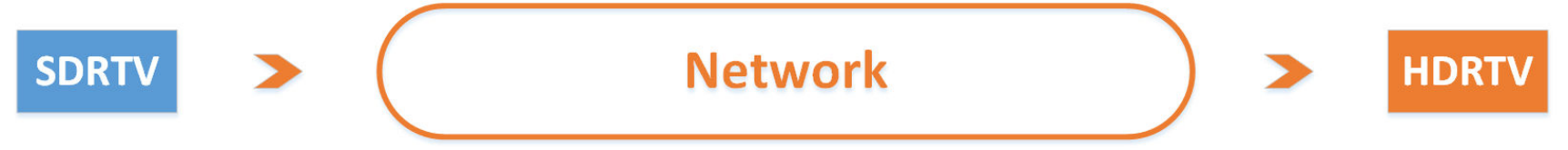}
\end{minipage}%
}

\subfigure[LDR-to-HDR based solution.]{
\label{LDR2HDR_based_solution}
\begin{minipage}[t]{\linewidth}
\centering
\vspace{-1pt}
\includegraphics[width=\linewidth]{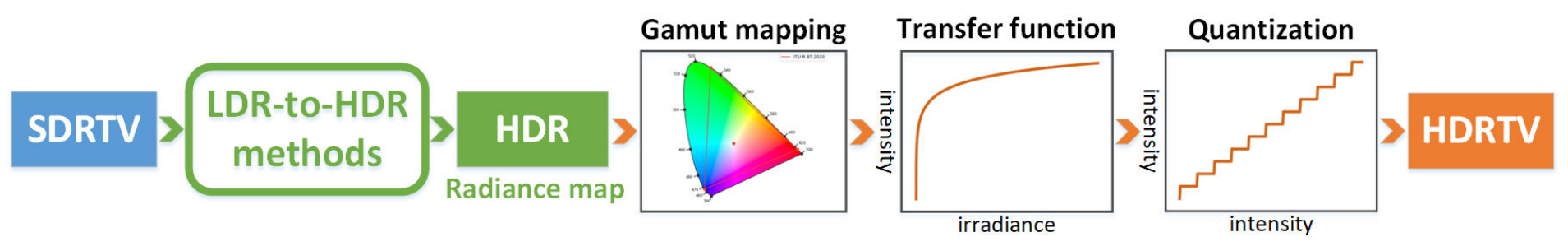}
\end{minipage}%
}
\caption{SDRTV-to-HDRTV solution pipelines.}
\label{solution_pipelines}
\vspace{-12pt}
\end{figure}

Based on the above pipeline, the SDRTV-to-HDRTV process can be formulated as:
\begin{equation}
   I_{H}=Q_{H}\circ f_{H}\circ M_{H}\circ T_{H} \circ T_{S}^{-1}\circ M_{S}^{-1}\circ f_{S}^{-1}\circ Q_{S}^{-1}(I_{S}),
\end{equation} 
where $T_{S}^{-1}, M_{S}^{-1}, f_{S}^{-1}, Q_{S}^{-1}$ are the inversions of corresponding operations. We propose a new solution pipeline as shown in Figure \ref{proposed_solution_pipeline}), based on two observations:
the one is that many critical operations, such as global tone mapping, OETF, and gamut mapping, are pixel-independent; the other one is that some operations, like local tone mapping and dequantization, depend on regional information.
%

%
To understand these operations, we define the mapping $f(\cdot)$ from the input $I_{in}$ to the output $I_{out}$ at $(x,y)$ as:
\begin{equation}
    I_{out}(x,y) = f(\Omega(I_{in}(x,y), \delta)),
\end{equation}
where $\Omega(I_{in}(x,y), \delta)$ is a local region. It composed of pixels whose distance from the center point $I_{in}(x,y)$ is no more than $\delta$. Specially, $\delta=1$ means that $\Omega(I_{in}(x,y), \delta)$ is equivalent to $I_{in}(x,y)$. For pixel-independent operations:
\begin{equation}
    I_{out}(x,y) = f(\Omega(I_{in}(x,y), 1)) = f(I_{in}(x,y)),
\end{equation}
and for region-dependent operations:
\begin{equation}
    I_{out}(x,y) = f(\Omega(I_{in}(x,y), \delta)), {\rm where}~\delta>1.
\end{equation}

Color conversion is crucial in SDRTV-to-HDRTV production \cite{colorconversion}, so we implement pixel-independent and region-dependent operations separately, i.e., global color mapping and local enhancement in Figure \ref{proposed_solution_pipeline}. Global color mapping should be image-adaptive (e.g., the average brightness and peak brightness are often used in tone mapping):
\begin{equation}
    I_{out}(x,y) = f(I_{in}(x,y)|I_{in}).
\end{equation}
Experiments in Section \ref{comparison_with_other_methods} demonstrate the effectiveness and efficiency of our design. Performing pixel-independent operations first reduces color transition artifacts compared to end-to-end solutions, as shown in Figure \ref{visual_comparison_w_wo_AGCM}.
This is likely due to the difficulty of convolution operators processing both pixel-independent low-frequency and region-dependent high-frequency transformations. For better performance, we optimize these operations jointly.
Due to severe information loss in SDRTV, particularly in highlights, previous methods \cite{hdrtvnet_conf} struggle to recover missing information. However, generative adversarial training enhances visual quality by improving color transitions in highlights, aligning predictions closer to HDRTV distribution.
We compare different SDRTV-to-HDRTV pipelines in Figure~\ref{solution_pipelines}.
Existing methods \cite{kim2019deep,kim2020jsi,xu2022fmnet} employ end-to-end networks in Figure~\ref{end2end_solution}. 
Since LDR-to-HDR has been extensively discussed, we also demonstrate a method pipeline based on LDR-to-HDR principles in Figure \ref{LDR2HDR_based_solution}. Specifically, the HDR radiance map is first generated. Then, gamut mapping is applied to convert the radiance map to the Rec.2020 gamut. The PQ OETF is subsequently used to compress the dynamic range. Finally, quantization is performed to produce the HDRTV output. The details of these operations may vary in actual processing, and thus we follow the pipeline used in \cite{kim2019deep, kim2020jsi}. 
Our solution uses a divide-and-conquer strategy, developing HDRTVNet++ with adaptive global color mapping, local enhancement, and highlight refinement.

\begin{figure*}[!t]
   \begin{center}
   \includegraphics[width=0.96\linewidth]{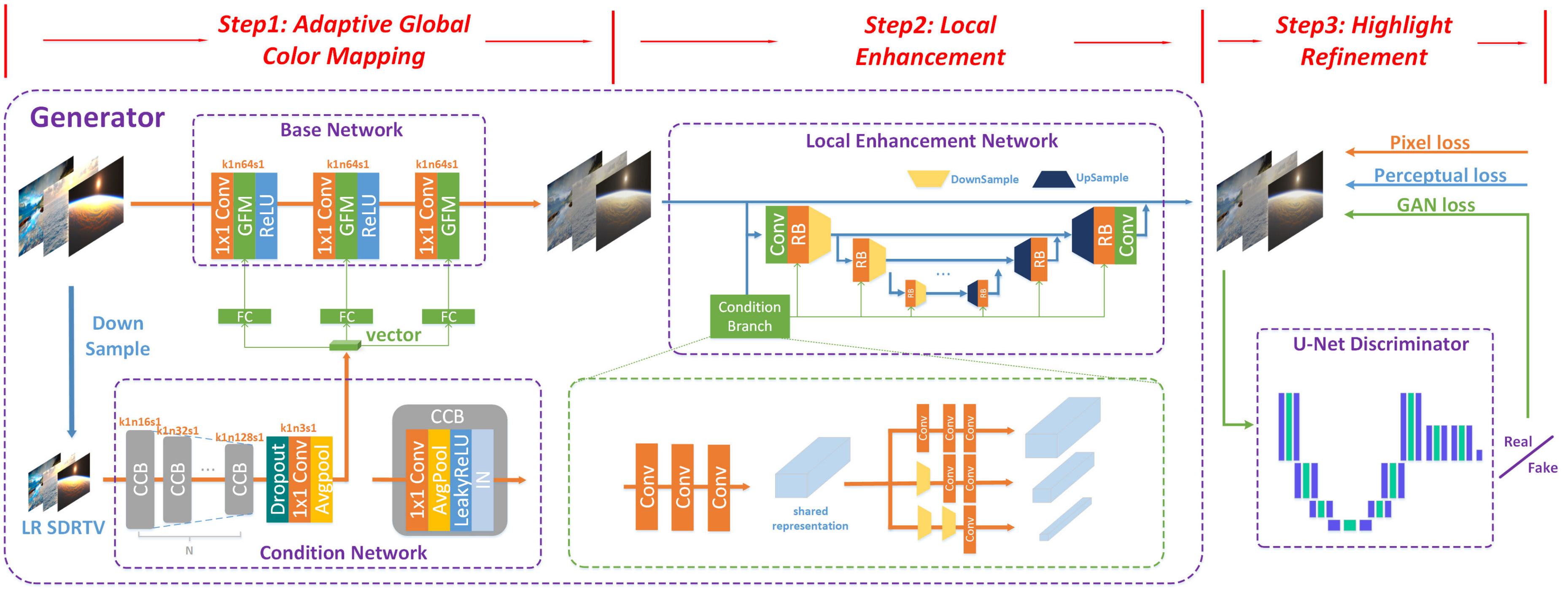}
   \end{center}
   \vspace{-8pt}
      \caption{The architecture of the proposed SDRTV-to-HDRTV method.}
   \label{Figure 2 architecture of three-step SDR-to-HDR}
   \vspace{-10pt}
\end{figure*}

\subsection{Adaptive Global Color Mapping}
Adaptive global color mapping (AGCM) aims for image-adaptive color conversion from the SDRTV domain to the HDRTV domain. We use the same network structure as the initial version. As depicted in Figure \ref{Figure 2 architecture of three-step SDR-to-HDR}, our model includes a base network and a condition network.

\subsubsection{Base Network}
The base network handles pixel-independent operations. For the input SDRTV image $I_{S}$, the mapping is denoted as:
\begin{equation}
   I_{B}(x,y)=f(I_{S}(x,y)), \forall (x,y)\in I_{S},
\end{equation}
where $I_{B}$ is the output of base network. As presented in CSRNet \cite{liu2022very}, a fully convolutional network with only $1\times 1$ convolutions and activations can achieve pixel-independent mapping. Therefore, we design the base network using $N_l$ convolutional layers with $1\times 1$ filters and $N_l$-1 ReLU activation functions, which is expressed as:
\begin{equation}
   I_{B}=Conv_{1\times 1}\circ (ReLU\circ Conv_{1\times 1})^{N_l-1}(I_{S}).
\end{equation}
The base network takes an 8-bit SDRTV image and outputs an HDRTV image encoded with 10-16 bits. Although it learns one-to-one color mapping, it performs well (see Table \ref{Table 1 comparison}). It is worth noting that this network can function like a 3D lookup table (3D LUT), but more efficiently (see Section \ref{Conversion_base_network_3DLUT}). 

\subsubsection{Condition Network}
Global priors are crucial for adaptive color mapping. To achieve image-adaptive mapping, we incorporate a condition network to modulate the base network. Prior works \cite{wang2018recovering, he2020conditional} focus on extracting spatial and local information as conditions. However, for the SDRTV-to-HDRTV problem, the global color mapping is mostly conditioned on global image statistics or color distribution, which are typically independent of spatial details. Our condition network extracts color-related information for adjustable mapping, as shown in Figure \ref{Figure 2 architecture of three-step SDR-to-HDR}. It comprises color condition blocks, convolution layers, feature dropout, and global average pooling. 

A color condition block (CCB) consists of a $1\times 1$ convolution, average pooling, LeakyReLU activation, and instance normalization \cite{dumoulin2016learned}:
\begin{equation}
   CCB(I_f)=IN\circ LReLU\circ avgpool\circ Conv_{1\times 1}(I_f),
\end{equation}
where $I_f$ is the input feature. The condition network processes a down-sampled SDRTV image $I_{S\downarrow}$ and outputs a condition vector $V$:
\begin{equation}
   V=GAP\circ Conv_{1\times 1}\circ Dropout\circ CCB^{N_c}(I_{S\downarrow}).
\end{equation}
Without local feature extraction, the overall condition network focus on deriving global priors. Dropout before the convolution and pooling layers, which acts like adding a multiplicative Bernoulli noise to features, is used to prevent overfitting. 

\subsubsection{Global Feature Modulation}
We introduce global feature modulation (GFM) to utilize global priors. GFM modulates the base network's intermediate features via scaling and shifting based on the condition vector:
\begin{equation}
   GFM(x_{i})=\alpha*x_{i}+\beta,
\end{equation}
where $x_{i}$ is the intermediate feature to be modulated, and $\alpha$, $\beta$ are scaling and shifting factors.
Overall, the AGCM network is formulated as:
\begin{equation}
    \begin{split}
   I_{AGCM}=GFM\circ Conv_{1\times 1}\circ(ReLU\circ \\GFM \circ Conv_{1\times 1})^{N_l-1}(I_{S}).
   \end{split}
\end{equation}
We optimize AGCM by minimizing the $L_1$ loss between the output and the ground-truth HDRTV image. 
In the initial version, we utilize the $L_2$ loss function to optimize the AGCM network, as it is commonly adopted in existing literature involving HDRTV conversion~\cite{kim2019deep} or color mapping~\cite{zeng2020learning} . This paper demonstrates that the $L_1$ loss function yields better results for the SDRTV-to-HDRTV problem (see Section~\ref{loss_comparison}).

\subsection{Local Enhancement}
\label{Local_Enhancement_method}
Following AGCM, Local Enhancement (LE) is crucial for SDRTV-to-HDRTV conversion. While AGCM provides significant performance, LE addresses region-dependent mappings.
Initially, a classic ResNet is used for LE \cite{hdrtvnet_conf}, but it has limited performance and large computational load. Inspired by \cite{chen2021hdrunet}, we employ a UNet structure for LE, consisting of a main and a condition branch, as illustrated in Figure \ref{Figure 2 architecture of three-step SDR-to-HDR}.
Specifically, The main branch is U-shape, and the condition branch generates vectors to modulate main branch features. The input $I_{AGCM} \in \mathbb{R}^{3\times H\times W}$ is transformed into high-dimensional features $F_0 \in \mathbb{R}^{C\times H\times W}$. A three-level encoder-decoder refines these features, using stride convolution and pixel-shuffle for downsampling and upsampling \cite{shi2016real}. Skip connections assist feature recovery.
In actual production, the resolution of SDRTV content is generally from 1K to 4K. The use of a U-shape structure can greatly reduce the computational burden required for processing.
The condition branch processes inputs through three convolutions for shared representation, generating hierarchical conditions for spatial feature modulation using SFT layers \cite{wang2018recovering}:
\begin{equation}
SFT(x_i) = m \odot x_i + n,
\end{equation}
where $\odot$ denotes the element-wise multiplication. $x_i \in\mathbb{R}^{C \times H \times W} $ is the intermediate features to be modulated. $m \in\mathbb{R}^{C \times H \times W}$ and $n \in\mathbb{R}^{C \times H \times W}$ are two condition maps predicted by the condition branch.
It is noteworthy that without AGCM, employing local enhancement with region-dependent operations can cause artifacts (see Figure \ref{visual_comparison_w_wo_AGCM}). 
To achieve better optimization, we further jointly train the AGCM and LE networks. Experiments show that joint training can still bring a slight performance improvement. Benefiting from our AGCM and LE, the proposed method can significantly outperform existing approaches with high efficiency for SDRTV-to-HDRTV. 

\subsection{Highlight Refinement}
\label{Highlight_Generation_method}

Highlight Refinement (HR) aims to address color disharmony in highlight regions, which is often caused by dynamic range and color gamut clipping. MSE-based models struggle with this ill-posed problem, so we introduce generative adversarial training to alleviate the color disharmony.
The initial version uses a separate UNet with a soft mask \cite{hdrtvnet_conf}, but it has limited visual improvement and high computational costs.
Instead, we leverage the pre-trained AGCM and LE networks as the generator, enhancing it with adversarial training to achieve better visual results, as shown in Figure \ref{Figure 2 architecture of three-step SDR-to-HDR}.
This approach brings two advantages. First, it aligns output distribution closer to the ground-truth. Since well-exposed regions have already been effectively processed in previous stages, this training will focus on improving color transition in highlight regions. Second, it avoids extra computational cost since there are no additional parameters to optimize.
The generator is defined as:
\begin{equation}
   I_{H}=Generator(I_{S})=LE(AGCM(I_{S})),
\end{equation}
where $LE(\cdot)$ and $AGCM(\cdot)$ represent the pretrained LE and AGCM networks.
We adopt the UNet-style network as the discriminator following \cite{wang2021real} and optimize the GAN training based on Relativistic GAN \cite{jolicoeur2018relativistic}. 
The overall loss function consists of $L_1$ loss, perceptual loss \cite{johnson2016perceptual} and GAN loss \cite{goodfellow2014generative,blau2018perception,ledig2017photo} as:
\begin{equation}
   L_{HR}=\lambda_{1} L_{1}+\lambda_{2} L_{Percep}+\lambda_{3} L_{GAN},
\end{equation}
where $\lambda_{1}, \lambda_{2}, \lambda_{3}$ are set to 0.01, 1 and 0.005, respectively.

\begin{figure}[t]
\centering
\includegraphics[width=1\linewidth]{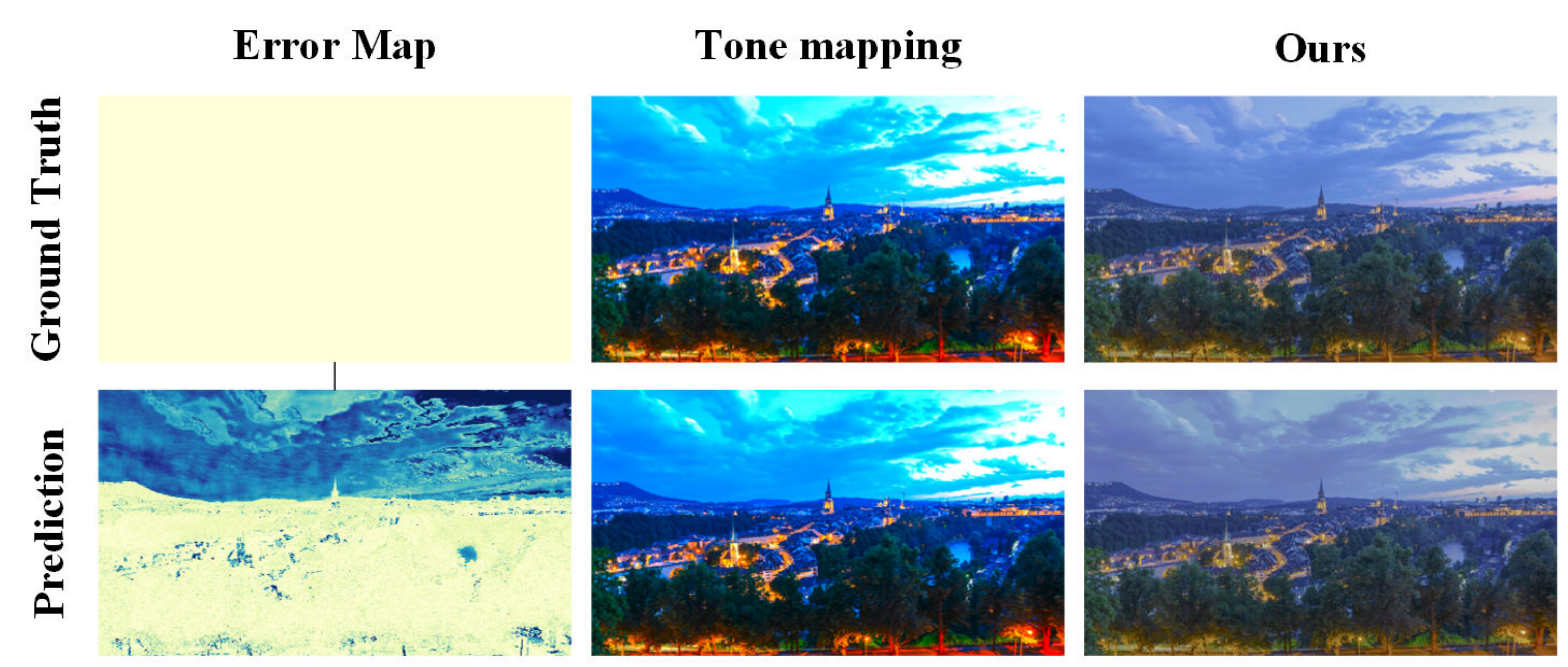}
\caption{Comparison of different visualization methods.}
\vspace{-10pt}
\label{visual_justify}
\end{figure}

\section{Experiments}
\label{Experiments}
\subsection{Experimental Setup}

\subsubsection{Dataset}
To address the scarcity of SDRTV/HDRTV data pairs for training and testing, we construct the HDRTV1K dataset. This dataset comprises 22 HDR videos (compliant with the HDR10 standard) and their SDRTV counterparts, sourced from YouTube following the methodology in \cite{kim2019deep}. All HDR videos are encoded using PQ-OETF within the Rec.2020 color gamut. We utilize 18 video pairs to create image pairs for training, reserving the remaining 4 for testing. To reduce content similarity, we sample one frame every two seconds, resulting in a training set of 1,235 images. The test set consists of 117 unique images extracted from the videos.

\begin{table*}[!t]
\begin{threeparttable}
\caption{Quantitative comparisons with existing methods.}
\label{Table 1 comparison}
\centering
\tabcolsep=4.0mm
\renewcommand{\arraystretch}{1.2}
\begin{tabular*}{\linewidth}{cc|c|ccccc}
\hline
\multicolumn{2}{c|}{Method} & Params$\downarrow$ & PSNR$\uparrow$ & SSIM$\uparrow$ & SR-SIM$\uparrow$ & $\Delta E_{ITP}\downarrow$ & HDR-VDP3$\uparrow$ \\ \hline
 & HuoPhyEO \cite{huo2014physiological} & - & 25.90 & 0.9296 & 0.9881 & 38.06 & 7.893 \\ 
\multirow{-2}{*}{LDR-to-HDR} & KovaleskiEO \cite{kovaleski2014high} & - & 27.89 & 0.9273 & 0.9809 & 28.00 & 7.431 \\ \hline
& ResNet \cite{he2016identity} & 1.37M & 37.32 & 0.9720 & 0.9950 & 9.02 & 8.391 \\  
 & Pixel2Pixel \cite{isola2017image} & 11.38M & 25.80 & 0.8777 & 0.9871 & 44.25 & 7.136 \\  
\multirow{-3}{*}{\begin{tabular}[c]{@{}c@{}}image-to-image\\ translation\end{tabular}} & CycleGAN \cite{zhu2017unpaired} & 11.38M & 21.33 & 0.8496 & 0.9595 & 77.74 & 6.941 \\ 
\hline
 & HDRNet \cite{gharbi2017deep} & 482K & 35.73 & 0.9664 & 0.9957 & 11.52 & 8.462 \\ 
 & CSRNet \cite{he2020conditional} & 36K & 35.04 & 0.9625 & 0.9955 & 14.28 & 8.400 \\  
\multirow{-3}{*}{\begin{tabular}[c]{@{}c@{}}photo\\ retouching\end{tabular}} & Ada-3DLUT \cite{zeng2020learning} & 594K & 36.22 & 0.9658 & 0.9967 & 10.89 & 8.423 \\ 
\hline
 & Deep SR-ITM \cite{kim2019deep} & 2.87M & 37.10 & 0.9686 & 0.9950 & 9.24 & 8.233 \\  
 & JSI-GAN \cite{kim2020jsi} & 1.06M & 37.01 & 0.9694 & 0.9928 & 9.36 & 8.169 \\
& FMNet \cite{xu2022fmnet}& 1.24M & 37.94 & \textbf{0.9747} & 0.9957 & 8.10 & 8.510 \\
 \multirow{-4}{*}{\begin{tabular}[c]{@{}c@{}}SDRTV-to-HDRTV\end{tabular}} & HDRTVDM \cite{guo2023learning}& 325k & 37.98 & 0.9707 & \underline{0.9974} & 8.84 & 8.610 \\
\hline
  & Base Network & 5K & 36.14 & 0.9643 & 0.9961 & 10.43 & 8.305 \\  
 & AGCM & 35K & 36.88 & 0.9655 & 0.9964 & 9.78 & 8.464 \\  
 & AGCM-LE & 1.41M & 37.61 & 0.9726 & 0.9967 & 8.89 & 8.613 \\
 \multirow{-4}{*}{\begin{tabular}[c]{@{}c@{}}\textbf{HDRTVNet\cite{hdrtvnet_conf}}\end{tabular}}& AGCM-LE-HG & 37.20M & 37.21 & 0.9699 & 0.9968 & 9.11 & 8.569 \\ 
 \hline
 & AGCM++ & 35K & 37.35 & 0.9666 & 0.9968 & 9.29 & 8.511 \\  
 & AGCM-LE++ & 591K & \underline{38.45} & 0.9739 & 0.9970 & \underline{7.90} & 8.666 \\ 
 & AGCM-LE++$^\dagger$ &591K & \textbf{38.60} & \underline{0.9745} & 0.9973 & \textbf{7.67} & \underline{8.696} \\
 \multirow{-4}{*}{\begin{tabular}[c]{@{}c@{}}\textbf{HDRTVNet++}\\ (ours)\end{tabular}} & AGCM-LE-HR++ &591K & 38.36 & 0.9735 & \textbf{0.9975} & 8.28  & \textbf{8.751} \\ 
 \hline
\end{tabular*}%
\begin{tablenotes}
\item[1] The best and second-best performance results are in \textbf{bold} and \underline{underline}.
\item[2] $^\dagger$ means the model is finetuned by joint training.
\end{tablenotes}
\end{threeparttable}
\vspace{-10pt}
\end{table*}

\subsubsection{Training details} 
For the proposed AGCM, the base network includes three convolutional layers with a $1 \times 1$ kernel size and 64 channels, while the condition network comprises four CCBs. Images are cropped to $480 \times 480$ with a step of 240 prior to training. During training, patches of size $480 \times 480$ are input into the base network, while full images downscaled by a factor of 4 are input into the condition network. We use a mini-batch size of 4 and employ the $L_1$ loss function and Adam optimizer for $1 \times 10^6$ iterations. The initial learning rate is $4 \times 10^{-4}$, decaying by a factor of 2 at $5 \times 10^5$ and $8 \times 10^5$ iterations.
For LE, the AGCM outputs serve as inputs. The mini-batch size is set to 8 with a patch size of $240 \times 240$. The initial learning rate is $1 \times 10^{-4}$, decaying by a factor of 2 every $2 \times 10^5$ iterations, across $1 \times 10^6$ iterations. The $L_1$ loss function and Adam optimizer are used for training.
In joint training, the AGCM and LE networks are optimized simultaneously with the $L_1$ loss function and Adam optimizer, using a batch size of 4 and a patch size of 192. The initial learning rate is $1 \times 10^{-4}$, decaying by 2 every $1 \times 10^5$ iterations, over $5 \times 10^5$ iterations.
Subsequently, the GAN model is trained with a batch size of 64 and a patch size of 128. The initial learning rate is $1 \times 10^{-4}$, with a total of $4 \times 10^5$ iterations. The learning rate decays by a factor of 0.5 at $5 \times 10^4$, $1 \times 10^5$, $2 \times 10^5$, and $3 \times 10^5$ iterations. All models are implemented using PyTorch and trained on NVIDIA 3090 GPUs. 

\subsubsection{Evaluation} 

We utilize five metrics for comprehensive evaluation: PSNR, SSIM, SR-SIM \cite{zhang2012sr}, HDR-VDP3 \cite{mantiuk2011hdr}, and $\Delta E_{ITP}$ \cite{ITP}. PSNR assesses SDRTV-to-HDRTV fidelity against ground truth HDRTV images. SSIM and SR-SIM are adopted to evaluate image structural similarity; SR-SIM, although designed for SDR images, is effective for HDR standards as shown in \cite{athar2019perceptual}. $\Delta E_{ITP}$ measures color differences, tailored for HDRTV content. HDR-VDP3 is an improved version of HDR-VDP2, which supports the Rec.2020 color gamut. For HDR-VDP3, evaluations are performed by setting the side-by-side'' task, rgb-bt.2020'' color encoding, 50 pixels per degree, and led-lcd-wcg'' for the rgb-display'' option.

HDRTV images are displayed in 16-bit PNG format without additional processing. Due to gamma EOTF decoding on SDR screens, they may appear darker than on HDR screens, yet visual differences remain discernible. 
Previous work \cite{kim2019deep, kim2020jsi} visualize HDRTV images using video players (i.e., MPC-HC player). However, this comparison may be unfair, because the software introduces unknown enhancement, particularly in highlight regions, leading to similar visual outcomes. 
Another approach is to use error maps to show the intensity difference between the generated result and the corresponding ground truth, while it may fail to reflect visual differences accurately.
In contrast, our method preserves highlight details and aligns closely with human perception, as demonstrated in Figure \ref{visual_justify}.

\subsection{Comparison with Existing Methods}
\label{comparison_with_other_methods}

We compare our results with four types of methods: SDRTV-to-HDRTV, image-to-image translation, photo retouching, and LDR-to-HDR. Since these methods are not all specifically designed for this task, necessary adjustments are made. 
For joint SR with SDRTV-to-HDRTV methods, we modify the stride of the first convolutional layer to 2 for downsampling to match input and output sizes.\footnote{We have also conducted experiments with removing the upsampling operation at the end of the networks, but this do not improve performance and significantly increased memory and runtime costs.}
For LDR-to-HDR methods, we process results as illustrated in Figure \ref{LDR2HDR_based_solution}, following the same steps as previous works~\cite{kim2019deep, kim2020jsi}. Note that All data-driven methods are retrained on our dataset. 

\textbf{Quantitative comparison.} As shown in Table \ref{Table 1 comparison}, our method significantly outperforms other methods on all metrics. Notably, our initial AGCM version achieves comparable performance to Ada-3DLUT with only 1/17 of its parameters.
By further optimizing the training of AGCM, the improved version, AGCM++, surpasses all compared methods including recent works FMNet~\cite{xu2022fmnet} and HDRTVDM~\cite{guo2023learning}. 
When equipped with the LE network and joint training, our approach achieves 38.60dB, surpassing all other approaches, including a 0.66dB gain over FMNet, a 0.62dB gain over HDRTVDM and about 1dB over the initial version HDRTVNet.
For the HR part, although generative adversarial training reduces PSNR performance, it achieves the best HDR-VDP3 perceptual quality scores.
All the quantitative results show the superiority of our method, and it is noteworthy that HDRTVNet++ is efficient and has much fewer parameters than other methods. 

\begin{figure*}[!t]
	\scriptsize
	\centering
	\setlength{\tabcolsep}{0.05cm}
	\begin{tabular}{ccccccc}
		\includegraphics[width=0.99in]{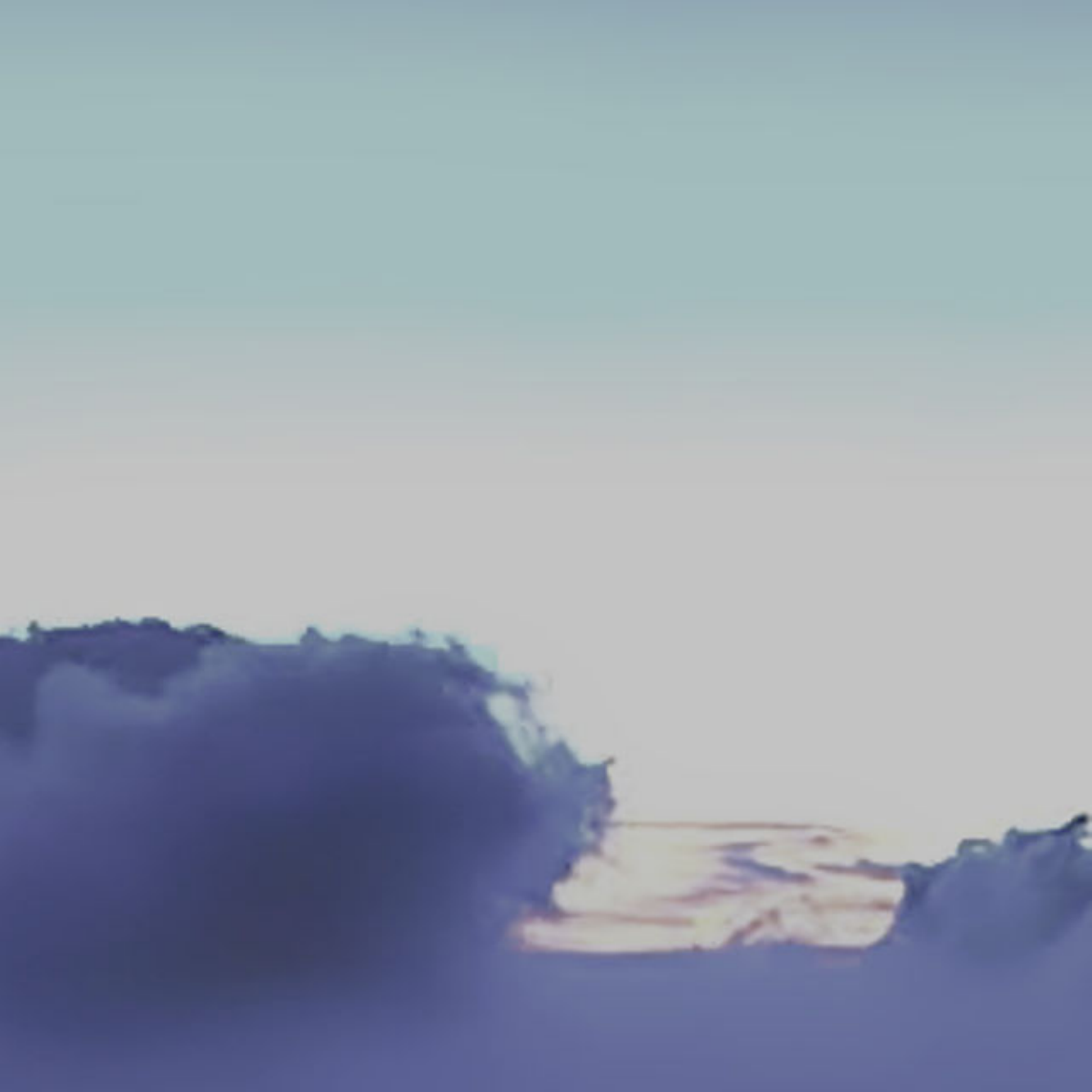}
		&\includegraphics[width=0.99in]{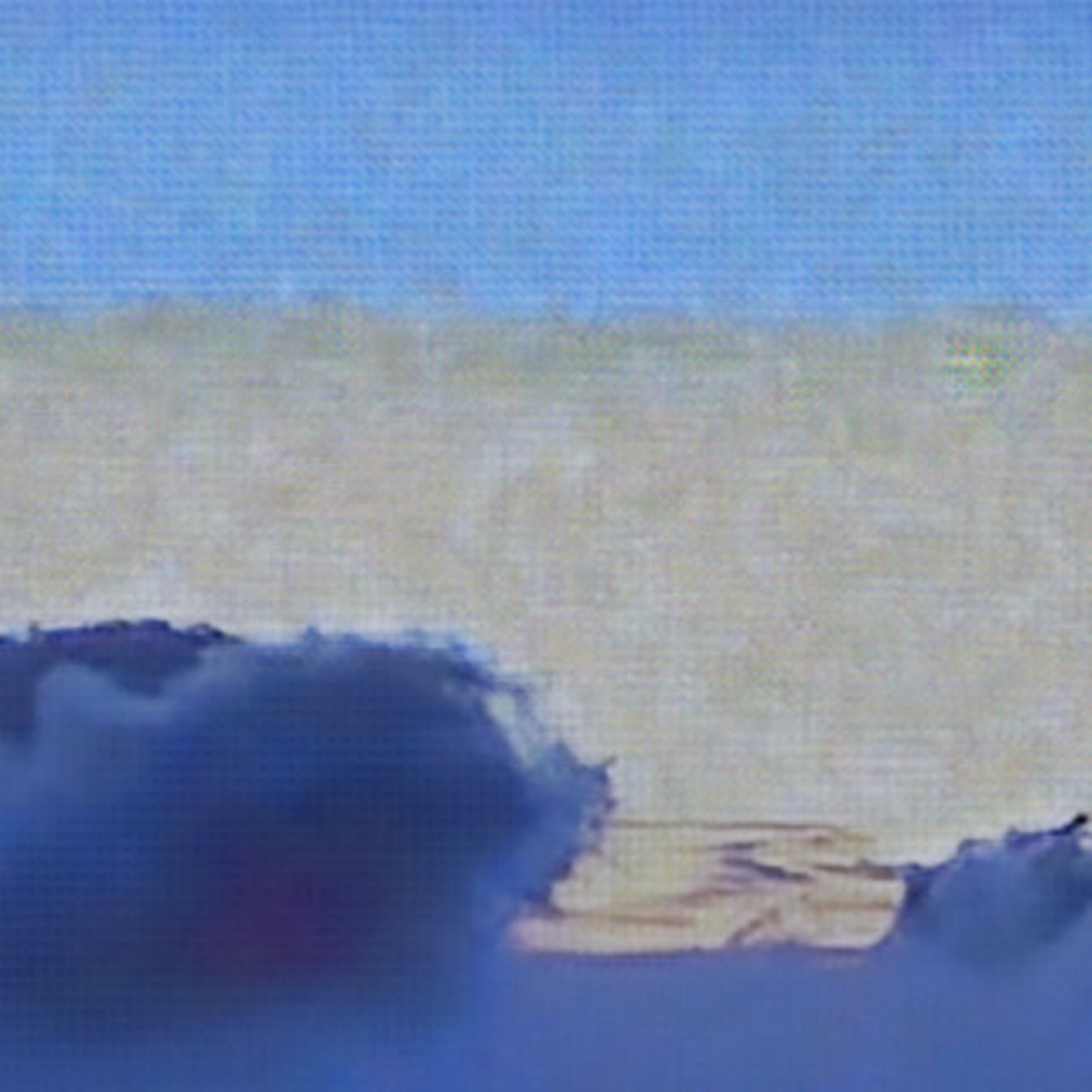}
		&\includegraphics[width=0.99in]{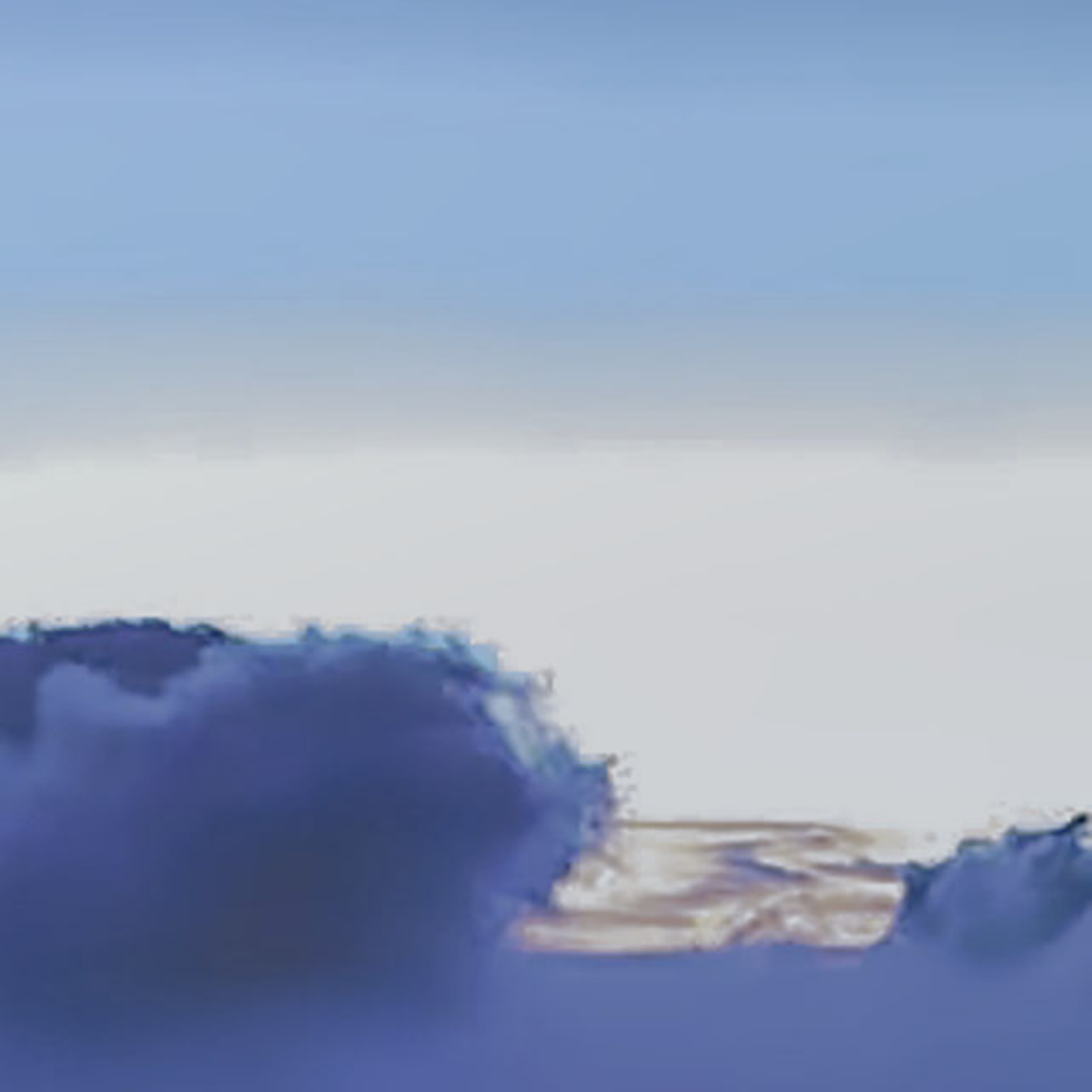}
		&\includegraphics[width=0.99in]{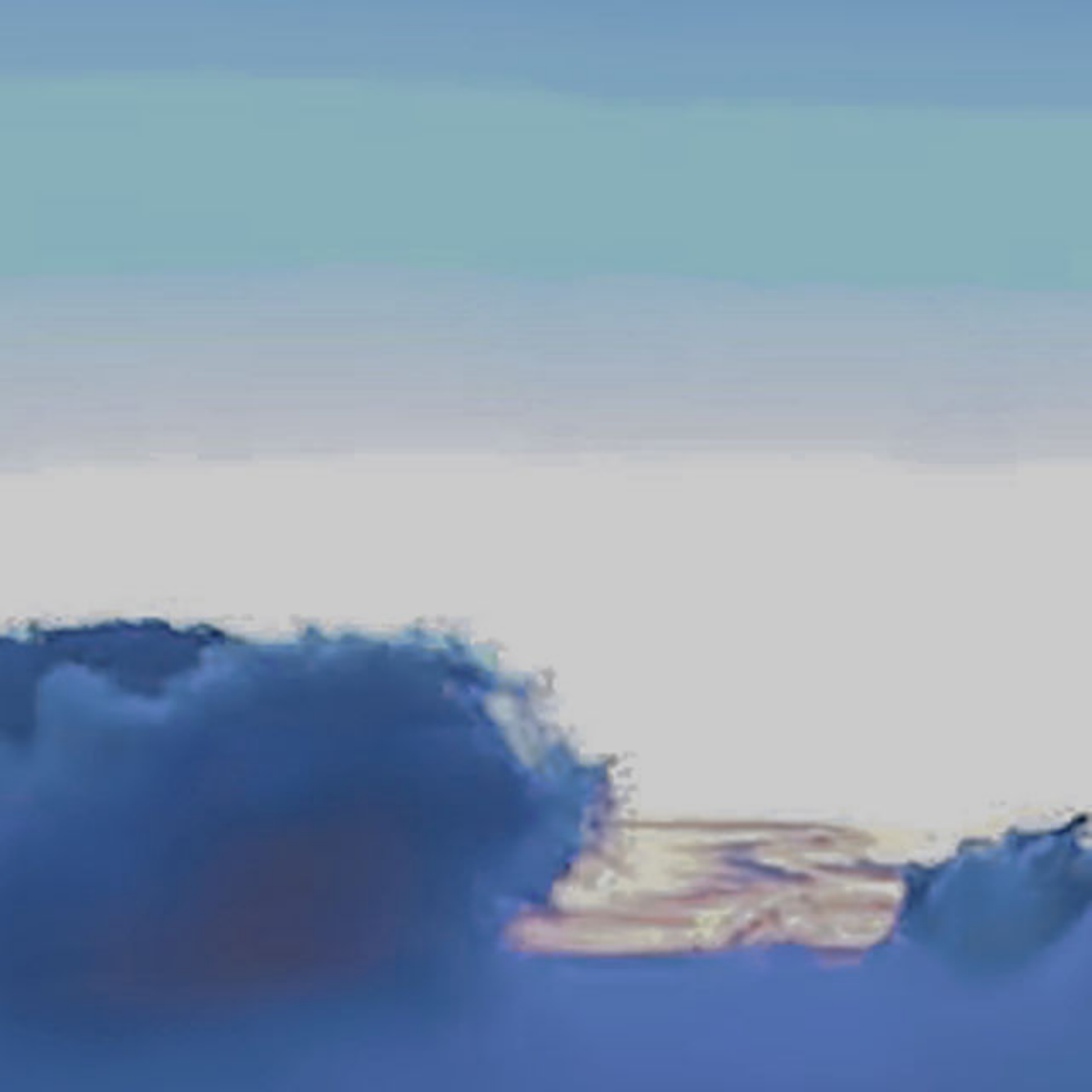}
		&\includegraphics[width=0.99in]{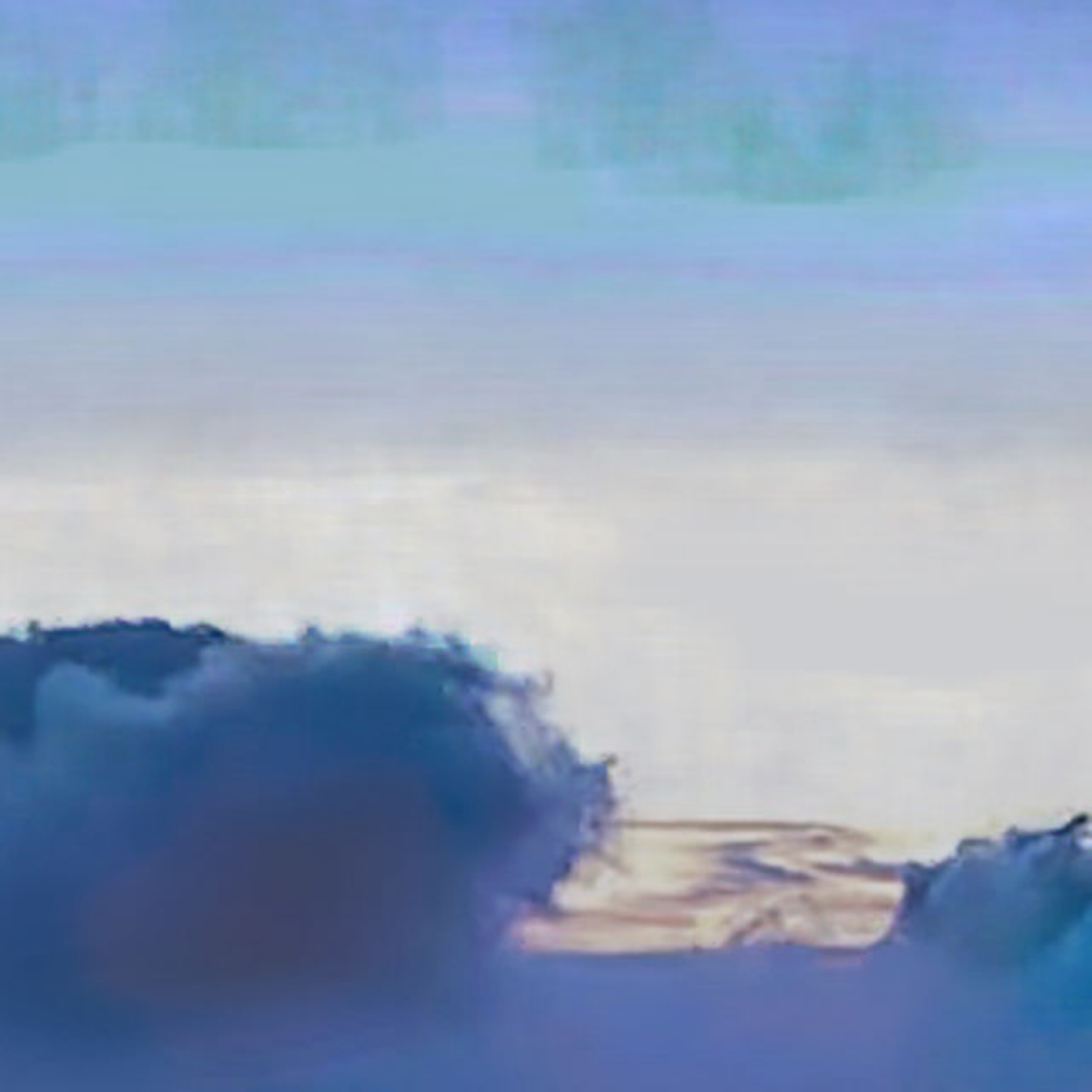}
		&\includegraphics[width=0.99in]{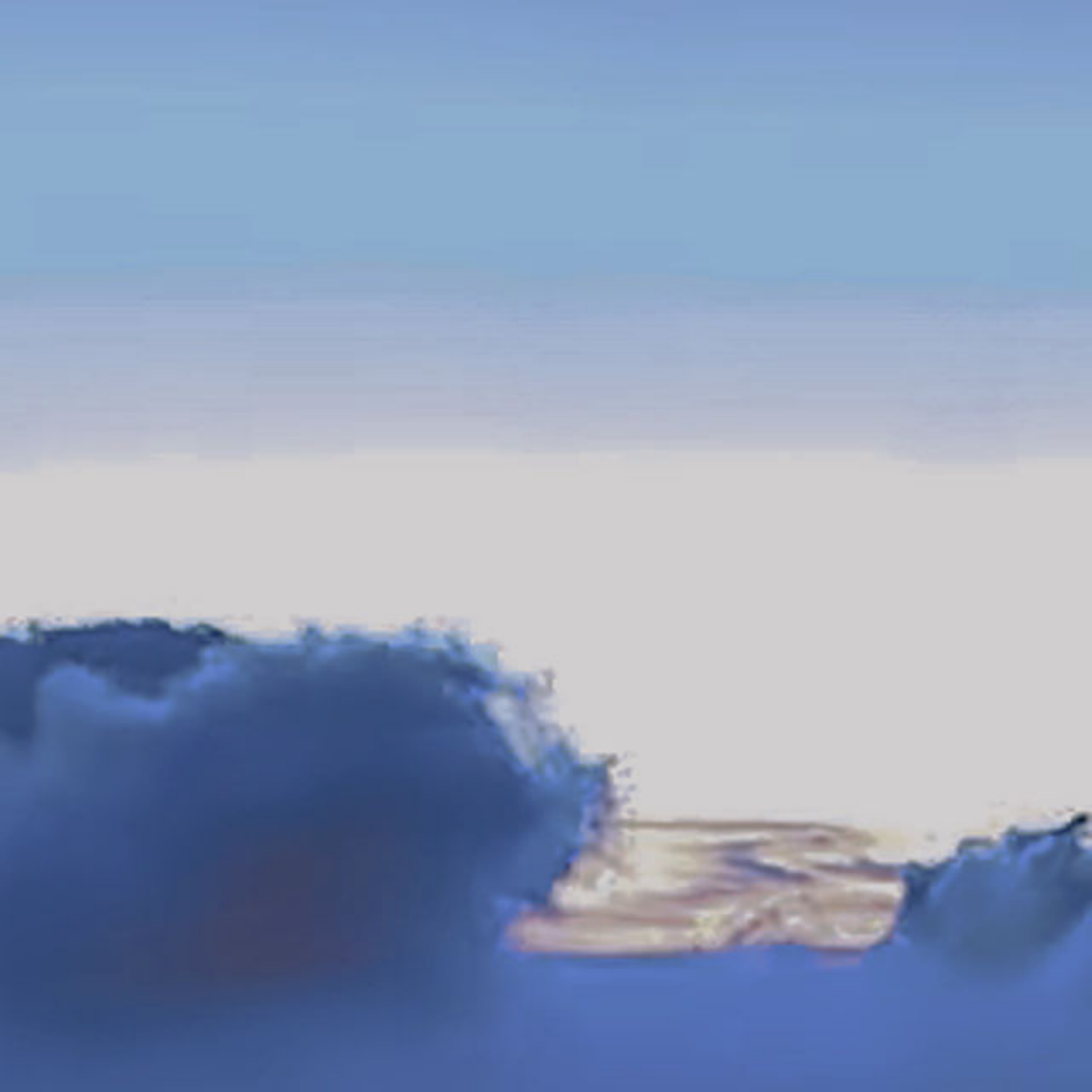}
		&\includegraphics[width=0.99in]{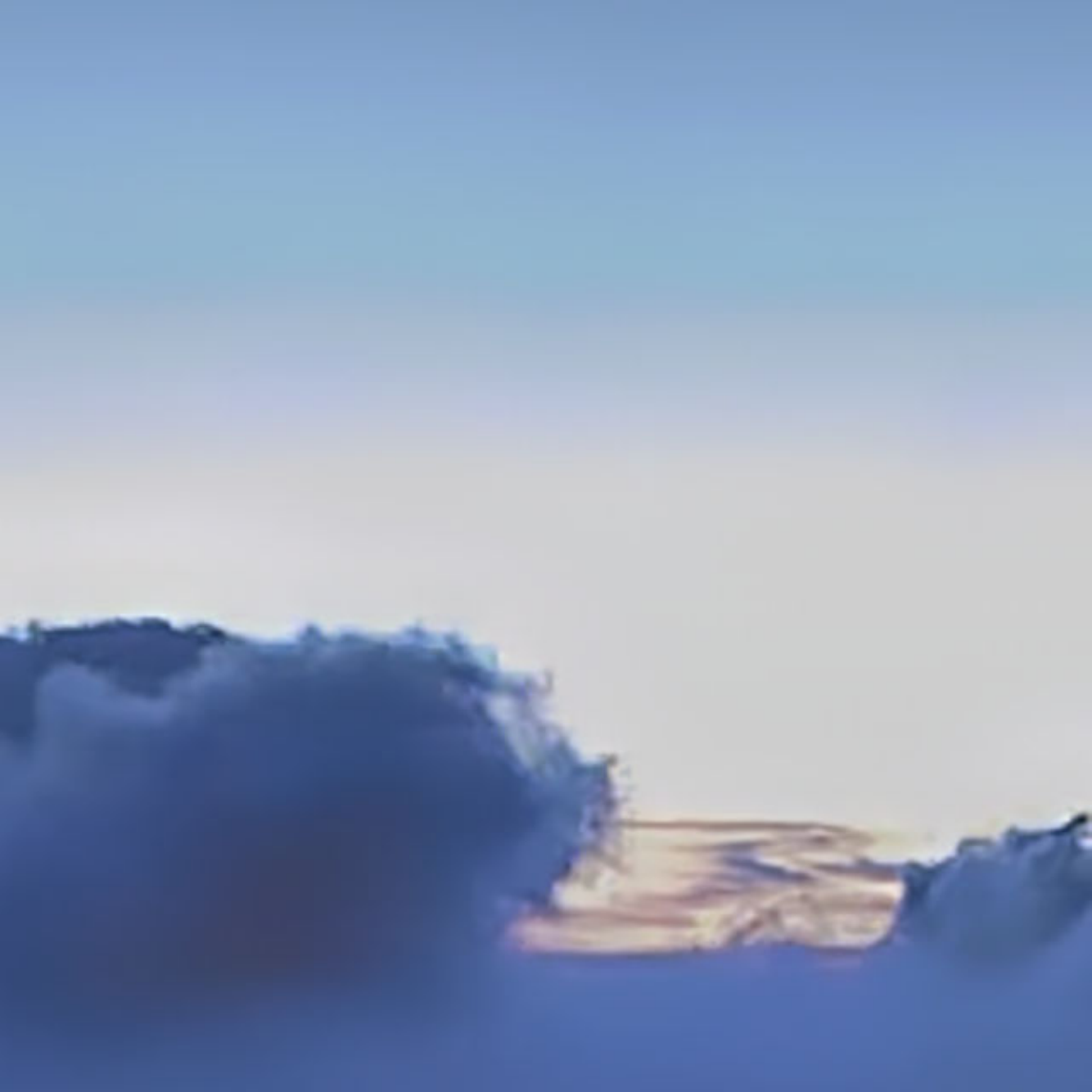}
		\\
		HuoPhyEO \cite{huo2014physiological}& Pixel2Pixel\cite{isola2017image}&HDRNet\cite{gharbi2017deep}& Ada-3DLUT\cite{zeng2020learning}&Deep SR-ITM\cite{kim2019deep}&AGCM++& AGCM-LE-HR++
		\\ 
  		\includegraphics[width=0.99in]{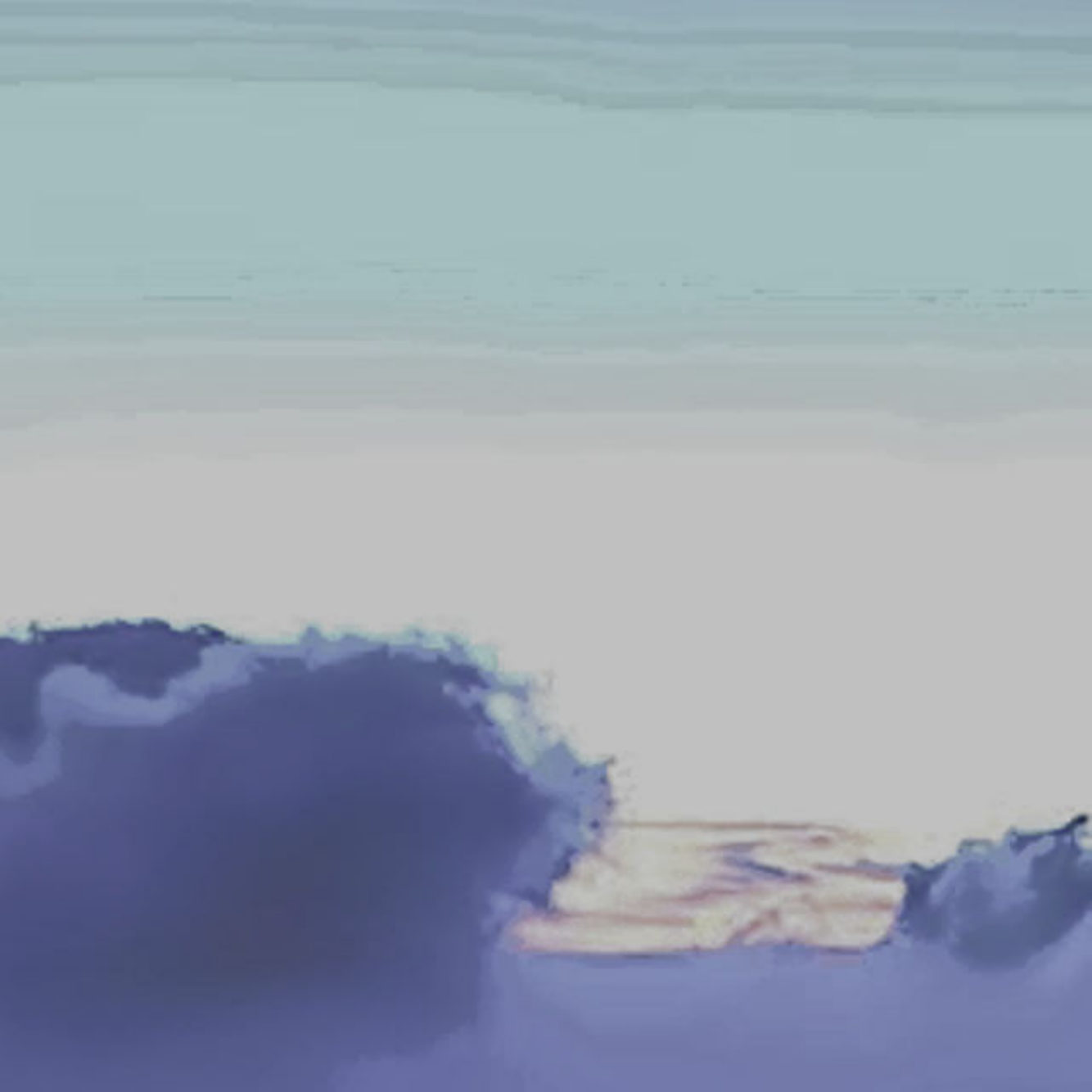}
		&\includegraphics[width=0.99in]{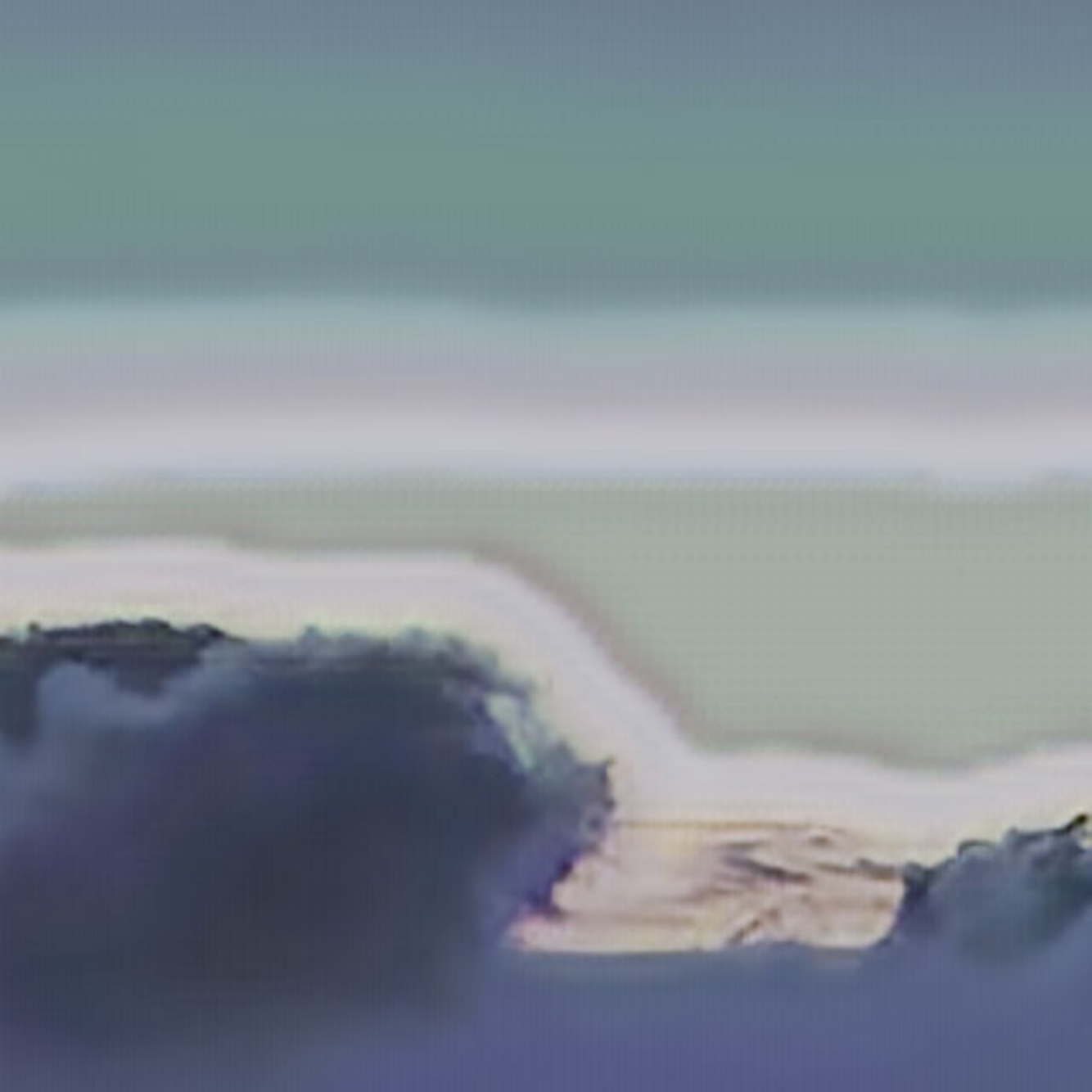}
		&\includegraphics[width=0.99in]{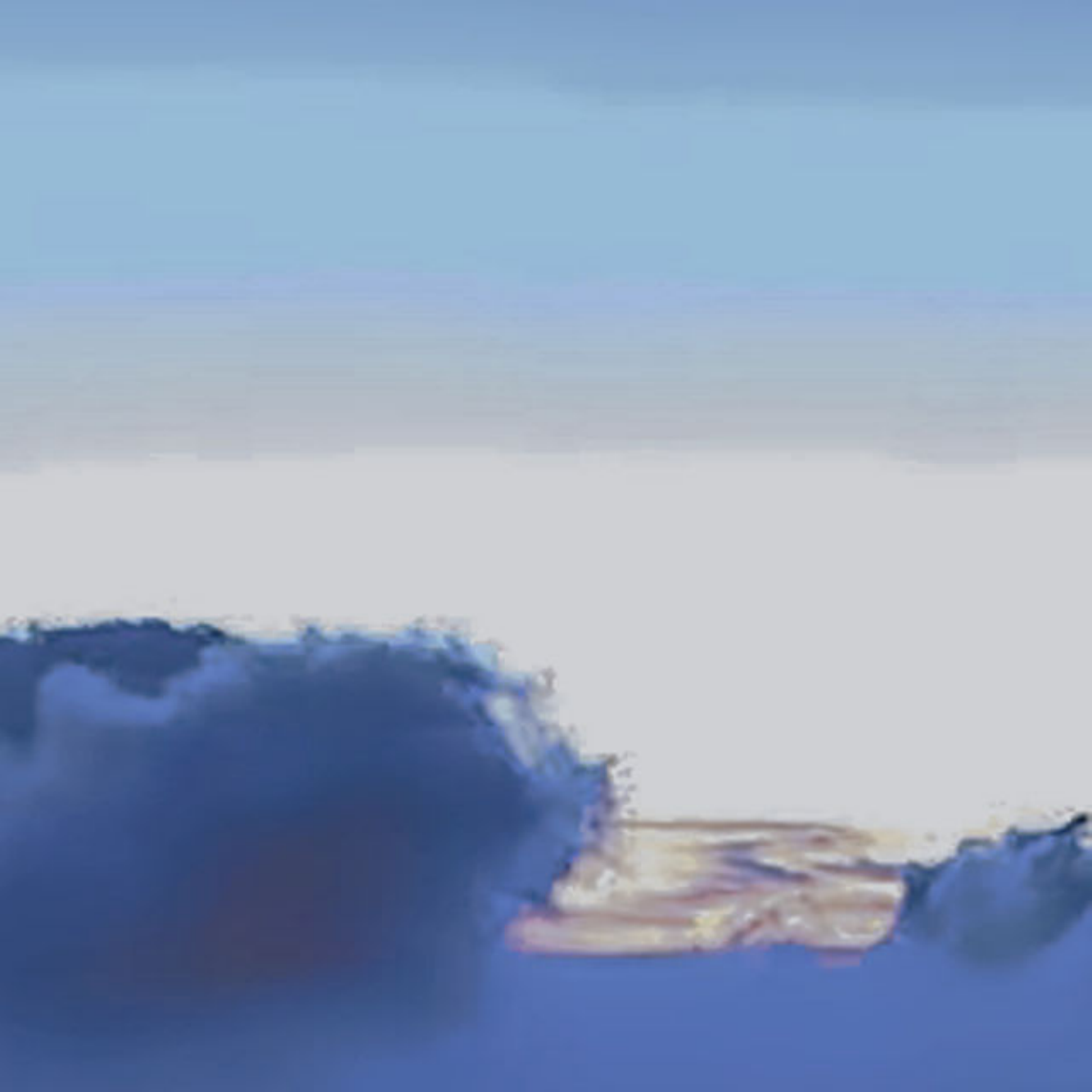}
		&\includegraphics[width=0.99in]{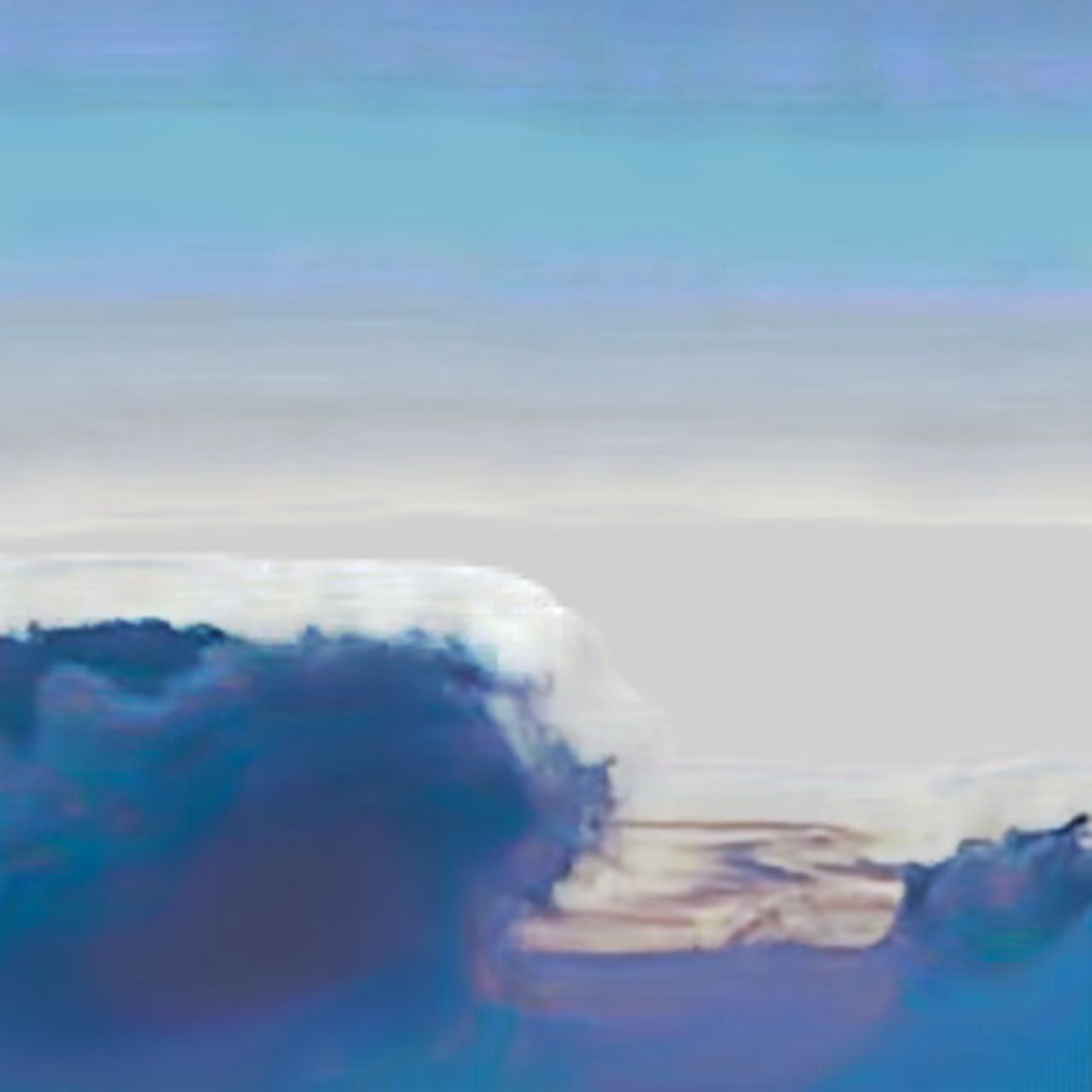}
		&\includegraphics[width=0.99in]{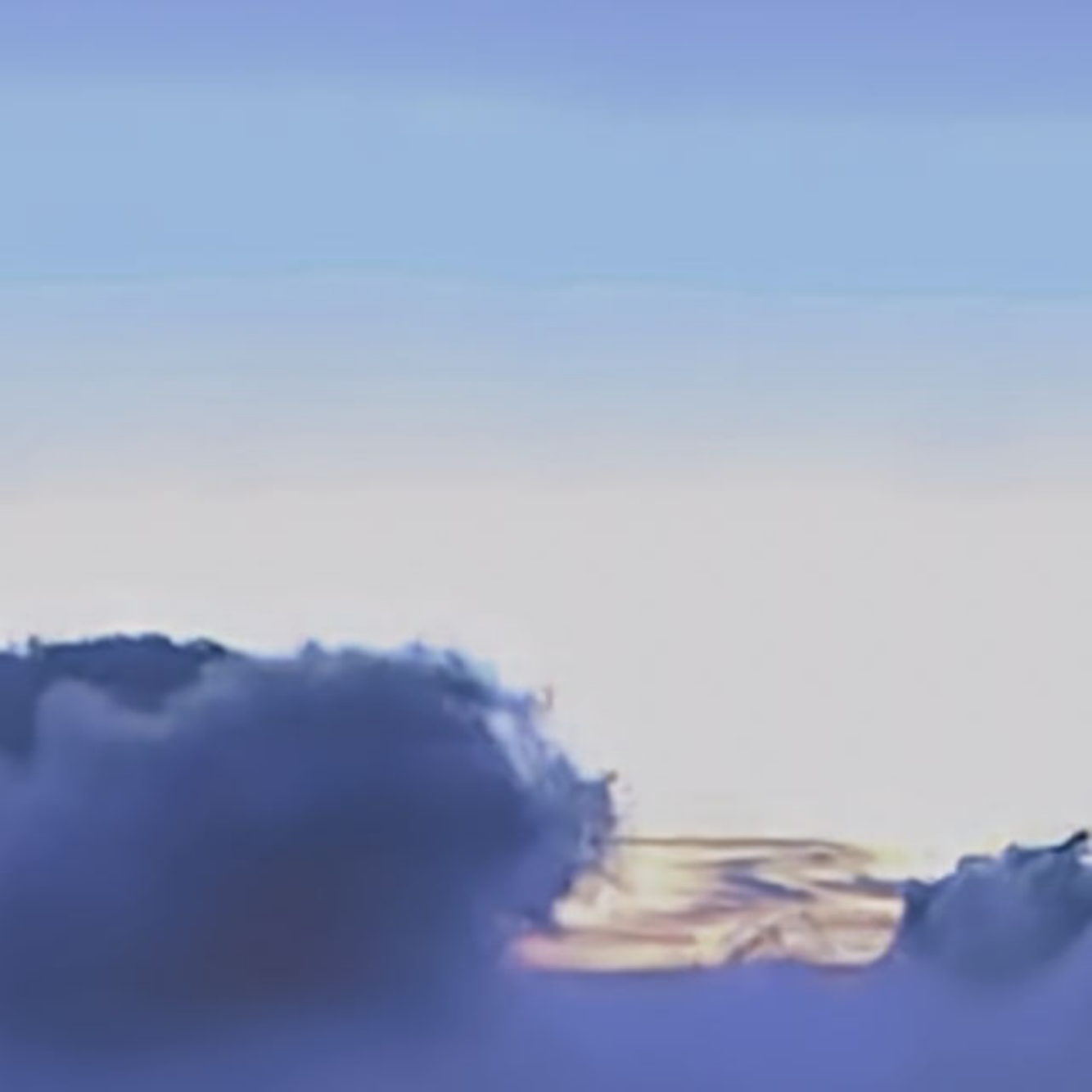}
		&\includegraphics[width=0.99in]{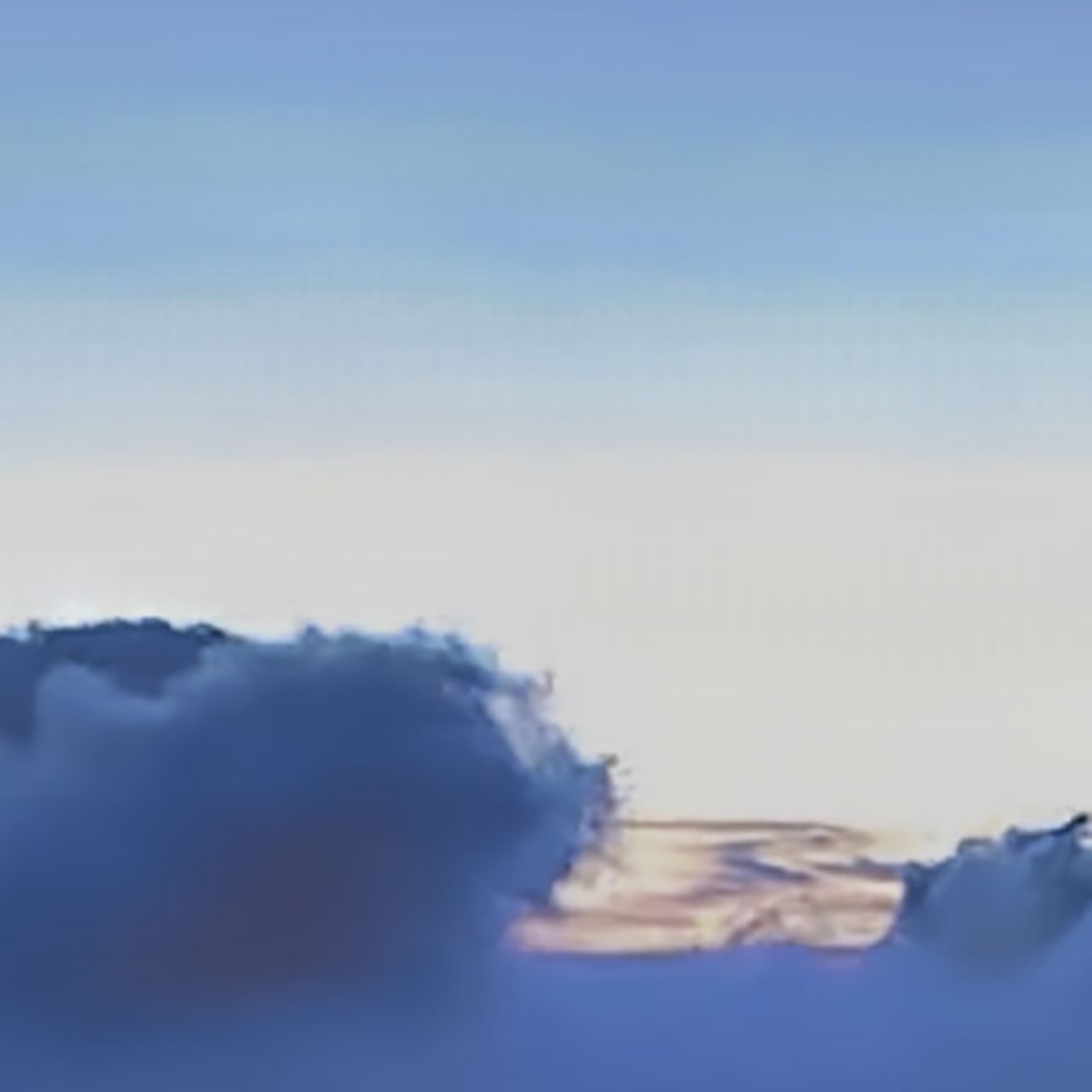}
		&\includegraphics[width=0.99in]{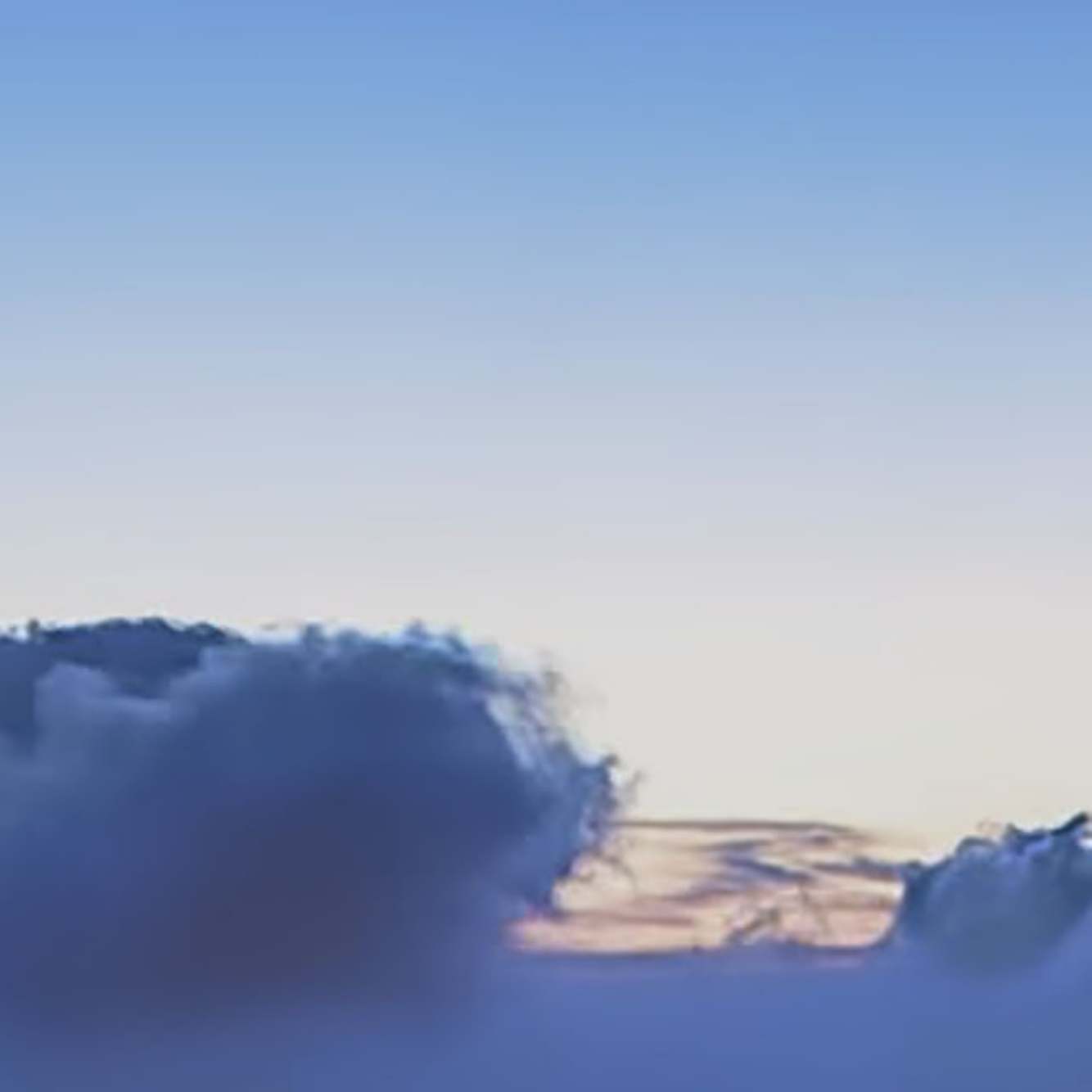}
		\\
		KovaleskiEO\cite{kovaleski2014high}& CycleGAN\cite{zhu2017unpaired}&CSRNet\cite{he2020conditional}& JSI-GAN\cite{kim2020jsi}&HDRTVNet\cite{hdrtvnet_conf}&AGCM-LE++& GT
		\\ 
  		\includegraphics[width=0.99in]{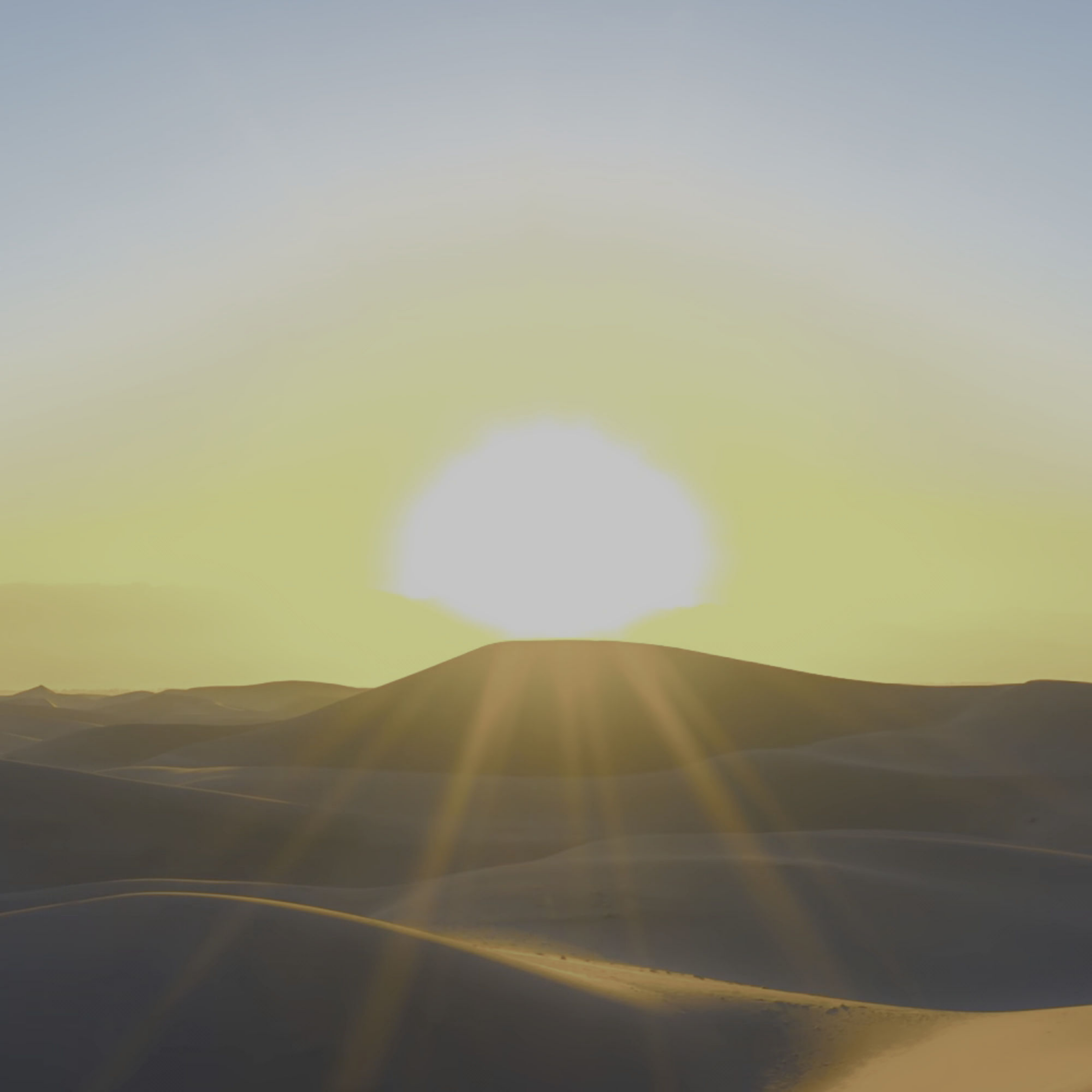}
		&\includegraphics[width=0.99in]{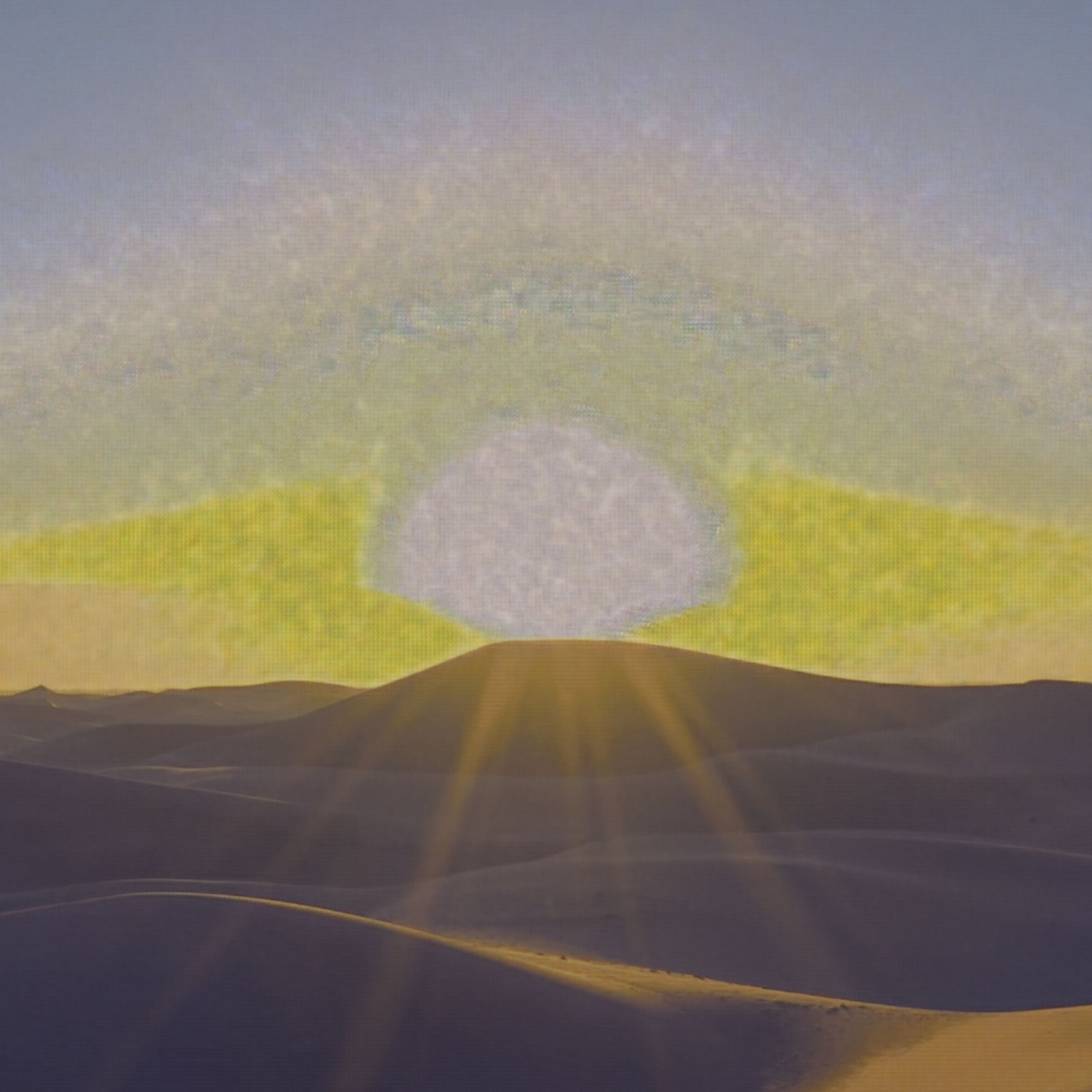}
		&\includegraphics[width=0.99in]{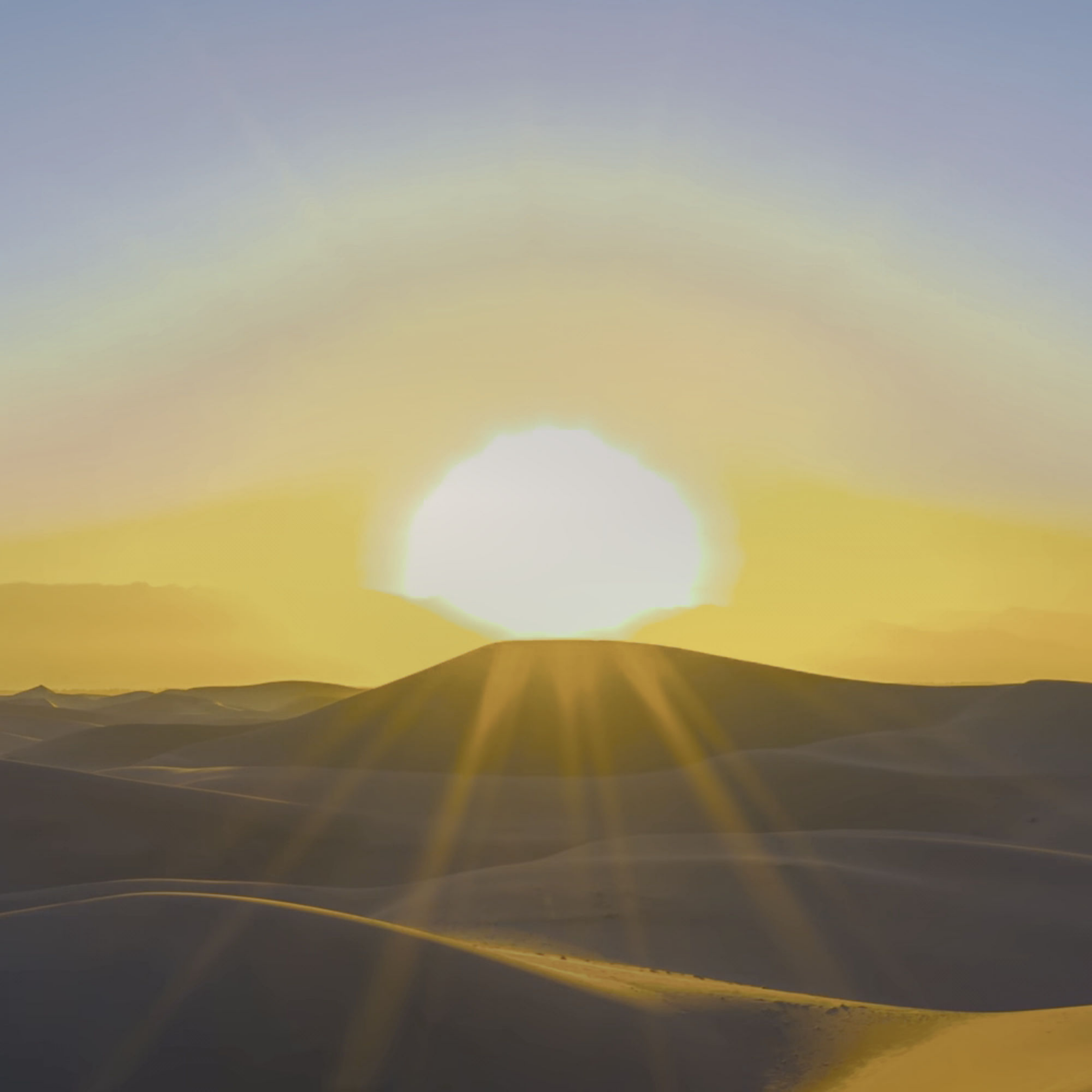}
		&\includegraphics[width=0.99in]{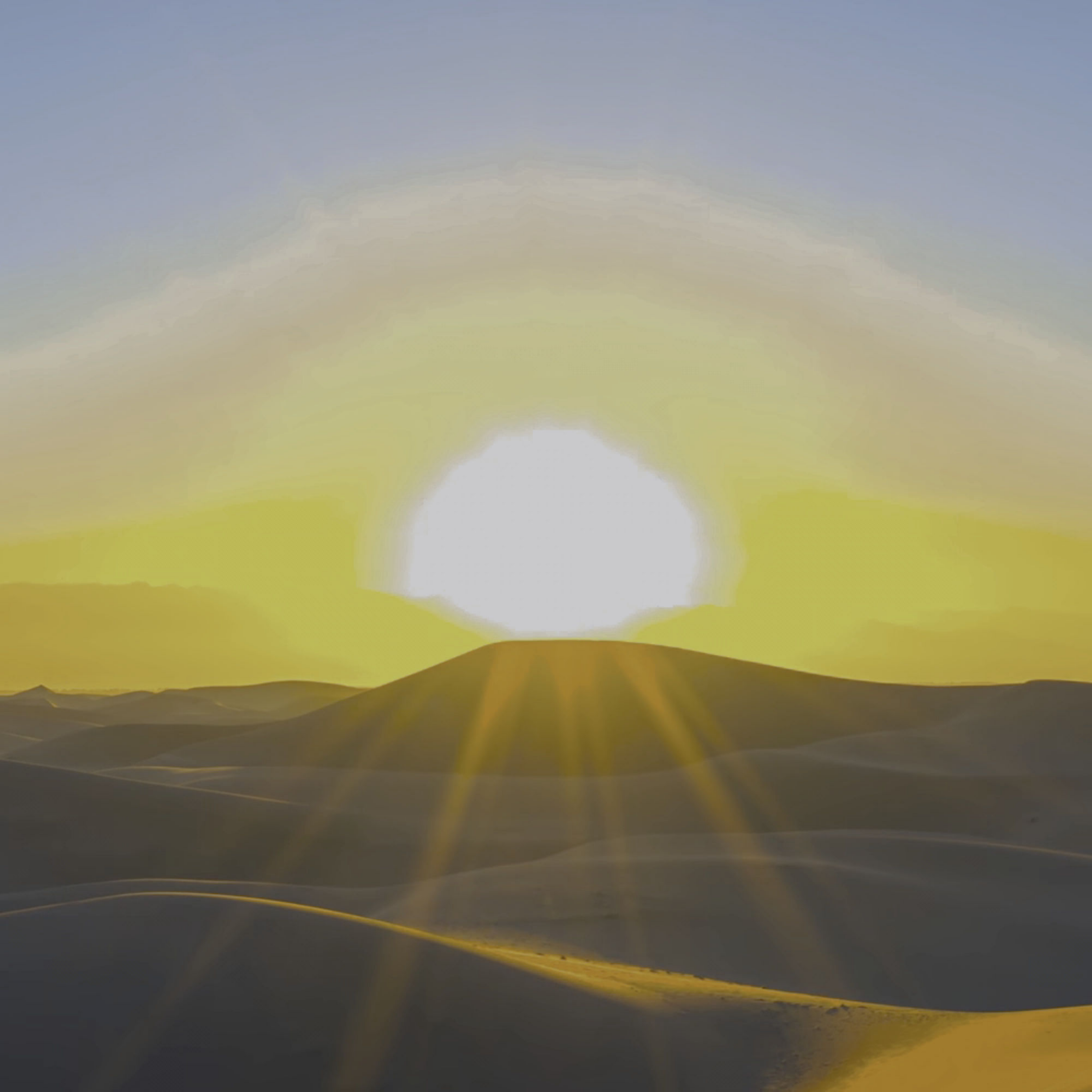}
		&\includegraphics[width=0.99in]{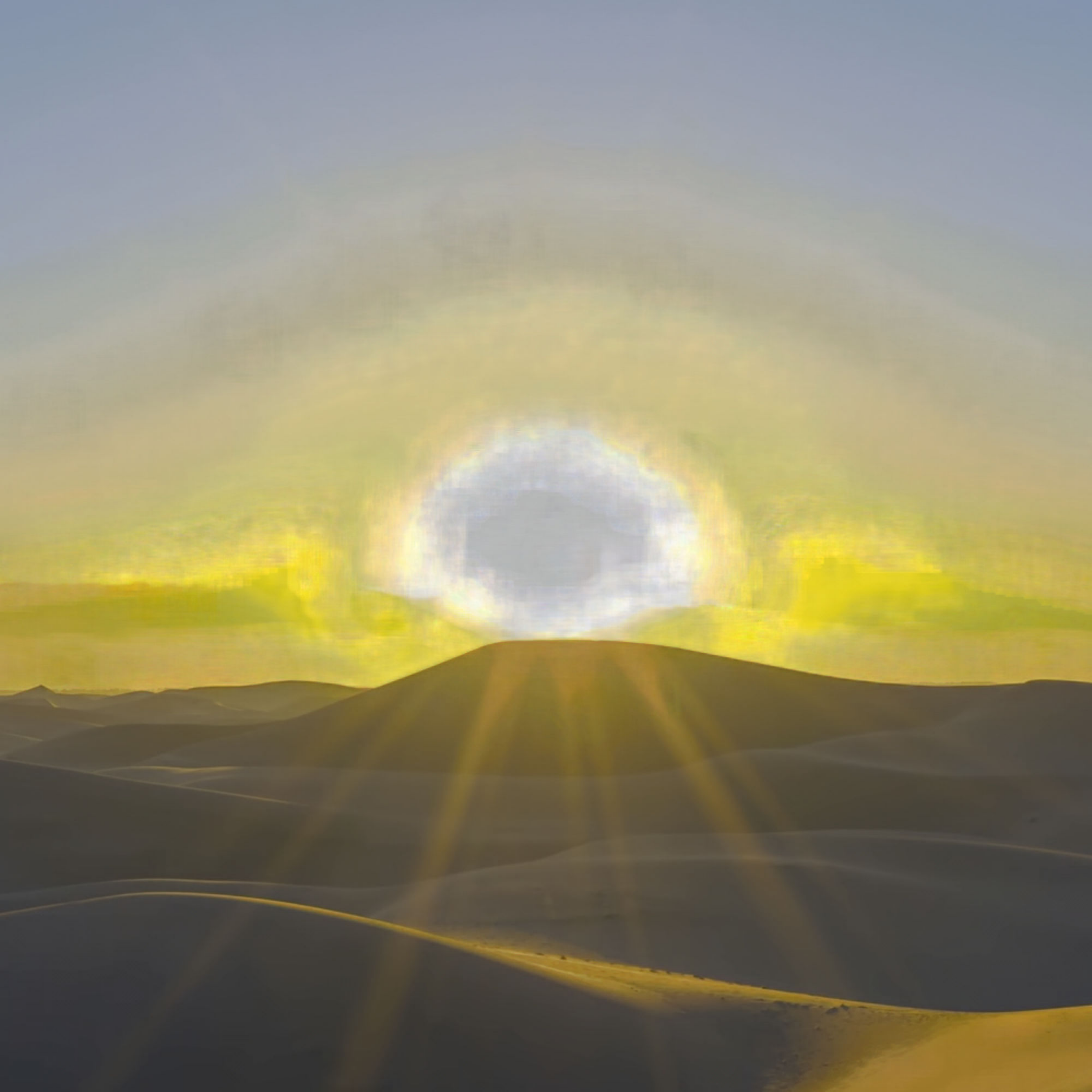}
		&\includegraphics[width=0.99in]{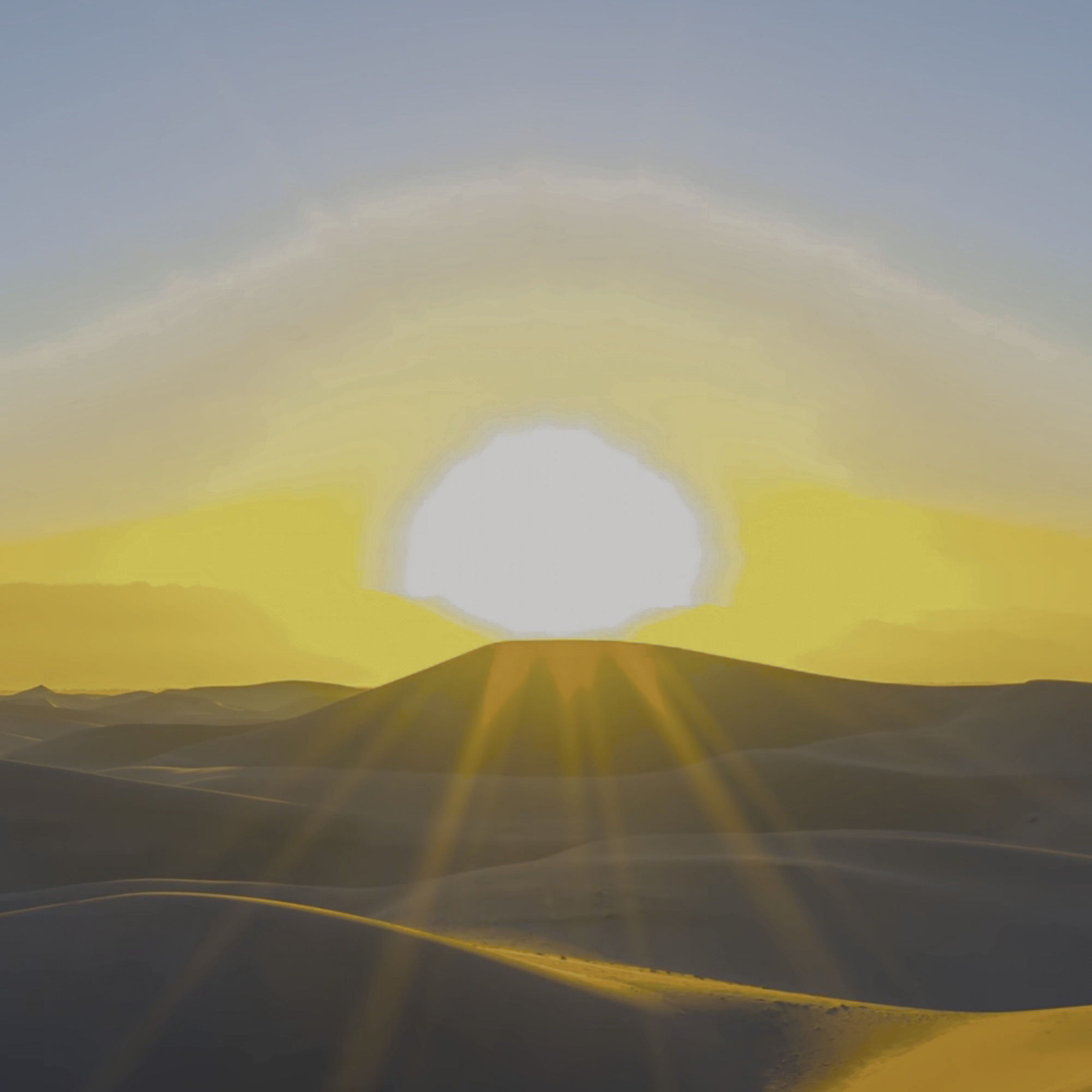}
		&\includegraphics[width=0.99in]{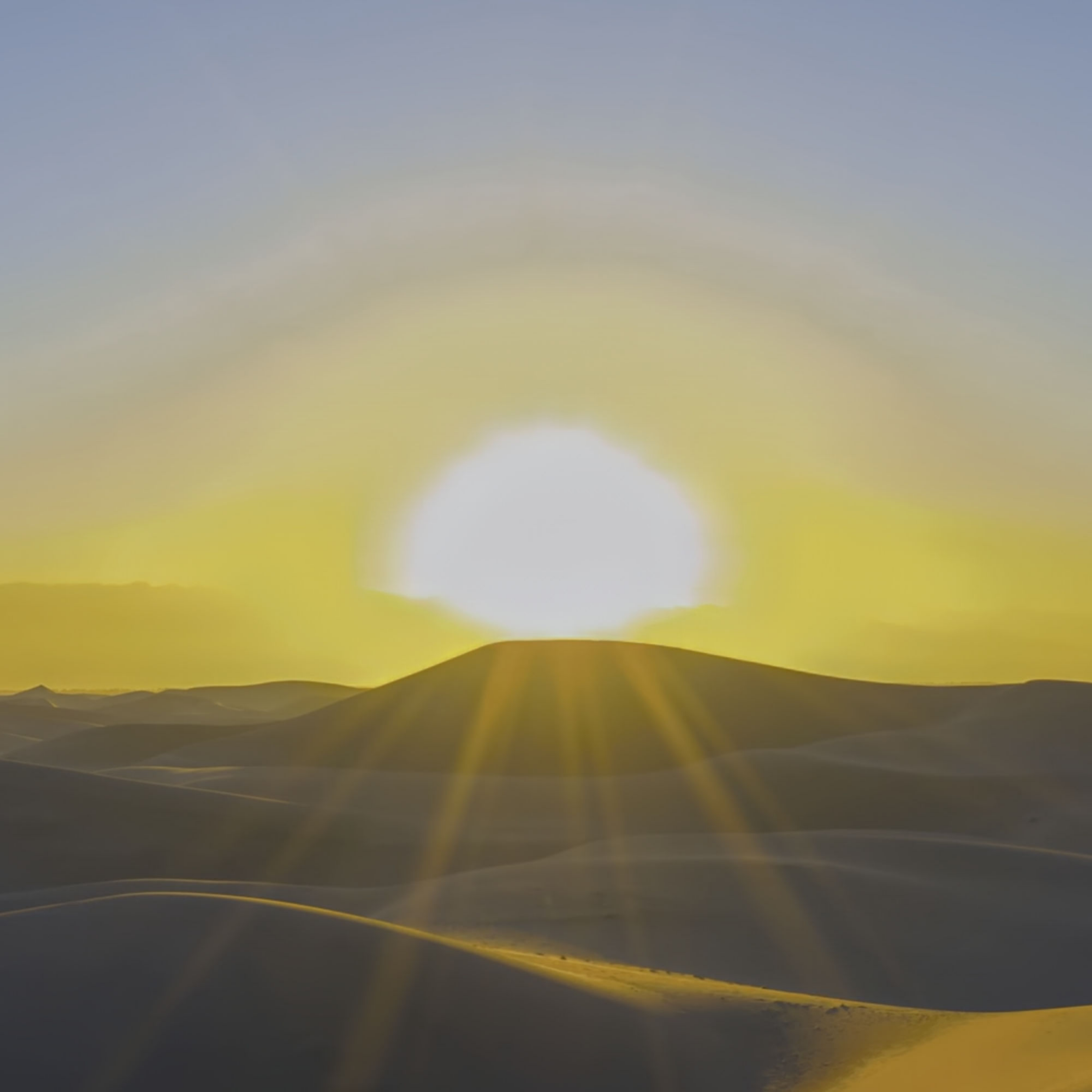}
		\\
		HuoPhyEO \cite{huo2014physiological}& Pixel2Pixel\cite{isola2017image}&HDRNet\cite{gharbi2017deep}& Ada-3DLUT\cite{zeng2020learning}&Deep SR-ITM\cite{kim2019deep}&AGCM++& AGCM-LE-HR++
		\\ 
  		\includegraphics[width=0.99in]{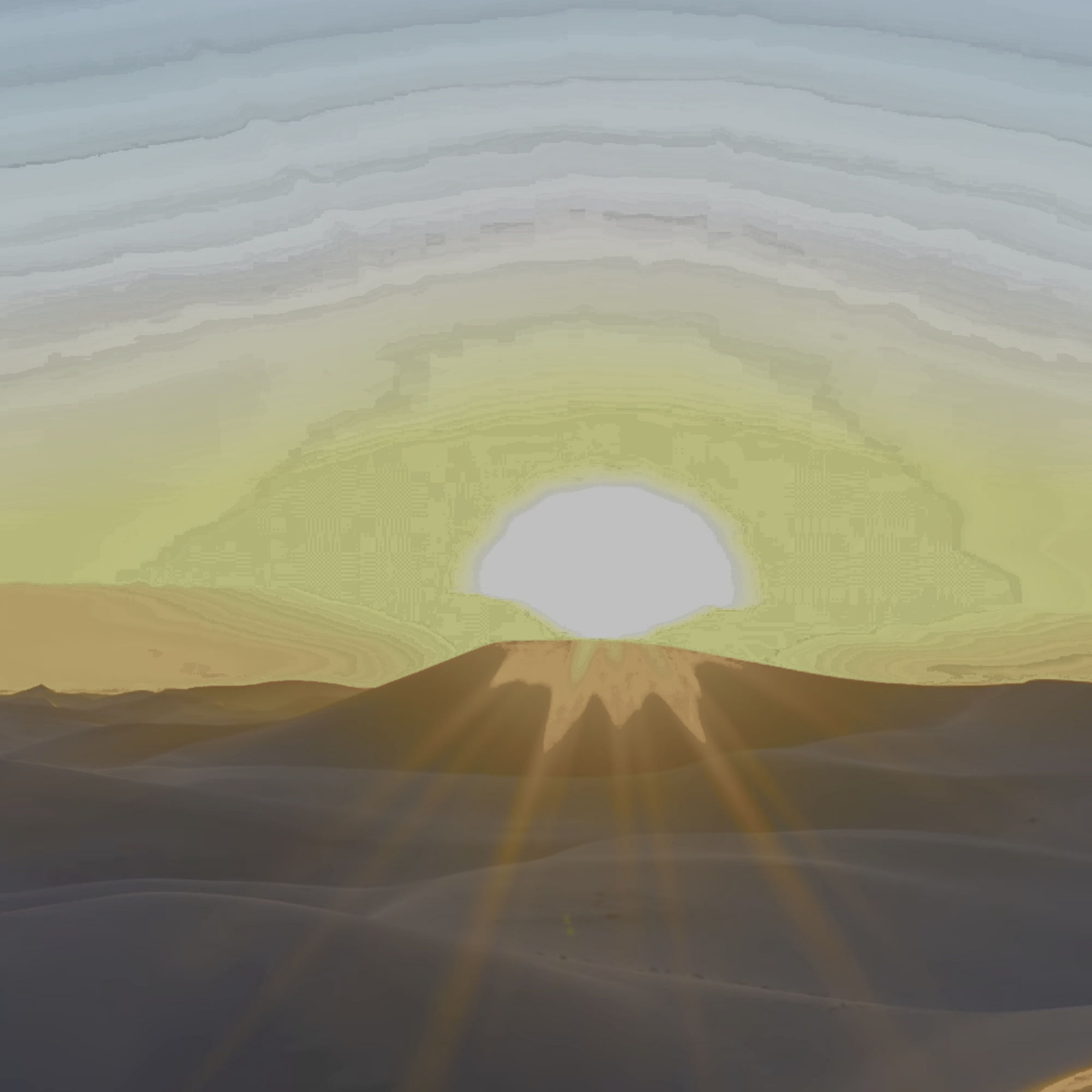}
		&\includegraphics[width=0.99in]{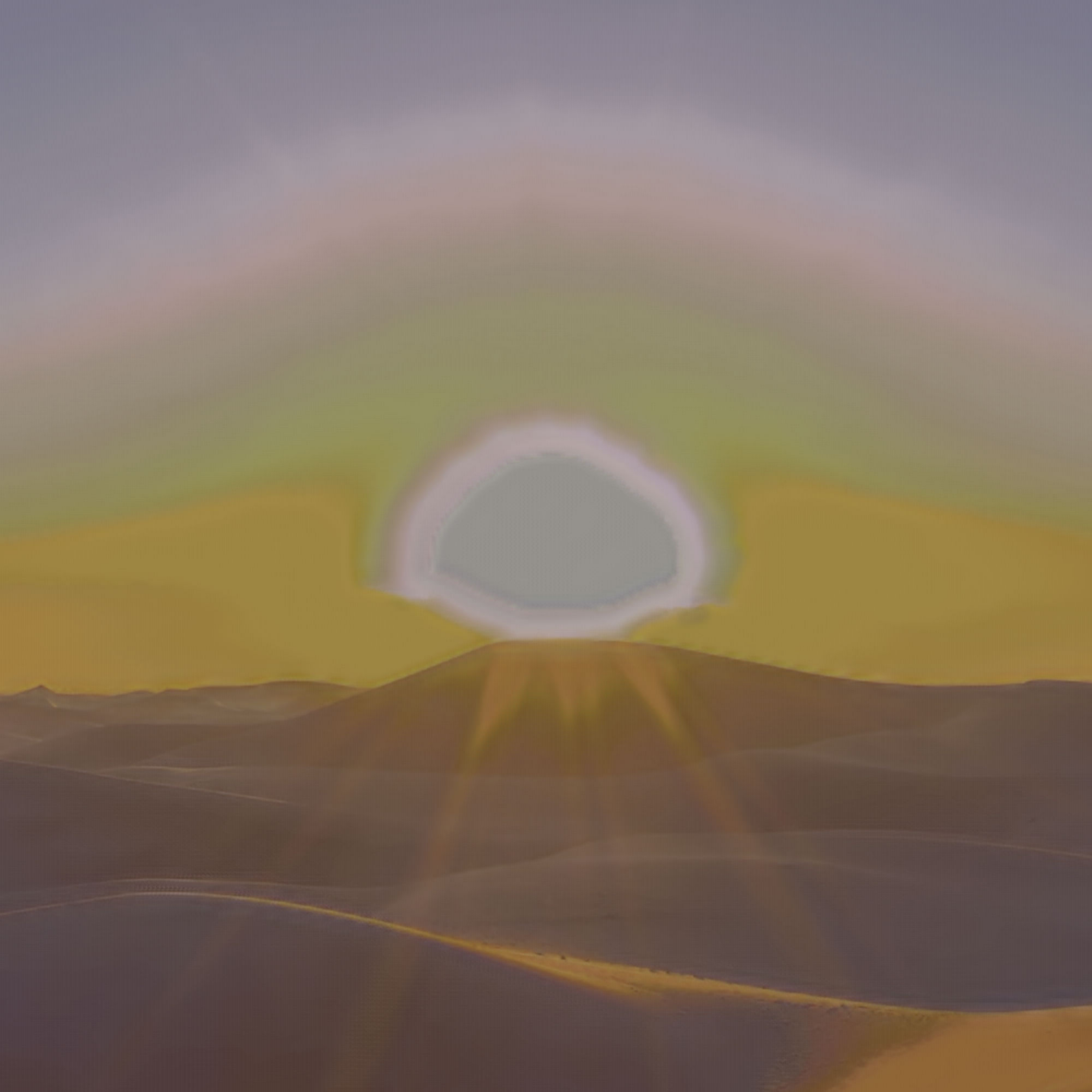}
		&\includegraphics[width=0.99in]{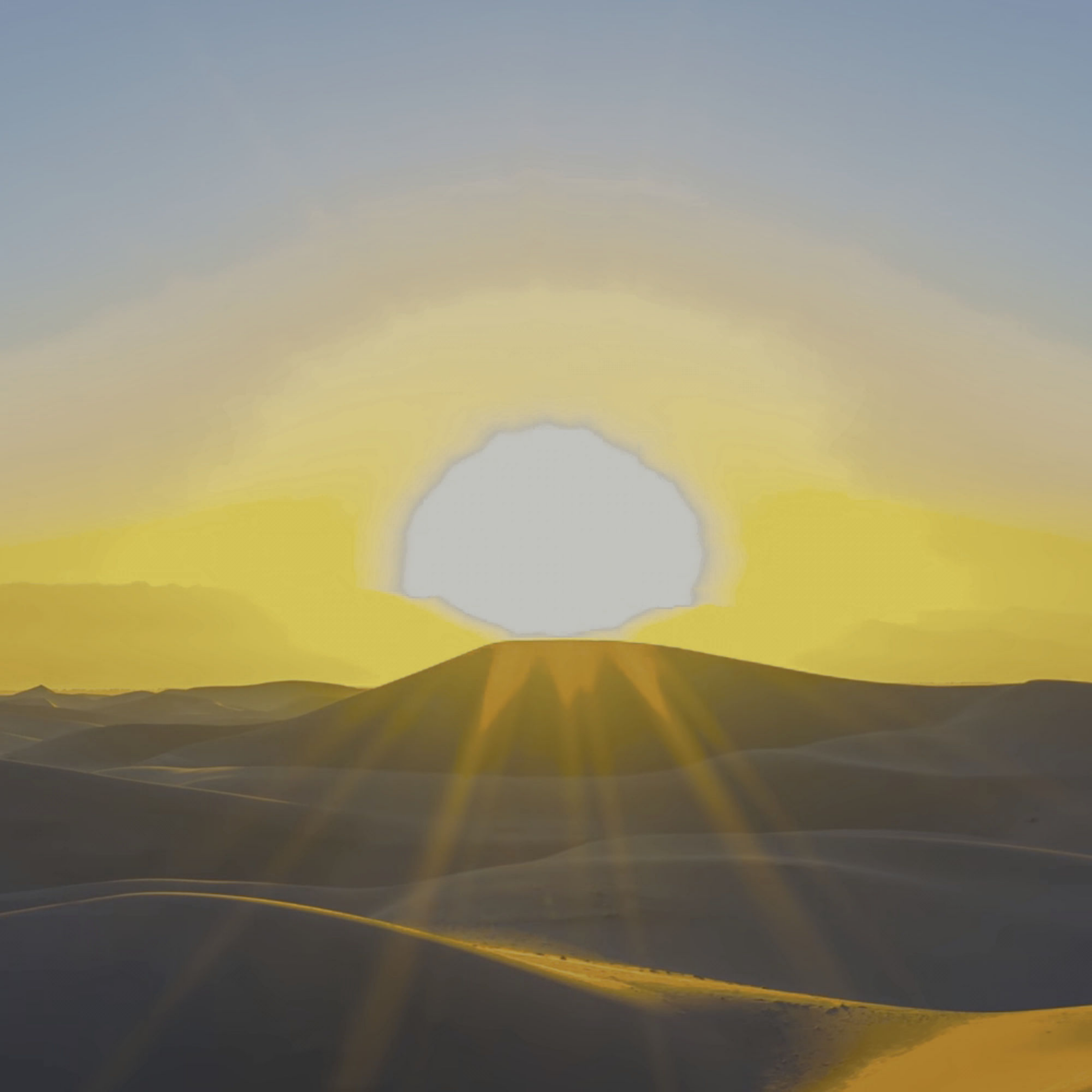}
		&\includegraphics[width=0.99in]{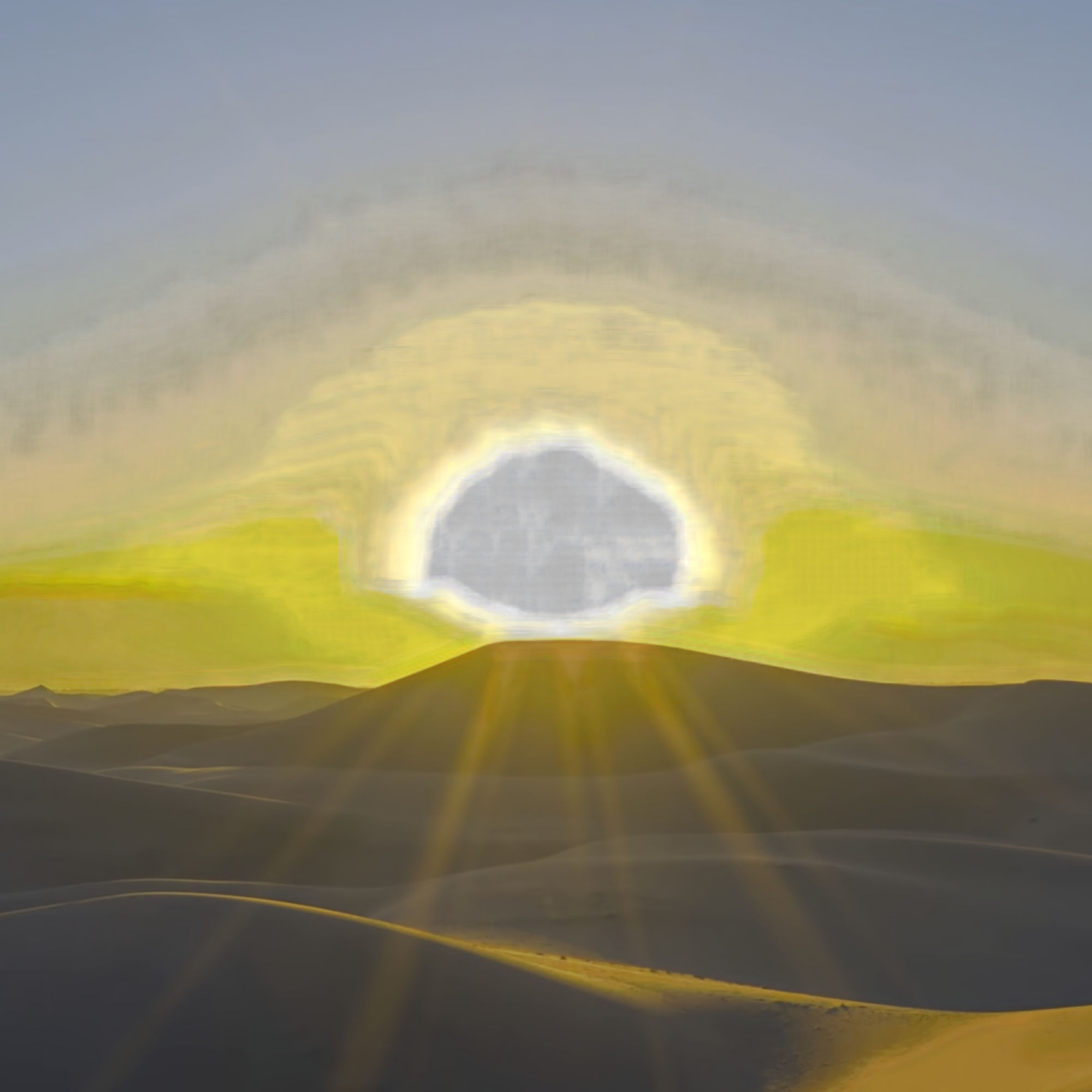}
		&\includegraphics[width=0.99in]{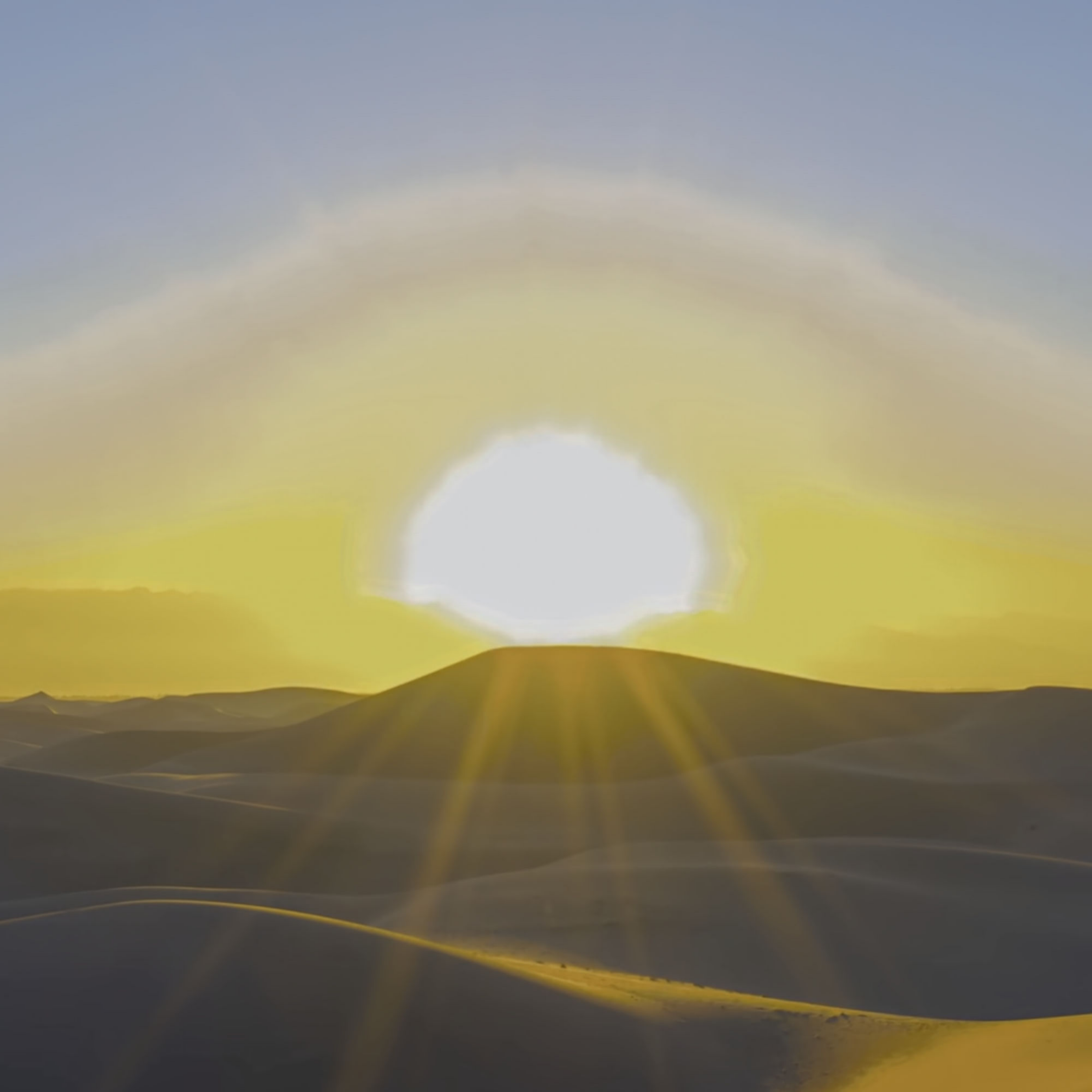}
		&\includegraphics[width=0.99in]{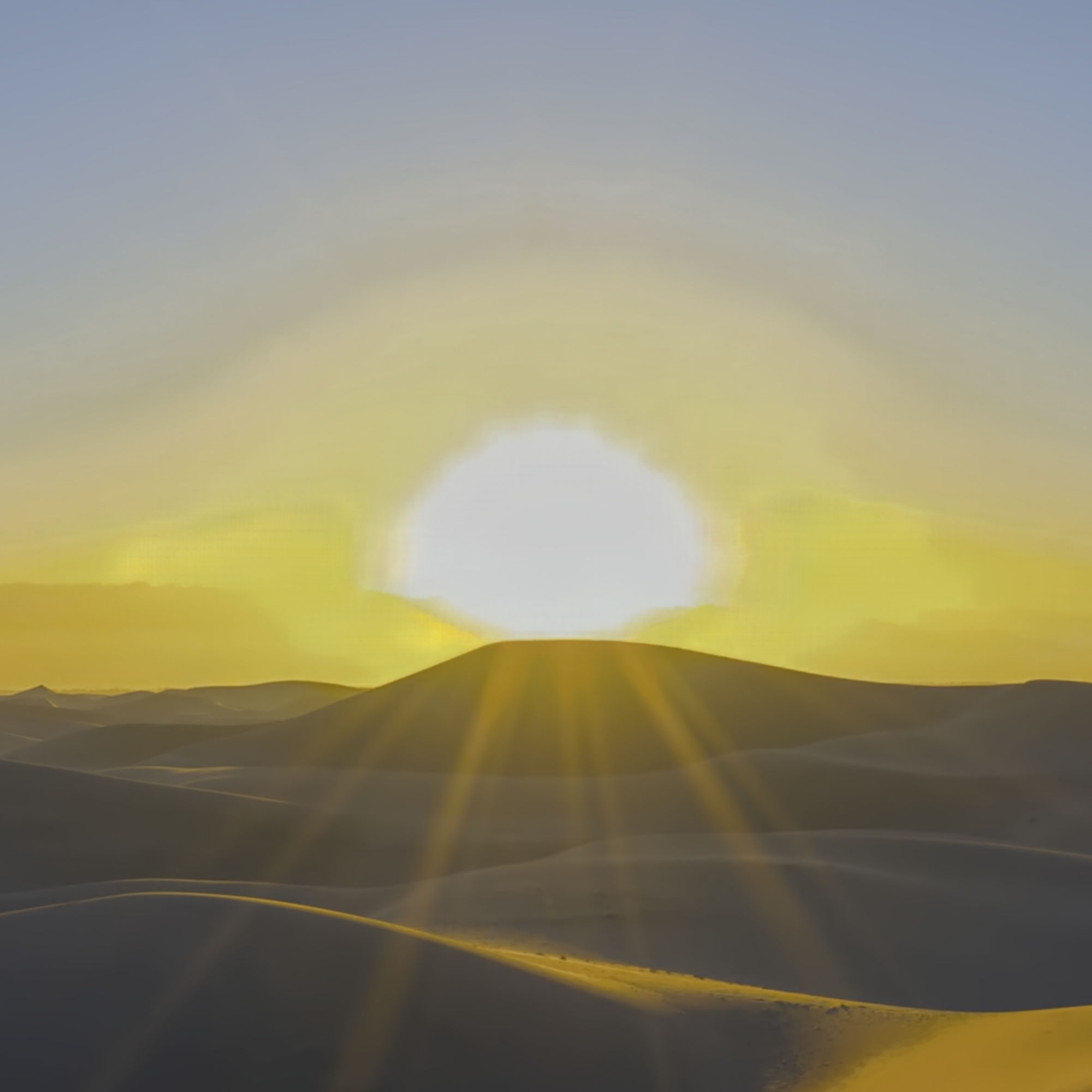}
		&\includegraphics[width=0.99in]{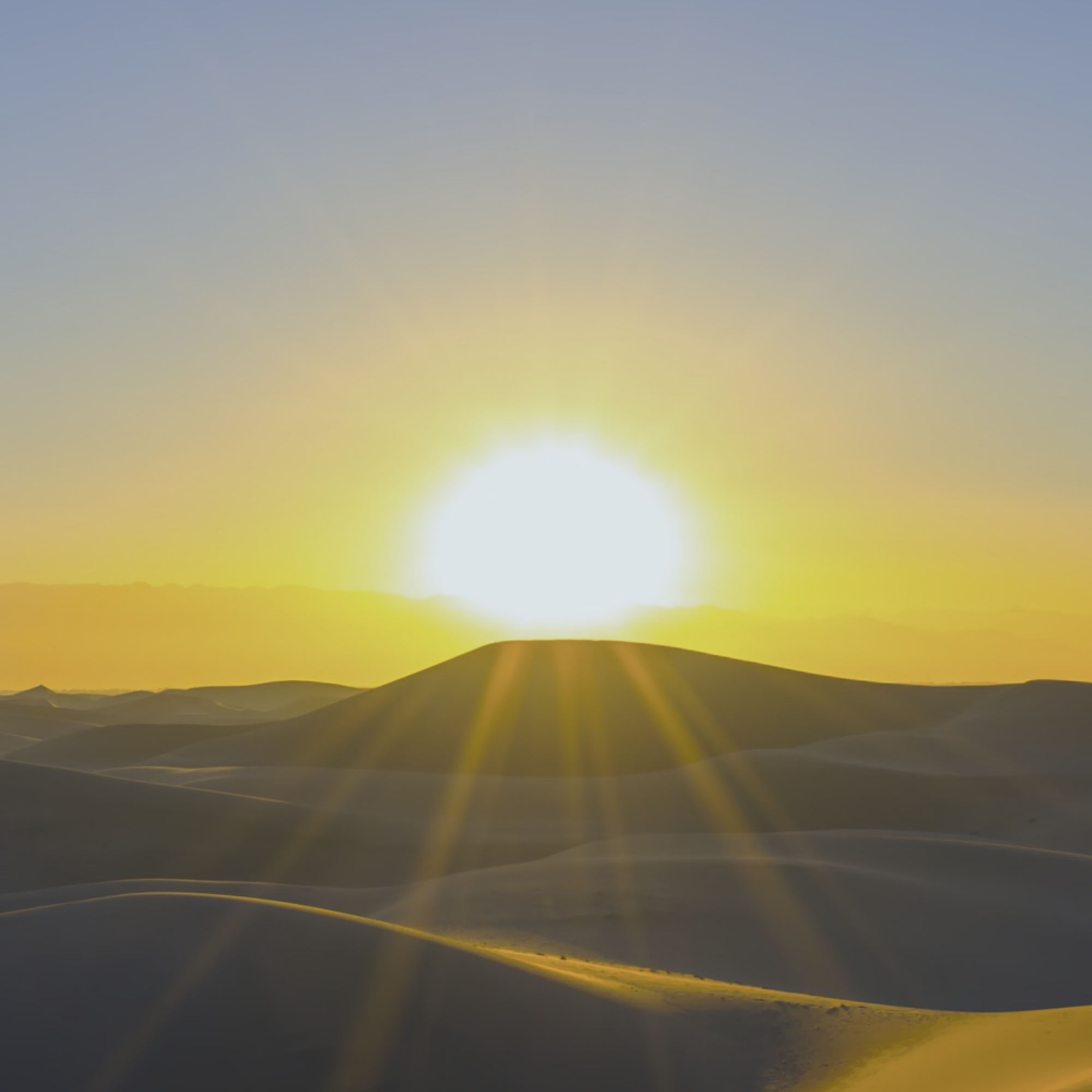}
		\\
		KovaleskiEO\cite{kovaleski2014high}& CycleGAN\cite{zhu2017unpaired}&CSRNet\cite{he2020conditional}& JSI-GAN\cite{kim2020jsi}&HDRTVNet\cite{hdrtvnet_conf}&AGCM-LE++& GT
		\\ 
	\end{tabular}
	\caption{Visual comparison on HDRTV1K.}
	\vspace{-3pt}
	\label{Figure 3 visualization} 
\end{figure*}

\textbf{Visual comparison.} Figure \ref{Figure 3 visualization} presents the results of visual comparison.
LDR-to-HDR and image-to-image translation methods often produce low-contrast images.
Except for HuoPhyEO \cite{huo2014physiological}, LDR-to-HDR-based, image-to-image translation, and SDRTV-to-HDRTV approaches all generate unnatural colors and noticeable artifacts. 
Photo retouching methods perform relatively better but suffer from color distortion.
In contrast, our method produces natural colors and high contrast akin to the ground truth, without additional artifacts. 
Notably, the visual quality improves with processing steps: AGCM $<$ AGCM-LE $<$ AGCM-LE-HR, demonstrating the effectiveness of our proposed solution pipeline.

\begin{figure*}
	\scriptsize
	\centering
	\setlength{\tabcolsep}{0.05cm}
	\begin{tabular}{ccccccc}
		\includegraphics[width=0.99in]{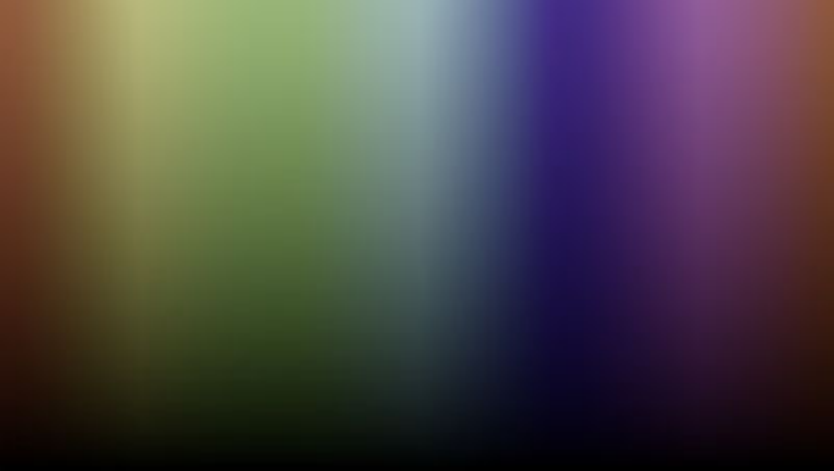}
		&\includegraphics[width=0.99in]{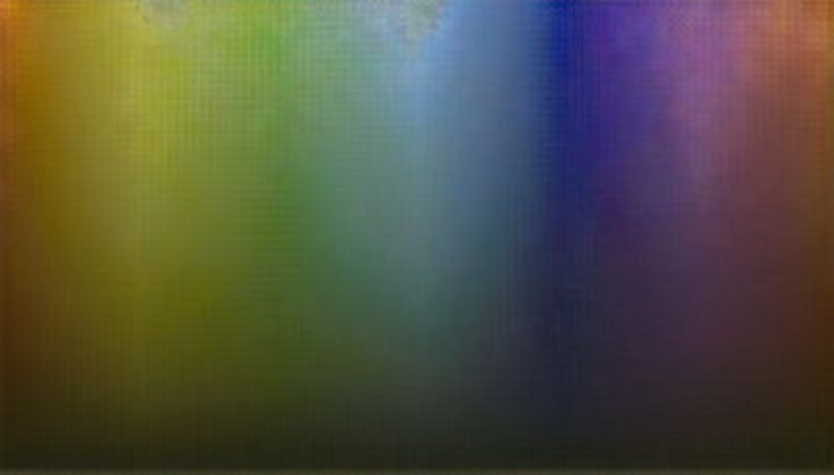}
		&\includegraphics[width=0.99in]{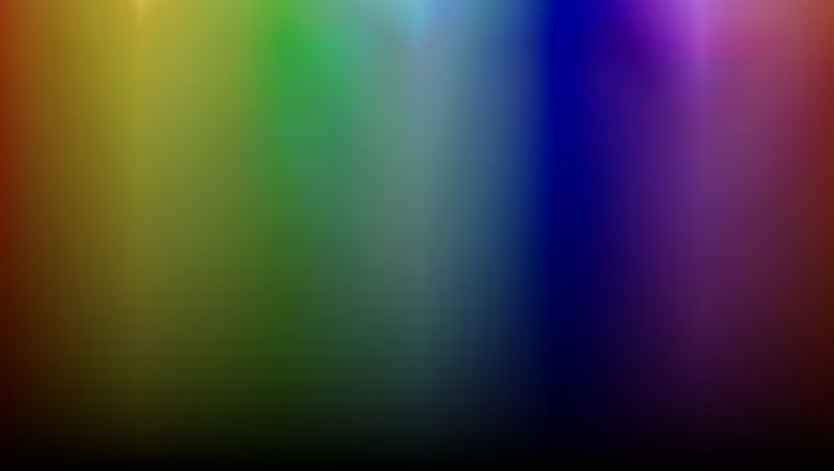}
		&\includegraphics[width=0.99in]{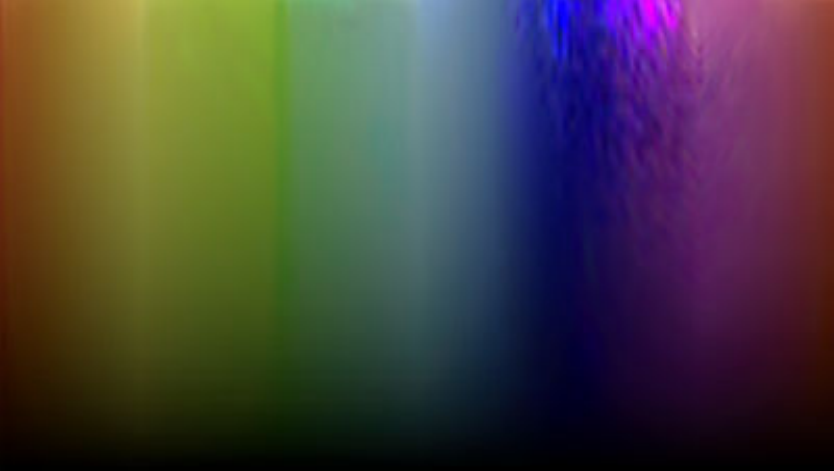}
		&\includegraphics[width=0.99in]{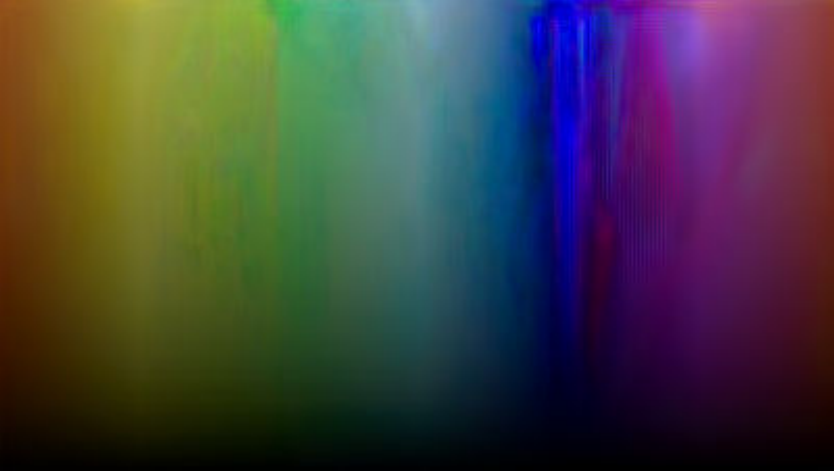}
		&\includegraphics[width=0.99in]{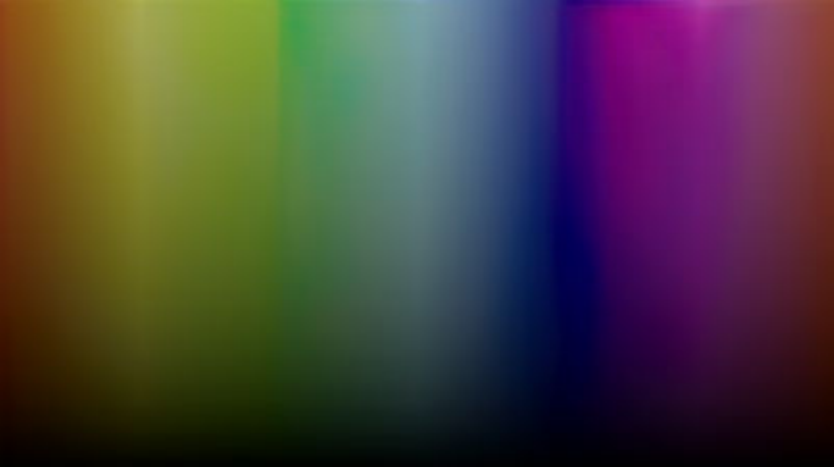}
		&\includegraphics[width=0.99in]{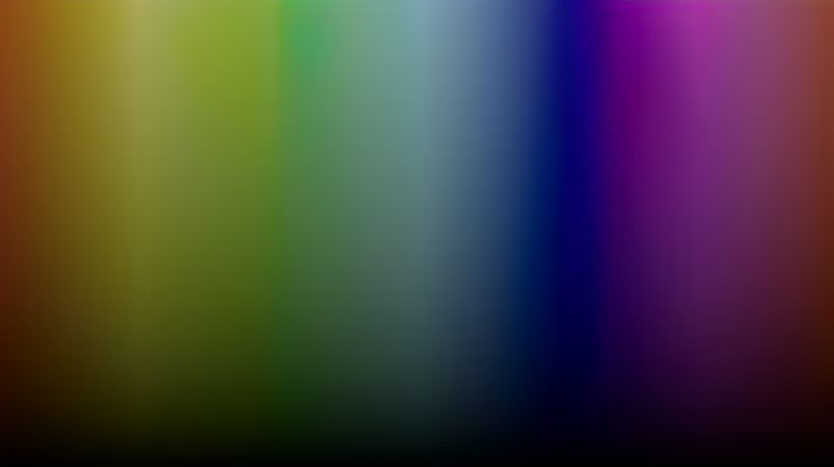}
		\\
		HuoPhyEO \cite{huo2014physiological}& Pixel2Pixel\cite{isola2017image}&HDRNet\cite{gharbi2017deep}& Ada-3DLUT\cite{zeng2020learning}&Deep SR-ITM\cite{kim2019deep}&AGCM++& AGCM-LE-HR++
		\\ 
  		\includegraphics[width=0.99in]{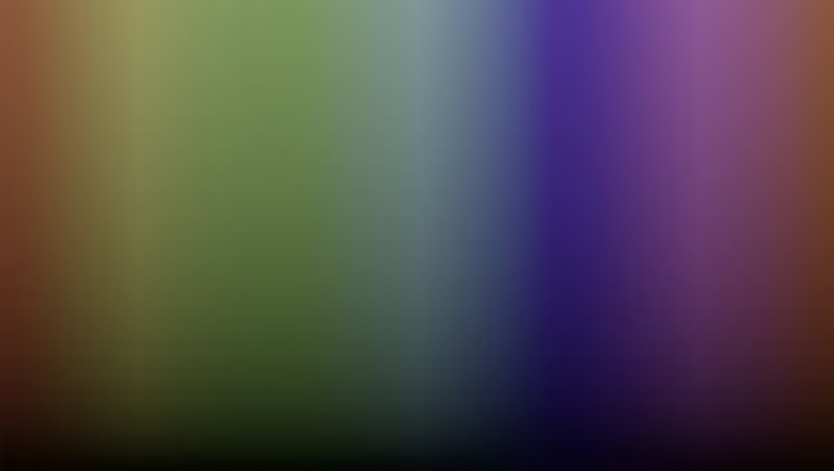}
		&\includegraphics[width=0.99in]{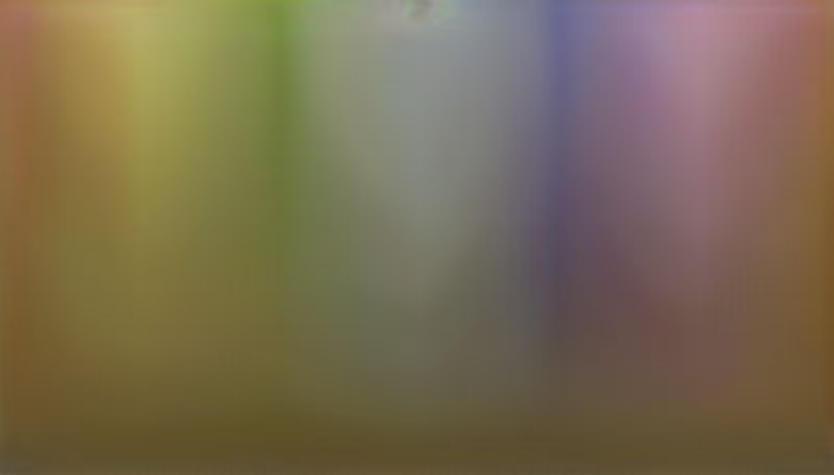}
		&\includegraphics[width=0.99in]{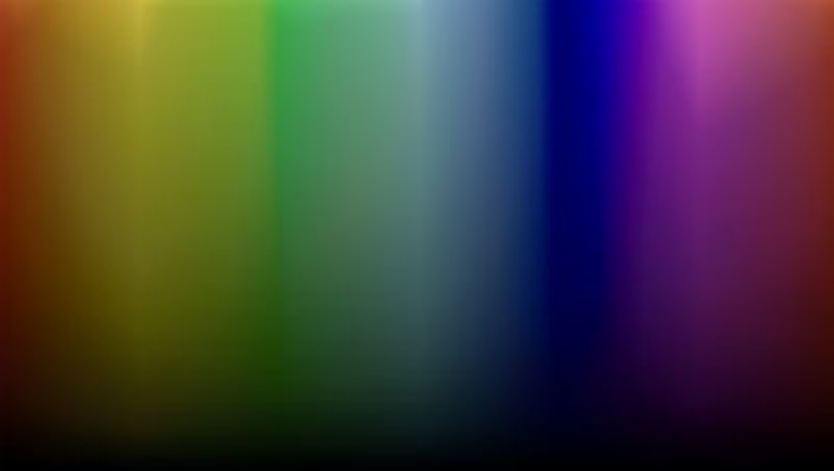}
		&\includegraphics[width=0.99in]{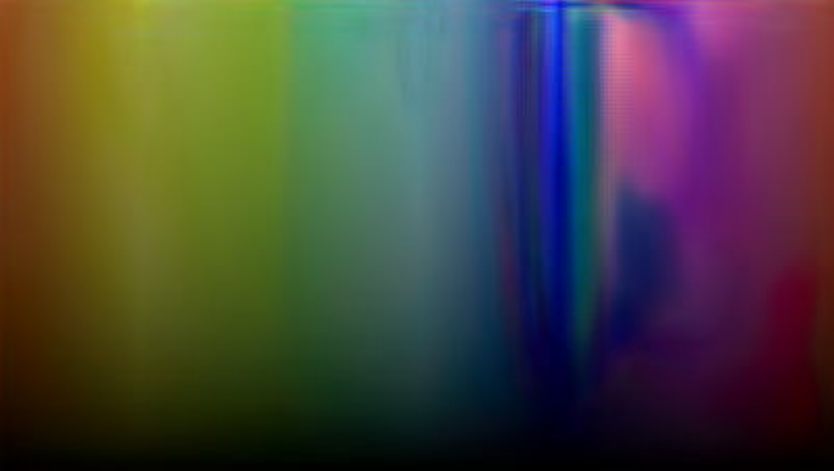}
		&\includegraphics[width=0.99in]{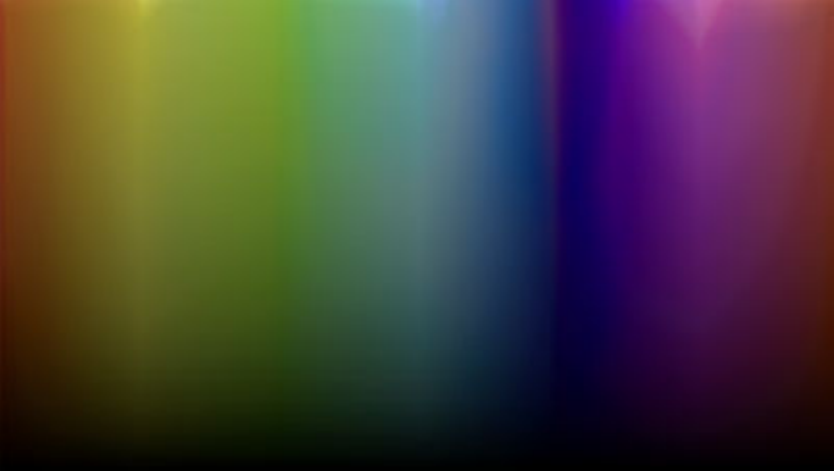}
		&\includegraphics[width=0.99in]{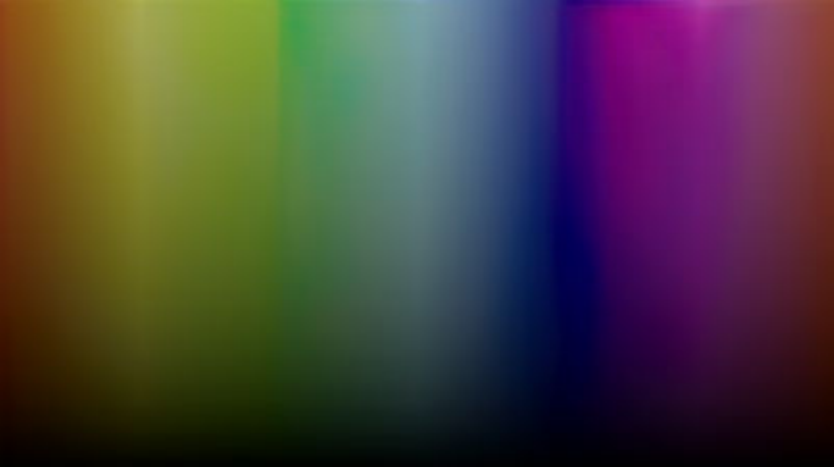}
		&\includegraphics[width=0.99in]{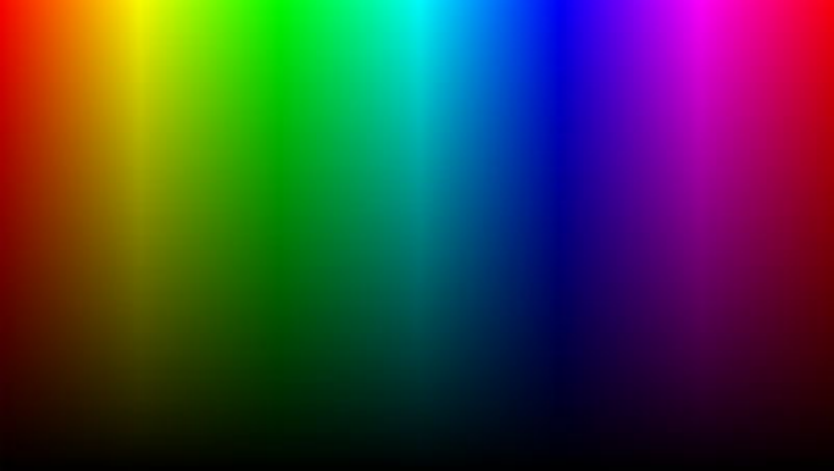}
		\\
		KovaleskiEO\cite{kovaleski2014high}& CycleGAN\cite{zhu2017unpaired}&CSRNet\cite{he2020conditional}& JSI-GAN\cite{kim2020jsi}&HDRTVNet\cite{hdrtvnet_conf}&AGCM-LE++& Input SDRTV
		\\ 
	\end{tabular}
	\caption{Visual comparison of the color transition test.}
	\vspace{-12pt}
	\label{Figure 4 color transition} 
\end{figure*}

\begin{table*}[ht]
\caption{Quantitative comparisons between methods w/ and w/o AGCM.}
\label{quantitative_comparison_w_wo_AGCM}
\centering
\tabcolsep=6.3mm
\begin{tabular*}{\linewidth}{c|c|ccccc}
\hline
Method & Params$\downarrow$ & PSNR$\uparrow$ & SSIM$\uparrow$ & SR-SIM$\uparrow$ & $\Delta E_{ITP}\downarrow$ & HDR-VDP3$\uparrow$ \\ \hline
LE(Basic3x3) & 40K & 36.98 & 0.9706 & \textbf{0.9989} & 9.63 & 8.368 \\
AGCM-LE(Basic3x3) & 75K & \textbf{37.50} & \textbf{0.9721} & 0.9988 & \textbf{9.13} & \textbf{8.580} \\ \hline
LE(ResNet) & 1.37M & 37.32 & 0.9720 & 0.9950 & 9.02 & 8.391 \\
AGCM-LE(ResNet) & 1.41M & \textbf{37.61} & \textbf{0.9726} & \textbf{0.9967} & \textbf{8.89} & \textbf{8.613} \\  \hline
LE(UNet) & 556K & 37.08 & 0.9705 & 0.9956 & 9.27 & 8.315 \\
AGCM-LE(UNet) & 591K & \textbf{38.60} & \textbf{0.9745} & \textbf{0.9973} & \textbf{7.67} & \textbf{8.696} \\  \hline
\end{tabular*}
\vspace{-3pt}
\end{table*}

\begin{figure*}[!t]
\subfigure[Visual results.]{
\label{visual_comparison_w_wo_AGCM(a)}
\begin{minipage}[t]{0.536\linewidth}
\includegraphics[scale=0.1365]{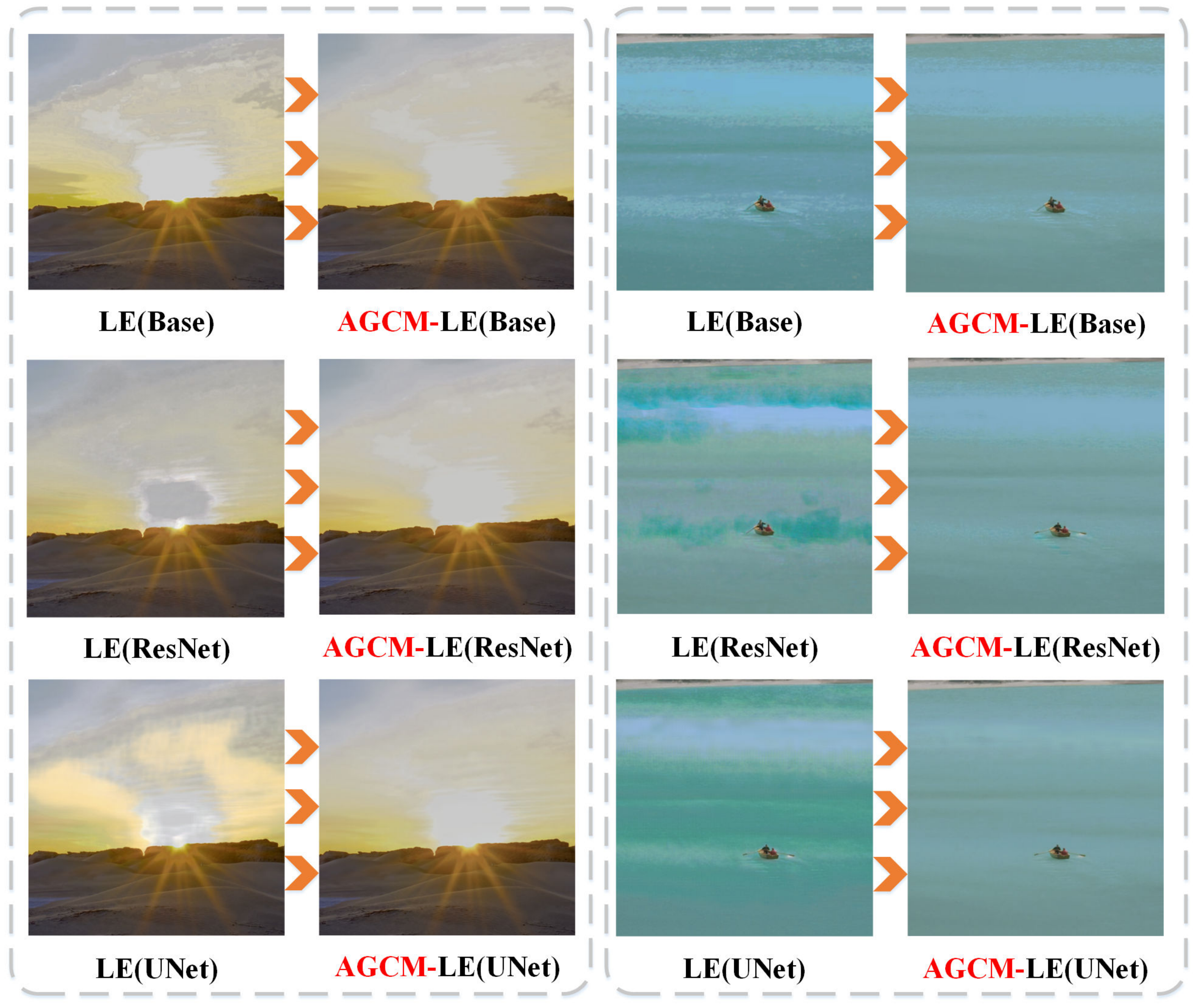}
\end{minipage}%
}
\subfigure[Color transition test.]{
\label{visual_comparison_w_wo_AGCM(b)}
\begin{minipage}[t]{0.45\linewidth}
\includegraphics[scale=0.137]{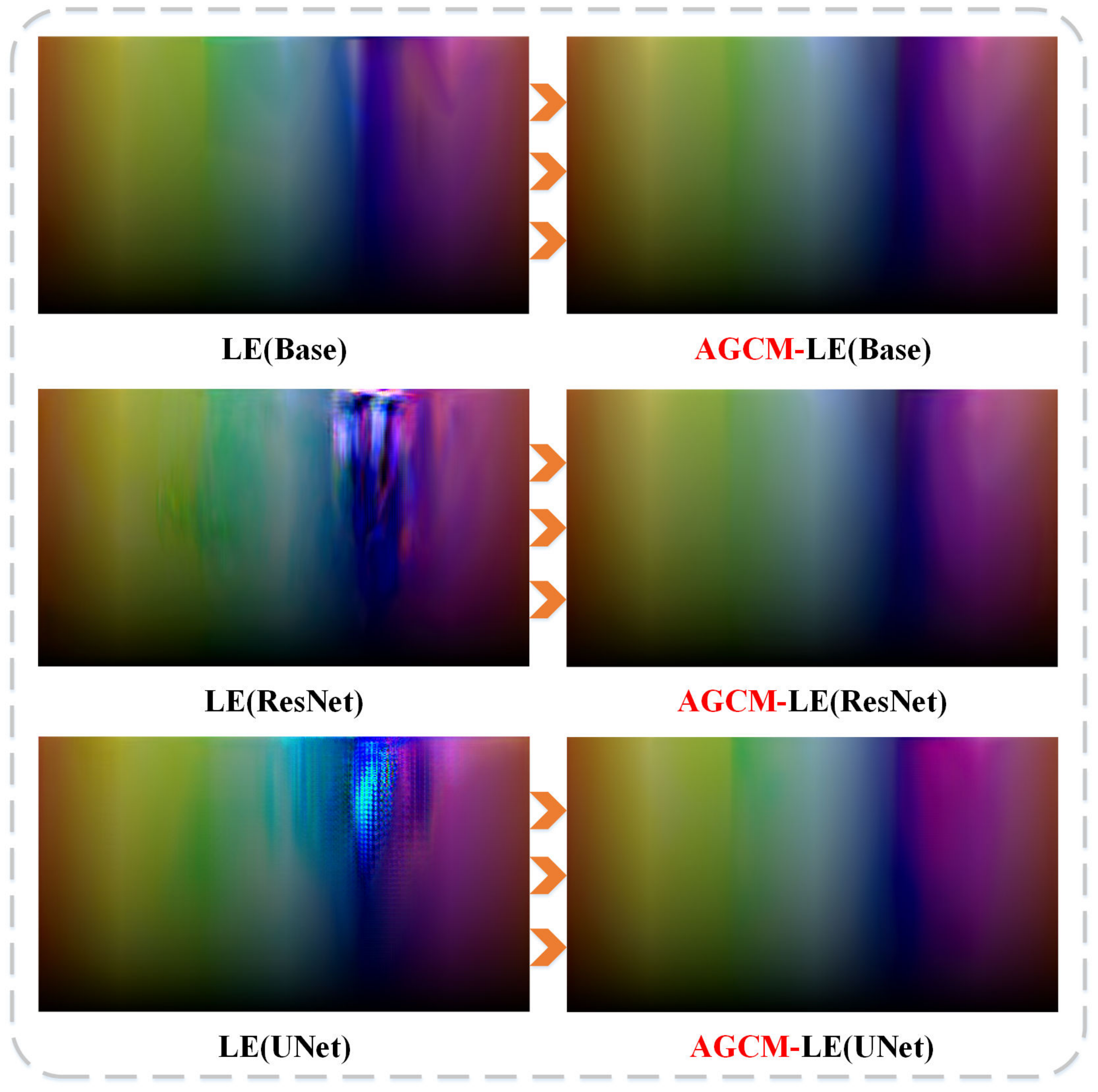}
\end{minipage}%
}
 \caption{Visual comparisons between methods w/ and w/o AGCM.}
\vspace{-3pt}
\label{visual_comparison_w_wo_AGCM}
\end{figure*}

\begin{figure}[!ht]
\centering
\begin{minipage}{0.46\linewidth}
\centering
\subfigure[The identity mapping of SDRTV colors.]{
\centering
\label{Identity_LUT}
\includegraphics[width=\linewidth]{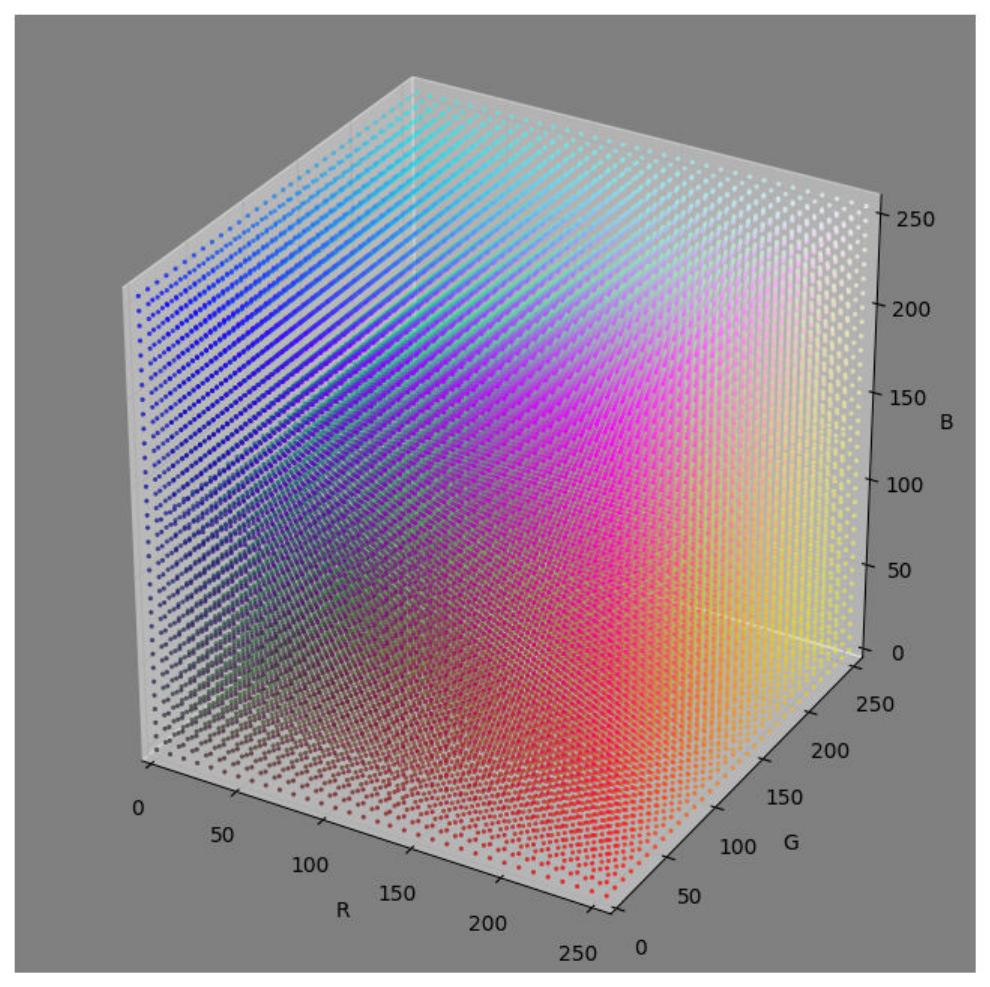}}
\end{minipage}\qquad
\begin{minipage}{0.46\linewidth}
\centering
\label{BasicNet_LUT33}
\subfigure[The SDRTV-to-HDRTV color mapping of the base network.]{
\centering
\includegraphics[width=\linewidth]{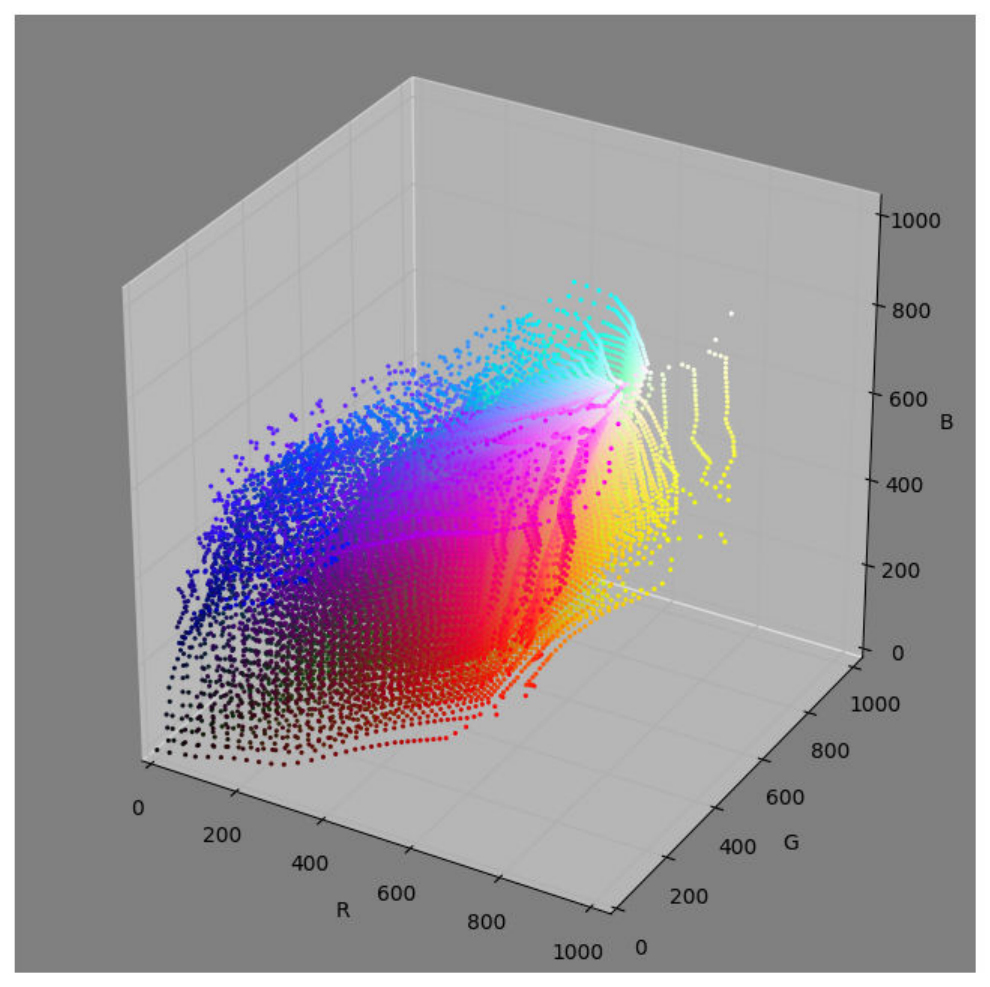}}
\end{minipage}
\caption{Illustrations of 3D LUTs.}
\label{tonemapping}
\vspace{-10pt}
\end{figure}

\subsection{Color Transition Test}
Previous methods often perform poorly in highlight regions, especially with color changes. We conduct a color transition test using a man-made color card as input, comparing outputs from different methods, as presented in Figure \ref{Figure 4 color transition}.
It can be observed that unnatural transitions and color blending appear in outputs from region-dependent methods (e.g., Deep SR-ITM, JSI-GAN, Pixel2Pixel, CycleGAN) and those based on region-dependent conditions (e.g., 3D-LUT).
In contrast, our method achieves smooth transition with AGCM, based on pixel-independent operations. 
Even when learning region-dependent mapping (e.g., AGCM-LE, AGCM-LE-HR), Our method avoids color transition artifacts. 
This showcases the superiority of our AGCM to deal with the color gamut conversion, and the effectiveness of our entire solution pipeline to resolve the complex SDRTV-to-HDRTV mappings.
Note that blue regions suffer the most severe unnatural color transition. This is due to the greater information loss appear in blue regions during color gamut compression in SDRTV production.

\begin{figure*}[!t]
   \begin{center}
   \includegraphics[width=\linewidth]{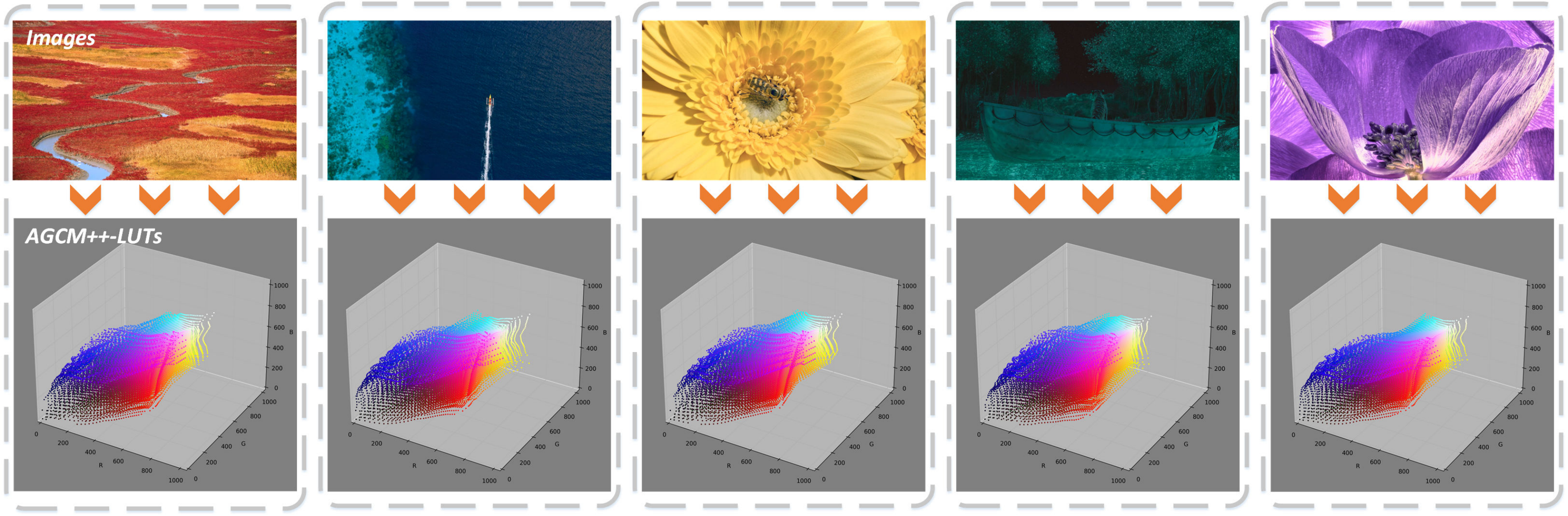}
   \end{center}
   \vspace{-8pt}
      \caption{3D LUTs generated by taking various images as input conditions of our AGCM network.}
    \vspace{-8pt}
   \label{LUT_comparison_of_various_images}
\end{figure*}

\begin{figure}[t!]
   \begin{center}
   \includegraphics[width=\linewidth]{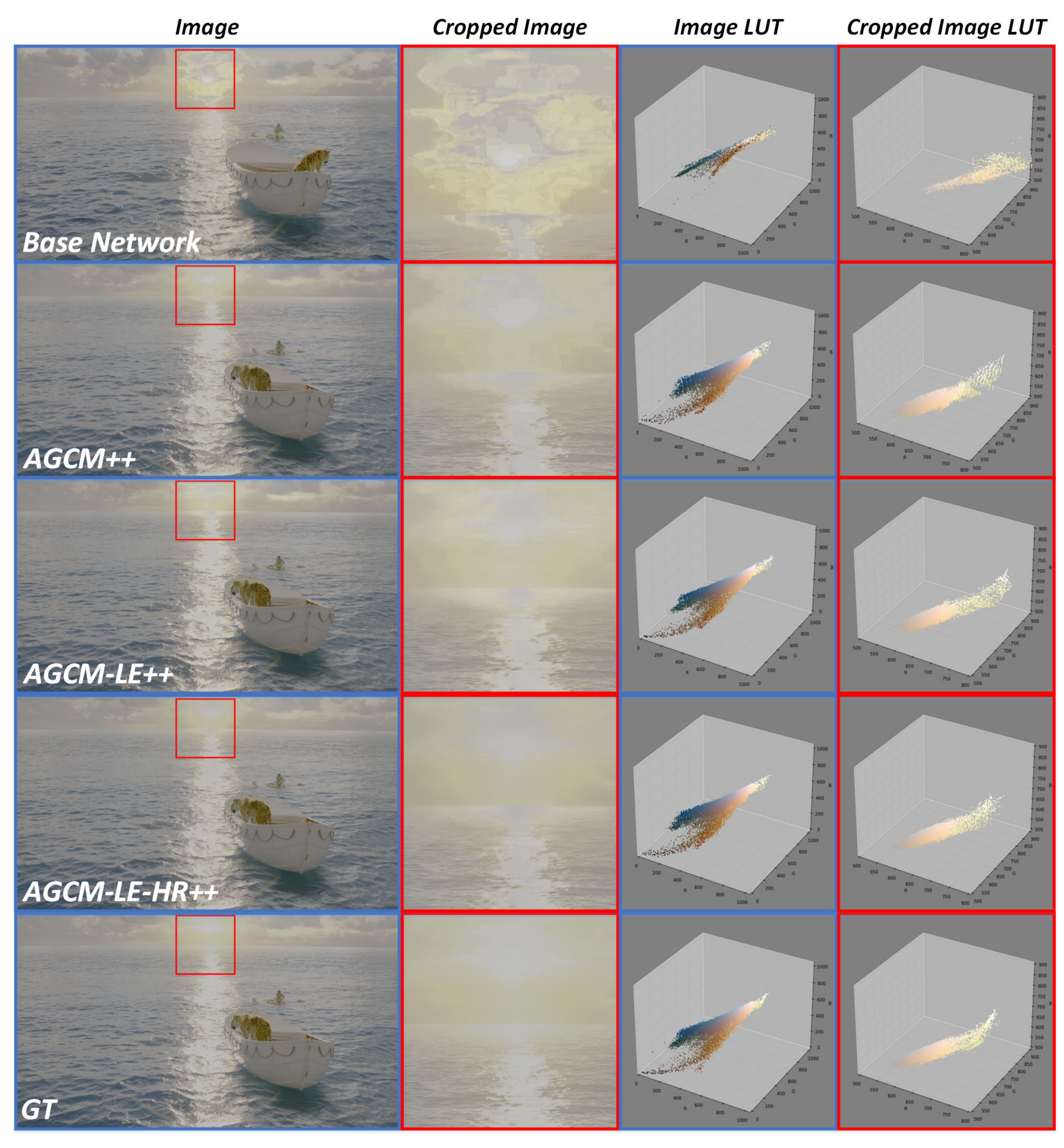}
   \end{center}
      \caption{Visual results and 3D LUT manifolds in different stages.}
    \vspace{-10pt}
   \label{Figure_LUT_ablation}
\end{figure}

\subsection{Significance of AGCM}
\label{Analysis_on_the_necessity_of_AGCM}
To demonstrate the necessity of AGCM in the entire SDRTV-to-HDRTV solution pipeline, we conduct comprehensive experiments comparing methods with and without AGCM.
In addition to using ResNet (used in the initial version) and UNet-based (in this paper) LE networks, we also implement a very small-scale LE network, denoted as Basic3x3. It has a very simple structure with only three layers of convolution with standard 3$\times$3 filters. 
As presented in Table \ref{quantitative_comparison_w_wo_AGCM}, methods that perform AGCM before LE (i.e. AGCM-LE), with limited additional parameters, achieve significantly higher performance than methods that learn the LE network directly, regardless of the LE network's scale. 
For visual comparisons, Figure \ref{visual_comparison_w_wo_AGCM(a)} shows that outputs from methods without AGCM exhibit noticeable artifacts in over-exposed and saturated regions. Figure~\ref{visual_comparison_w_wo_AGCM(b)} also demonstrates that methods without AGCM perform poorly in the color transition test. 
Additionally, we can observe an interesting phenomenon that when comparing the different LE networks (without AGCM), larger networks tend to produce more artifacts. 
The smallest LE network Basic3x3 produce the best visual results despite the lowest quantitative performance. 
All these results indicate that optimizing pixel-independent and region-dependent mapping together is challenging for an end-to-end LE network, and simply improve the LE network cannot address this challenge.
Nevertheless, we find that performing AGCM prior to local enhancement is crucial for the final performance. This suggests that addressing color mapping before enhancement can effectively mitigates the optimization challenge and achieve considerable SDRTV-to-HDRTV results.

\subsection{Analysis of color mapping via LUT manifold}
\label{Analysis_of_color_mapping_via_LUT_manifold}
In this section, we provide an analysis tool by visualizing the Look-Up Tables (LUTs) manifold to intuitively evaluate the model's function at various stages. 
We begin by illustrating the LUT manifold and its visualization. In a 3D LUT cube, each point has four basic attributes: color and three coordinate values, which determine the position of the color within the current domain. 
Figure \ref{Identity_LUT} shows the 3D LUT cube of the identity mapping of SDRTV colors, where the coordinate values of each point correspond to the three-channel values of its color in the SDRTV domain. For instance, the color (R:128, G:128, B:128) is located at the position (128, 128, 128) in the space. 
Figure \ref{BasicNet_LUT33} presents the LUT manifold of SDRTV-to-HDRTV color mapping using the base network. Within this cube, the coordinate values of each point correspond to its corresponding HDRTV color. It can be observed that SDRTV colors (0-255) map to HDRTV colors (0-1023).
To demonstrate image-adaptive functionality, Figure \ref{LUT_comparison_of_various_images} shows LUT manifolds changing with different inputs, indicating effective color conditioning of our AGCM. 
We further demonstrate the functionality of different steps using LUT manifold visualization in Figure \ref{Figure_LUT_ablation}.
Firstly, the base network can only learn a one-to-one color mapping throughout the dataset, resulting in a non-smooth color transition of the LUT manifold in highlight areas and severe artifacts in the output. 
In contrast, the color condition network helps the base network learn image-adaptive color mapping, eliminating artifacts in the results generated by AGCM and densifying the LUT manifold. 
Due to local enhancement, the LUT manifold becomes more compact and smooth, allowing an SDRTV color to be mapped to multiple HDRTV colors through region-dependent operations (i.e., convolutions). It can handle one-to-many color mapping and greatly improve visual quality. 
Lastly, we can see that highlight refinement further compacts and densifies the LUT manifold, and the results also have natural color transition in highlight regions.
The above LUT manifold analysis intuitively reflects the role of different stages in our SDRTV-to-HDRTV solution pipeline.

\subsection{Network Investigation}
In this section, we conduct comprehensive experiments to investigate the specific network design in our method. 
\textbf{Adaptive Global Color Mapping.} We first examine the effects of the depths of the base network and the condition network in AGCM, as shown in Table \ref{ablation_study_AGCM_on_base_model} and Table \ref{ablation_study_AGCM_on_condition_network}. 
We vary the depth of the base network from 2 to 5 and set the depth of the condition network from 3 to 6.
Experimental results show that a base network depth of 3 and a condition network depth of 5 achieves the best performance, thereby being set as the default setting in our method.
We also conduct an ablation study on the key components of the proposed AGCM. As shown in Figure \ref{ablation_study_AGCM}, removing the Dropout layer slightly reduces performance, indicating the effectiveness of Dropout in the condition network.
Notably, we can see that without instance normalization will result in a significant performance drop, performing only slightly better than the base network without the condition.
This illustrates that this normalization layer is critical for the proposed color condition. 
We believe this is because instance normalization can help the network learn the key property (i.e., contrast) as the condition.

\textbf{Local Enhancement.} We improved upon the preliminary ResNet-style network by introducing a UNet-style network for local enhancement (See Section \ref{Local_Enhancement_method}).
For fair comparison, we adopt the same AGCM model, i.e., AGCM++, prior to the LE network.
As shown in Table \ref{ablation_study_LE}, our LE++ outperforms previous LE by over 0.8dB with about half the parameters, indicating the superiority of LE++.
There are two primary reasons for the success of our LE++ design: (1) its U-shape design enables stronger representation ability to learn useful features, particularly for high-resolution inputs; and (2) its ability to handle spatially varying mappings via conditional branches makes it particularly suitable for operations that vary across different regions, such as local tone mapping operators.

\begin{table*}[ht]
\centering
\tabcolsep=7mm
\begin{threeparttable}[b]
\caption{Quantitative comparisons on the depth of the base network in AGCM.}
\label{ablation_study_AGCM_on_base_model}

\begin{tabular*}{\linewidth}{c|c|ccccc}
\hline
Method & Params$\downarrow$ & PSNR$\uparrow$ & SSIM$\uparrow$ & SR-SIM$\uparrow$ & $\Delta E_{ITP}\downarrow$ & HDR-VDP3$\uparrow$ \\ \hline
AGCM-C5B2\tnote{1}      & 30.2K & 36.65                 & 0.9664 & 0.9966 & 10.11 & 8.497 \\
AGCM-C5B3               & 35.3K & \textbf{37.35}        & \textbf{0.9666} & \textbf{0.9968} & 9.29 & \textbf{8.511} \\
AGCM-C5B4               & 40.3K & 37.31                 & 0.9662 & 0.9966 & \textbf{9.16} & 8.497 \\
AGCM-C5B5               & 45.4K & 37.31                 & 0.9670 & 0.9969 & 9.35 & 8.504 \\ \hline
\end{tabular*}%
\begin{tablenotes}
\item [1] \textit{B} means the depth of the base network, while \textit{C} represents the depth of the condition network.
\end{tablenotes}
\end{threeparttable}
\end{table*}

\begin{table*}[ht]
\centering
\tabcolsep=7.0mm
\begin{threeparttable}[b]
\caption{Quantitative comparisons on the depth of the condition network in AGCM.}
\label{ablation_study_AGCM_on_condition_network}

\begin{tabular*}{\linewidth}{c|c|ccccc}
\hline
Method & Params$\downarrow$ & PSNR$\uparrow$ & SSIM$\uparrow$ & SR-SIM$\uparrow$ & $\Delta E_{ITP}\downarrow$ & HDR-VDP3$\uparrow$ \\ \hline
AGCM-C3B3\tnote{1}      & 8.4K & 36.42                  & 0.9645 & 0.9966 & 10.25 &  8.492 \\
AGCM-C4B3               & 13.9K & 36.84                 & \textbf{0.9667} & 0.9965 & 9.56 & 8.509 \\
AGCM-C5B3               & 35.3K & \textbf{37.35}        & 0.9666 & \textbf{0.9968} & \textbf{9.29} & \textbf{8.511} \\
AGCM-C6B3               & 118.8K & 37.05                & 0.9663 & 0.9966 & 9.56 & 8.486 \\ \hline
\end{tabular*}%
\begin{tablenotes}
\item [1] \textit{B} means the depth of the base network, while \textit{C} represents the depth of the condition network.
\end{tablenotes}
\end{threeparttable}
\end{table*}

\begin{table*}[t!]
\centering
\tabcolsep=6.5mm
\begin{threeparttable}[b]
\caption{Quantitative comparisons on the critical layers in AGCM.}
\label{ablation_study_AGCM}
\centering
\begin{tabular*}{\linewidth}{c|c|ccccc}
\hline
Method & Params$\downarrow$ & PSNR$\uparrow$ & SSIM$\uparrow$ & SR-SIM$\uparrow$ & $\Delta E_{ITP}\downarrow$ & HDR-VDP3$\uparrow$ \\ \hline
BaseModel-B3            & 4.6K & 36.37                  & 0.9556 & 0.9963 & 10.22 & 8.397 \\
AGCM-woDropout\tnote{1} & 35.3K & 37.00                 & 0.9658 & \textbf{0.9968} & 9.39 & 8.497 \\
AGCM-woIN\tnote{2}      & 34.8K & 36.60                 & 0.9645 & 0.9967 & 11.36 & 8.473 \\
AGCM                    & 35.3K & \textbf{37.35}        & \textbf{0.9666} & \textbf{0.9968} & \textbf{9.29} & \textbf{8.511} \\ \hline

\end{tabular*}%
\begin{tablenotes}
\item [1] \textit{woDropout} means the Dropout layers are disabled. \item [2] \textit{woIN} indicates the Instance Normalization layers are disabled.
\end{tablenotes}
\end{threeparttable}
\end{table*}

\begin{table*}[t!]
\centering
\tabcolsep=7.7mm
\begin{threeparttable}[b]
\caption{Quantitative comparisons on different networks for LE.}
\label{ablation_study_LE}
\centering
\begin{tabular*}{\linewidth}{c|c|ccccc}
\hline
Method & Params$\downarrow$ & PSNR$\uparrow$ & SSIM$\uparrow$ & SR-SIM$\uparrow$ & $\Delta E_{ITP}\downarrow$ & HDR-VDP3$\uparrow$ \\ \hline
LE\tnote{1}                 & 1368K & 38.32               & 0.9736 & 0.9971 & 8.14      & 8.635 \\
LE++               & 556K  & 38.45               & 0.9739 & 0.9970 & 7.90     & 8.666\\
\hline

\end{tabular*}%
\begin{tablenotes}
\item [1] \textit{LE} denotes the network in the initial version \cite{hdrtvnet_conf} trained based on the same AGCM++ outputs.
\end{tablenotes}
\end{threeparttable}
\end{table*}

\subsection{Loss Comparison}
\label{loss_comparison}
\begin{figure}[t]
\centering
\includegraphics[width=\linewidth]{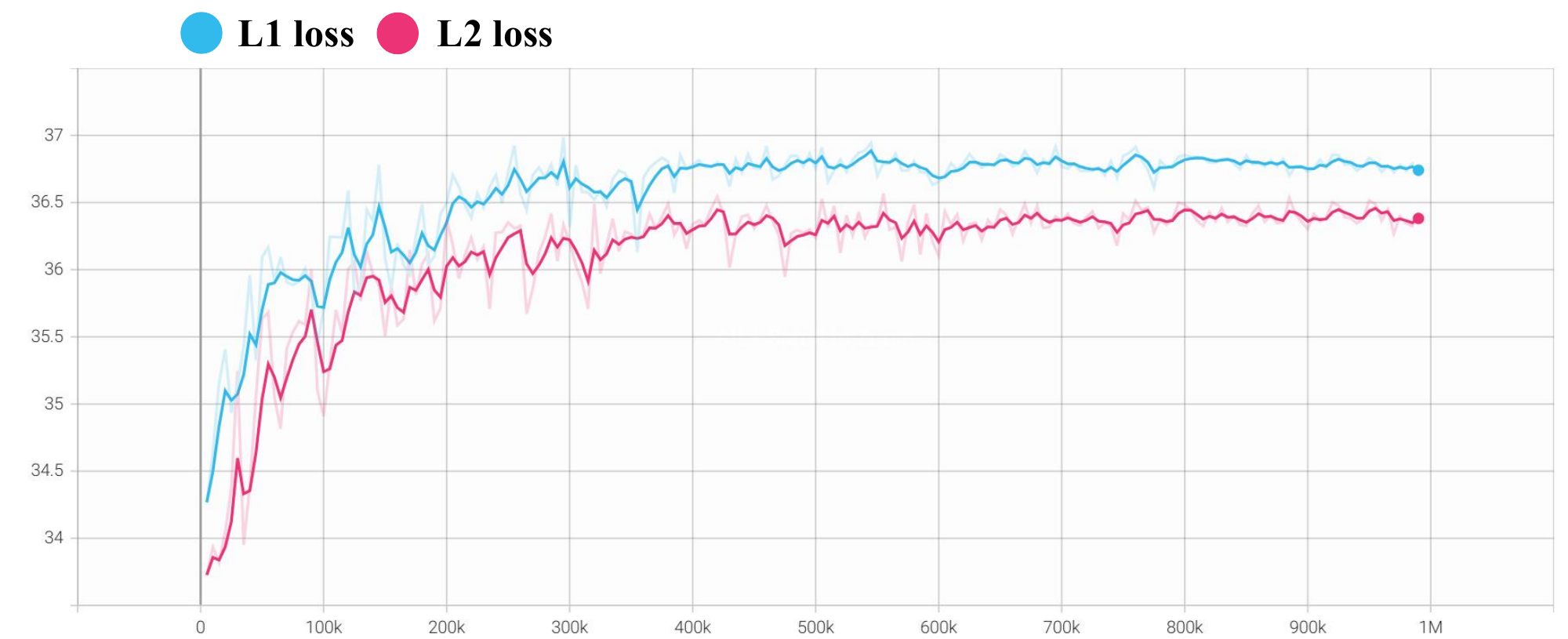}
\caption{Training curves based on two different loss functions.}
\vspace{-10pt}
\label{loss}
\end{figure}

In previous works involving SDRTV-to-HDRTV~\cite{kim2019deep,kim2020jsi}, the $L_2$ loss function is commonly employed to optimize networks, as well as in tasks dominated by color mapping~\cite{zeng2020learning}. Thus, we adopt this loss function to optimize AGCM in the initial work. 
Our experiments, however, indicate that the choice of loss function significantly impacts performance for SDRTV-to-HDRTV. As shown in Figure~\ref{loss}, we compared training curves using $L_1$ and $L_2$ loss functions. The model optimized with $L_1$ loss exhibits notably better results than with $L_2$ loss. Previous studies, such as \cite{wang2004image}, have also demonstrated that $L_1$ loss often achieves better convergence and higher final performance in image restoration.
Therefore, in this study, we utilize $L_1$ loss to optimize the AGCM network. As demonstrated in Table~\ref{Table 1 comparison}, AGCM++ achieves significantly improved performance compared to the original AGCM.


\subsection{Conversion from base network to 3D LUT}
\label{Conversion_base_network_3DLUT}
3D LUTs are widely used in practical applications for image style and tone manipulation, especially in movie production, as part of the digital media process. 
There are many tools editing images by directly modifying 3D LUTs, thereby constructing available SDRTV-to-HDRTV 3D LUTs is of great value for practical applications. 
In this section, we show that the base network in our method can be converted into a SDRTV-to-HDRTV 3D LUT with acceptable performance loss.
Concretely, we take a 3D lattice composed of SDRTV colors as the input to the network and obtain the corresponding HDRTV colors. We can then build a lookup table based on these paired data. When performing color transformation, we can use lookup and trilinear interpolation operations as \cite{zeng2020learning}.
As shown in Figure~\ref{Figure_LUT_ablation}, we present the results of two settings for converting our base network to 3D LUT.
Conversion to a small 3D LUT with 33 nodes (i.e., 3DLUT\_s33) results in minimal performance drop (0.16dB), while a large 3D LUT with 64 nodes (i.e., 3DLUT\_s64) almost maintains performance.
This demonstrates the flexibility of our base network to be converted an available SDRTV-to-HDRTV 3D LUT.
Furthermore, our base network' efficiency (with only 5k parameters) allows for efficient training, and it allows modulation according to the condition network to generate customized 3D LUTs.

\subsection{User Study}
\begin{figure}[t]
\centering
\includegraphics[width=\linewidth]{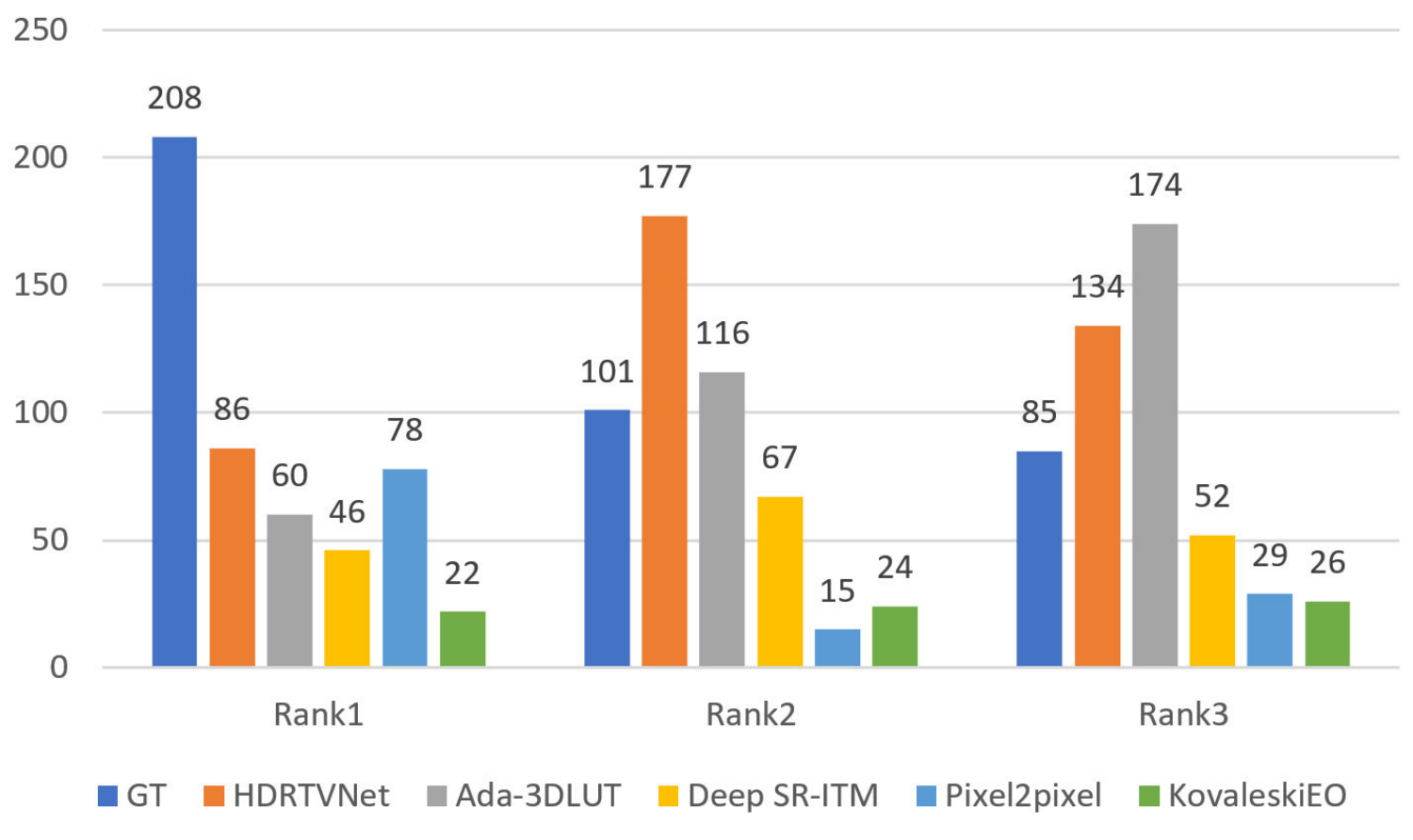}
\caption{User study rankings for different methods. Rank 1 means the best subjective feeling.}
\vspace{-10pt}
\label{user_study_1}
\end{figure}

\begin{figure}[t]
\centering
\includegraphics[width=\linewidth]{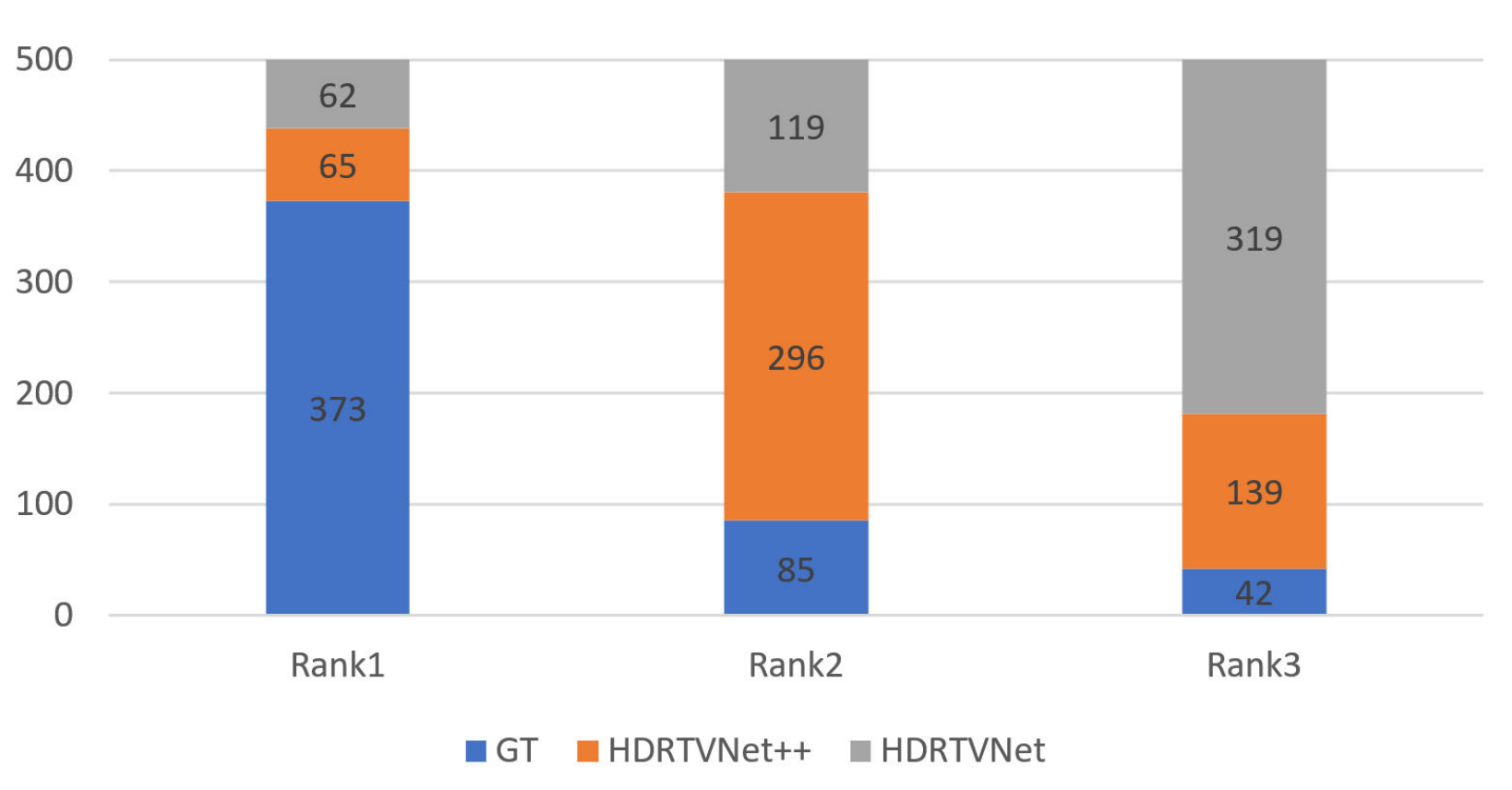}
\caption{User study rankings for HDRTVNet and HDRTVNet++. Rank 1 means the best subjective feeling.}
\vspace{-7pt}
\label{user_study_2}
\end{figure}

In the initial version, we first conduct a user study with 20 participants to evaluate HDRTVNet's visual quality compared to four top-performing methods.
For the experimental setup, a total of 25 images are randomly selected from the testing set and display in a darkroom on an HDR TV (Sony X9500G with a peak brightness of 1300 nits) set to the Rec.2020 color gamut and HDR10 standard.
We then instruct the participants to consider three main factors when evaluating the images: (1) the presence of obvious artifacts and unnatural colors, (2) the naturalness and comfort of the overall color, brightness, and contrast, and (3) the perception of contrast between light and dark levels and highlight details. 
Based on these principles, participants rank the results in each scenario. 
The results of five approaches including Ada-3DLUT \cite{zeng2020learning}, Deep SR-ITM \cite{kim2019deep}, Pixel2pixel \cite{isola2017image}, KovaleskiEO \cite{kovaleski2014high} and HDRTVNet \cite{hdrtvnet_conf}, along with the ground-truth images are compared.
When ranking the images for a scene, participants are able to view six images from different methods simultaneously or compare any two images at will until they decide on the order. We display the counts of different results in the top three ranks, as shown in Figure \ref{user_study_1}. The ground truth (GT) and HDRTVNet account for 41.6$\%$ (208 counts) and 17.2$\%$ (86 counts) of the results considered to have the best visual quality, respectively. Similarly, HDRTVNet accounts for 35.4$\%$ of the results considered to have the second-best visual quality. In conclusion, the results of HDRTVNet are only inferior to the GT in terms of visual quality in subjective evaluation.
We further conduct a user study to compare the proposed HDRTVNet++ with the previous HDRTVNet~\cite{hdrtvnet_conf}.
Participants are asked to rank each set of images in this experiment using the same settings as above.
As shown in Figure \ref{user_study_2}, the ground-truth still achieve the best visual quality, while HDRTVNet++ shows a better ranking over HDRTVNet.
HDRTVNet++ accounts for 59.2$\%$ (296 counts) of the results considered to have the second-best visual quality, which greatly outperforms HDRTVNet 23.8$\%$ (119 counts).
These results demonstrate the superiority of our method in terms of objective visual quality.

\begin{table*}[t]
\begin{threeparttable}[b]
\caption{Quantitative comparisons between the base network and its converted 3D LUTs.}
\label{lut_comparison}
\centering
\tabcolsep=7.2mm
\begin{tabular*}{\linewidth}{c|c|ccccc}
\hline
Method & Params$\downarrow$ & PSNR$\uparrow$ & SSIM$\uparrow$ & SR-SIM$\uparrow$ & $\Delta E_{ITP}\downarrow$ & HDR-VDP3$\uparrow$ \\ \hline
Base network & 5k & 36.14 & 0.9643 & 0.9961 & 10.43 & 8.305 \\ \hline
3DLUT\_s33\tnote{1} & 108k & 35.98 & 0.9645 & 0.9958 & 10.60 & 8.322 \\
3DLUT\_s64 & 786k & 36.13 & 0.9643 & 0.9960 & 10.46 & 8.309 \\ \hline
\end{tabular*}%
\begin{tablenotes}
\item [1] The s33 or s64 represents the size of LUT. LUTs of these two sizes are commonly used in the actual production.
\end{tablenotes}
\end{threeparttable}
\end{table*}

\section{Conclusion}
We have introduced a new SDRTV-to-HDRTV solution pipeline, leveraging a divide-and-conquer strategy based on the SDRTV/HDRTV formation process. Additionally, we developed HDRTVNet++ to address this challenge effectively.
Our approach distinguishes between pixel-independent and region-dependent operations in the formation pipeline, allowing us to implement adaptive global color mapping and local enhancement separately. We design a new color condition network, which offers improved performance with fewer parameters compared to existing methods, to facilitate SDRTV-to-HDRTV color mapping. For enhanced visual results, we employ generative adversarial training to refine highlights.
Furthermore, we construct a new HDRTV dataset for rigorous training and testing. Comprehensive experiments confirm the superiority of our solution, demonstrating significant improvements in both quantitative metrics and visual quality.

\bibliographystyle{IEEEtran}
\bibliography{bibtex}


 





\end{document}